\documentclass[sn-mathphys,iicol]{sn-jnl}

\jyear{2022}%

\theoremstyle{thmstyleone}%
\theoremstyle{thmstyletwo}%

\theoremstyle{thmstylethree}%

\usepackage{float} 

\usepackage{tabularx}
\usepackage{longtable}
\usepackage{booktabs}
\usepackage[T1]{fontenc}
\usepackage[utf8]{inputenc}

\usepackage{graphicx}
\usepackage{comment}
\usepackage{amsmath,amssymb} % define this before the line numbering.
\usepackage{color}
\newcommand{\tabincell}[2]{\begin{tabular}{@{}#1@{}}#2\end{tabular}}
\usepackage{makecell}
\usepackage{caption}
\usepackage{enumerate}
\usepackage{multirow}
\usepackage{multicol}
\usepackage{subfigure}
\usepackage{float}
\usepackage{bbding}
\usepackage{xcolor}
\usepackage{grffile}

\usepackage{mathtools}
\usepackage[linesnumbered,ruled,vlined,algo2e]{algorithm2e}
\SetCommentSty{mycommfont}

\newcommand{\etc}{\textit{etc}.}
\newcommand{\ie}{\textit{i}.\textit{e}.}
\newcommand{\eg}{\textit{e}.\textit{g}.}

\begin{document}

\title[mode=title]{BioDrone: A Bionic Drone-based Single Object Tracking Benchmark for Robust Vision}

\author[1,2]{\fnm{Xin} \sur{Zhao}}\email{xzhaopersonal@foxmail.com}
\author[2]{\fnm{Shiyu} \sur{Hu}}\email{hushiyu2019@ia.ac.cn}
\author[3]{\fnm{Yipei} \sur{Wang}}\email{220213711@seu.edu.cn}
\author[2]{\fnm{Jing} \sur{Zhang}}\email{jing\_zhang@ia.ac.cn}
\author[4]{\fnm{Yimin} \sur{Hu}}\email{ymhu2015@sinano.ac.cn}
\author[4]{\fnm{Rongshuai} \sur{Liu}}\email{rsliu2020@sinano.ac.cn}
\author[5]{\fnm{Haibin} \sur{Ling}}\email{ hling@cs.stonybrook.edu}
\author[6]{\fnm{Yin} \sur{Li}}\email{yin.li@wisc.edu}
\author[4]{\fnm{Renshu} \sur{Li}}\email{lirenshu2021@gusulab.ac.cn}
\author[4]{\fnm{Kun} \sur{Liu}}\email{liukun2021@gusulab.ac.cn}
\author[4]{\fnm{Jiadong} \sur{Li}}\email{jdli2009@sinano.ac.cn}

\affil[1]{\orgdiv{School of Computer and Communication Engineering}, \orgname{University of Science and Technology Beijing}, \orgaddress{\city{Beijing}, \country{China}}}

\affil[2]{\orgdiv{Institute of Automation}, \orgname{Chinese Academy of Sciences}, \orgaddress{\city{Beijing}, \country{China}}}

% \affil[2]{\orgdiv{School of Artificial Intelligence}, \orgname{University of Chinese Academy of Sciences}, \orgaddress{\city{Beijing}, \country{China}}}

\affil[3]{\orgdiv{School of Instrument Science and Engineering}, \orgname{Southeast University}, \orgaddress{\city{Nanjing}, \country{China}}}

\affil[4]{\orgdiv{Suzhou Institute of Nano-tech and Nano-bionics}, \orgname{Chinese Academy of Sciences}, \orgaddress{\city{Suzhou}, \country{China}}}

\affil[5]{\orgdiv{Department of Computer Science}, \orgname{Stony Brook University}, \orgaddress{\city{Stony Brook}, \country{USA}}}

\affil[6]{\orgdiv{Biostatistics \& Medical Informatics Computer Sciences}, \orgname{University of Wisconsin-Madison}, \orgaddress{\city{Madison}, \country{USA}}}

\abstract{
Single object tracking (SOT) is a fundamental problem in computer vision, with a wide range of applications, including autonomous driving, augmented reality, and robot navigation. The robustness of SOT faces two main challenges: \textit{tiny target} and \textit{fast motion}. These challenges are especially manifested in videos captured by unmanned aerial vehicles (UAV), where the target is usually far away from the camera and often with significant motion relative to the camera. 
To evaluate the robustness of SOT methods, we propose \textbf{BioDrone} -- the first \textbf{bio}nic \textbf{drone}-based visual benchmark for SOT. Unlike existing UAV datasets, BioDrone features videos captured from a flapping-wing UAV system with a major camera shake due to its aerodynamics. BioDrone hence highlights the tracking of tiny targets with drastic changes between consecutive frames, providing a new robust vision benchmark for SOT. To date, BioDrone offers the largest UAV-based SOT benchmark with high-quality fine-grained manual annotations and automatically generates frame-level labels, designed for robust vision analyses. Leveraging our proposed BioDrone, we conduct a systematic evaluation of existing SOT methods, comparing the performance of 20 representative models and studying novel means of optimizing a SOTA method (KeepTrack \cite{KeepTrack}) for robust SOT. Our evaluation leads to new baselines and insights for robust SOT. Moving forward, we hope that BioDrone will not only serve as a high-quality benchmark for robust SOT, but also invite future research into robust computer vision. 
The database, toolkits, evaluation server, and baseline results are available at \url{http://biodrone.aitestunion.com}.
    }

\maketitle

\section{Introduction}
\label{sec:introduction}

\begin{figure*}[t!]
    \centering
    \includegraphics[width=\linewidth]{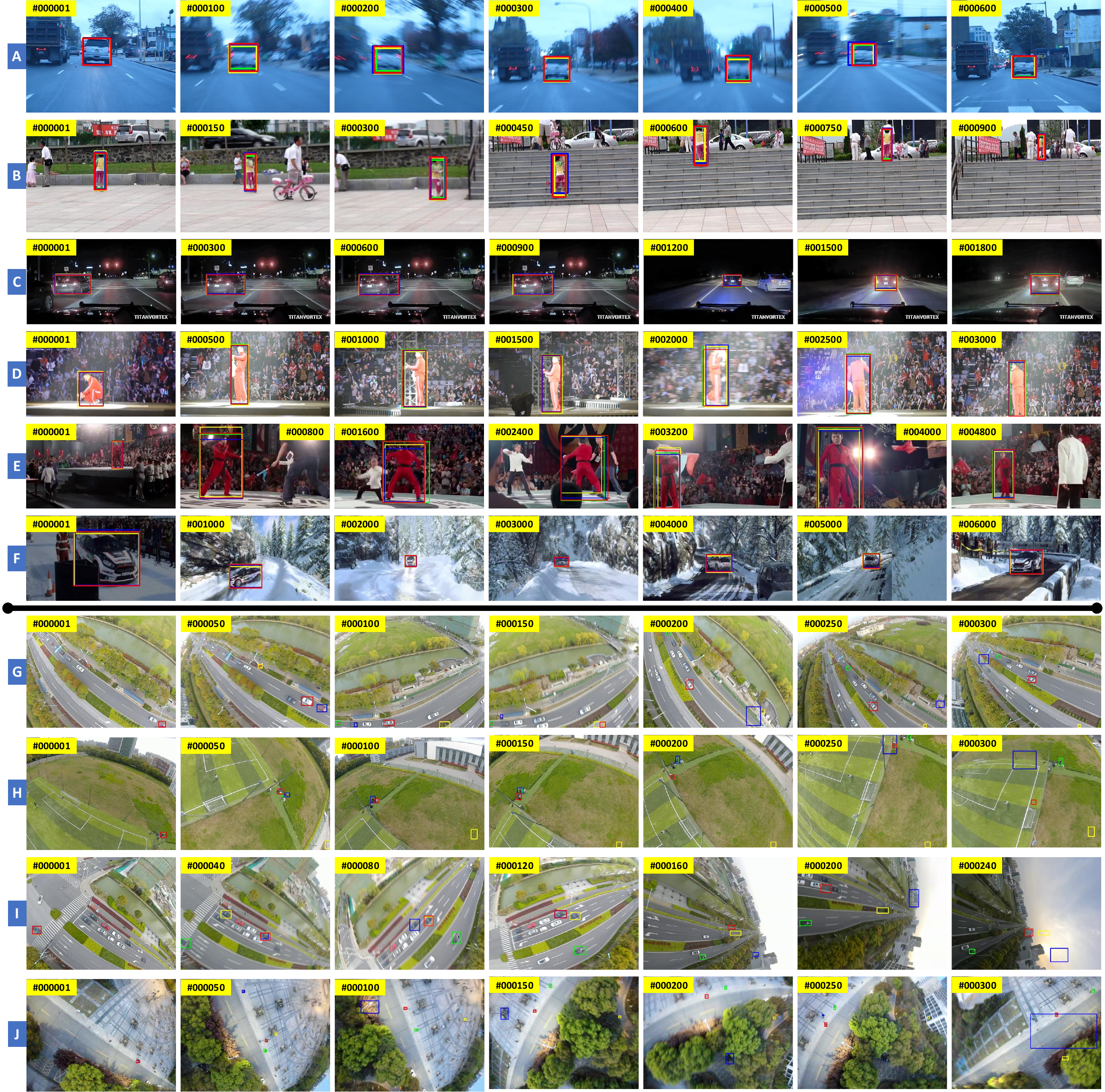}
    \caption{
        This paper aims to study the robust vision problem in visual object tracking; thus, we propose a bionic drone-based SOT benchmark named BioDrone to support this goal. 
        In this figure, we compare BioDrone (G to J) with generic SOT benchmarks represented by VOT short-term tracking competition \cite{VOT2018,VOT2019} (A to B), LaSOT \cite{LaSOT} (C to D), VideoCube \cite{GIT} (E to F). 
        Here we select the same object categories (car and person) in different benchmarks, and add performances of state-of-the-art (SOTA) tracking methods for better comparison (\textcolor{green}{$\blacksquare$} green bounding-box represents ground-truth, \textcolor{yellow}{$\blacksquare$} yellow bounding-box  represents KeepTrack \cite{KeepTrack}, \textcolor{blue}{$\blacksquare$} blue bounding-box represents MixFormer \cite{MixFormer}, \textcolor{red}{$\blacksquare$} red bounding-box represents SiamRCNN \cite{SiamRCNN}). 
        Compared to other benchmarks, BioDrone highlights the challenges of \textit{tiny target} and \textit{fast motion}. The above factors can affect appearance and motion information, bringing troubles to most tracking algorithms on BioDrone. Most SOTA methods lose the target after tens of frames on BioDrone, but they perform well for thousands of frames on other benchmarks.
    }
    \label{fig:motivation}
    \end{figure*}

\begin{figure*}[t!]
    \centering
    \includegraphics[width=0.8\linewidth]{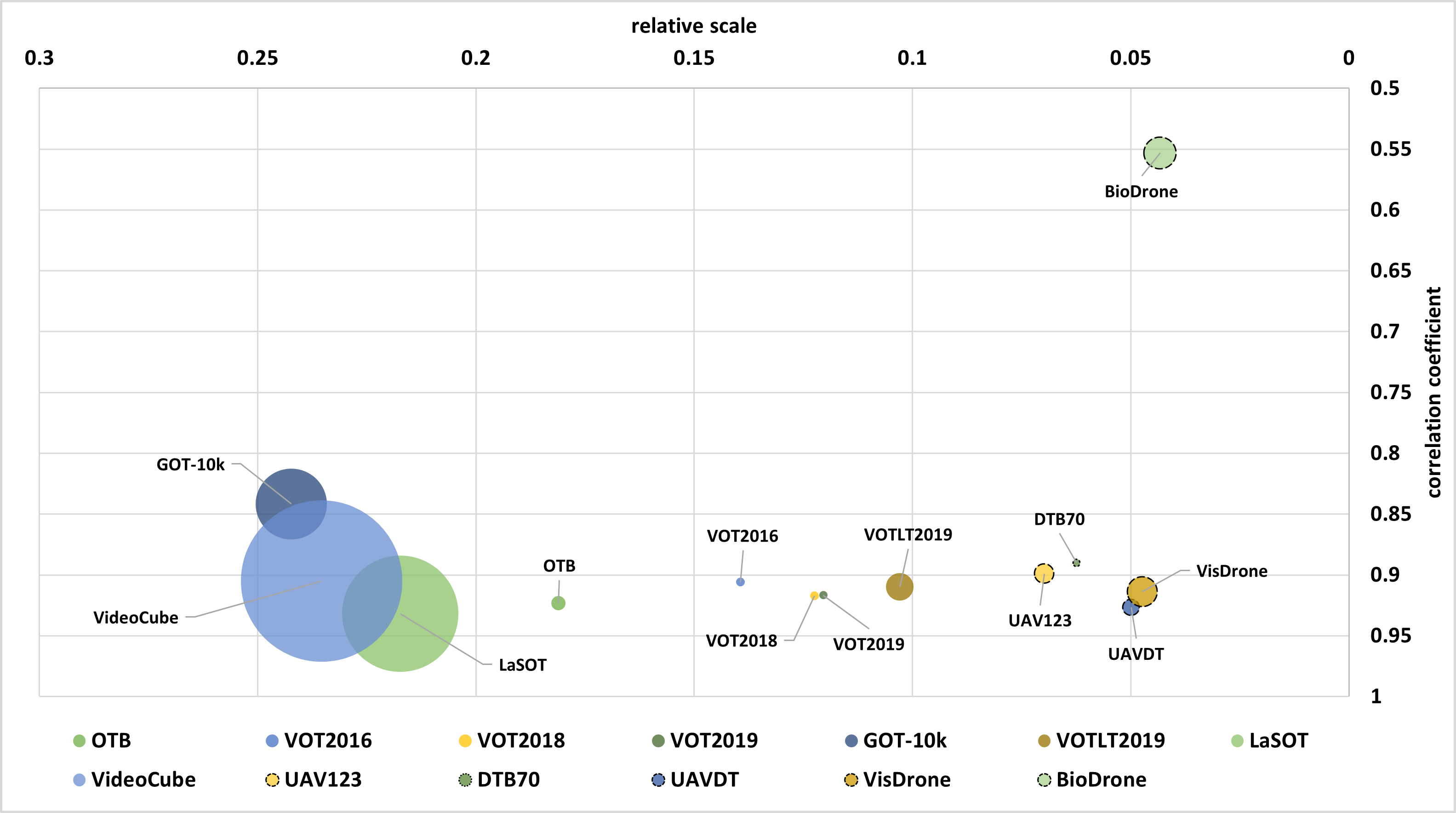}
    \caption{
        Summary of existing SOT benchmarks, including classical benchmarks (OTB100 \cite{OTB2015}, VOT2016 \cite{VOT2016}, VOT2018 \cite{VOT2018}, VOT2019 \cite{VOT2019}, GOT-10k\cite{GOT-10k}, VOTLT2019 \cite{VOT2019}, LaSOT \cite{LaSOT}, Videocube \cite{GIT}), and UAV-based benchmarks (UAV123 \cite{UAV123}, UAVDT \cite{UAVDT}, DTB70 \cite{DTB70}, VisDrone \cite{VisDrone}). 
        The bubble diameter is in proportion to the total frames of a benchmark. The bubbles with dashed borders represent UAV-based benchmarks.
        The horizontal coordinate represents the average relative scale of the target, and the vertical coordinate represents the average correlation coefficient between consecutive frames.
        The proposed BioDrone has a \textit{smaller target size} and \textit{more drastic frame changes} between consecutive frames, with higher demands on the robustness of tracking algorithms.
    }
    \label{fig:sot-bubble}
    \end{figure*}

Single object tracking (SOT) \cite{GOT-10k, GIT}, an essential computer vision task that aims to locate a user-specified moving target, has attracted numerous researchers to propose effective tracking algorithms \cite{KCF, SiamFC, SiamRPN, KeepTrack,MixFormer}. Although existing methods have been widely used in application scenarios like self-driving \cite{kong2022human,dendorfer2021motchallenge}, augmented reality \cite{abu2018augmented,gauglitz2011evaluation} and robot navigation \cite{robot-navigation, ramakrishnan2021exploration}, key challenges like \textit{tiny target} and \textit{fast motion} can still affect the robustness of algorithms. 
SOT is commonly formulated as a sequential decision process (\ie, tracking the current frame should rely on previous frames' tracking results), and corresponding tracking algorithms highly depend on the target's appearance and motion information during execution.
However, the tiny target means that the available appearance information is limited, while fast motion increases the difficulty in modeling motion information, and even the relative movement of the target and camera can disrupt motion continuity.
Therefore, building a high-quality environment for researching the aforementioned challenging factors can contribute to enhancing the robustness of trackers. 

Regrettably, the majority of SOT datasets are designed for generic scenarios, with a primary focus on addressing generalization issues. Thus, they always encompass a wide range of target categories and scene categories, resulting in a sparse distribution of the aforementioned challenging factors. 
Consequently, there is a necessity to establish a dedicated environment that incorporates densely distributed challenging factors to facilitate robustness research.
Compared with generic scenarios that are recorded by fixed or handheld cameras, visual tracking based on unmanned aerial vehicles (UAVs or drones) highlights challenges and requires more visual robustness. 
(1) \textit{Tiny target}: the aerial overhead view causing the target size of a UAV-based system to be much smaller than other traditional datasets. 
(2) \textit{Fast motion}: unlike fixed cameras, UAV-based datasets include both camera and target motion, resulting in frequent and drastic target position changes in consecutive frames.
(3) \textit{Abrupt variation}: due to the long distance between the target and UAV-mounted camera, a slight movement of UAV will lead to a drastic change in its viewpoint, making the visual information (both foreground and background) shift drastically between consecutive frames.

High-quality UAV-based benchmarks with the above challenging factors are critical to developing robust visual tracking algorithms. 
Although existing works have provided an important basis (Table~\ref{tab:datasets}), they still have several shortcomings: 

\begin{itemize}
\item \textbf{Small-scale dataset.} Early UAV datasets \cite{CARPK, DOTA} usually cover only a few thousand images. Although recent works have improved the dataset scale, 
the size of any single task remains relatively small \cite{UAVDT,VisDrone,BIRDSAI}, often insufficient to support data-driven vision algorithms. 
\item \textbf{Scarcity of UAV-based data.} Most UAV datasets \cite{DOTA,UAV123,DTB70,BIRDSAI} contain multiple data sources, such as data collection from websites or data generated from the UAV simulators, but lack UAV data collected in real scenarios. 
\item \textbf{Limited UAV types.} 
% The typical UAVs can be divided into three categories. Compared with fixed-wing or rotary-wing devices, flapping-wing UAVs are entirely designed with bionic structures. 
% It has broader application prospects and introduces more challenges due to its flapping-wing structure. 
UAVs can be classified into fixed-wing, rotary-wind, and flapping-wing vehicles. Among the three, bionic UAVs with flapping-wing structure remains under exploration. 
However, the existing UAV datasets \cite{CARPK,DOTA,UAV123,DTB70,UAVDT,VisDrone,BIRDSAI} all use fixed-wing or rotary-wing UAVs for data collection and lack attention to visual data from the flapping-wing UAVs.
\end{itemize}

The above problems motivate us to focus on new challenges posed by the aerodynamic structure of flapping-wing drones. 
Using the Large Wingspan bionic flight platform, a flapping-wing aircraft with cutting-edge flight performance made by our team, we construct the first \textbf{bio}nic \textbf{drone}-based visual benchmark \textbf{BioDrone} for SOT task. 
% Compared with existing works, we have summarized our characteristics and contributions as follows:
We summarize the characteristics of our benchmark and our contributions as follows. 

\begin{figure*}[t!]
    \begin{center}
    \subfigure[Fixed-wing UAV \cite{hu2022sensaturban}.]
    {
        \includegraphics[width=0.3\linewidth]{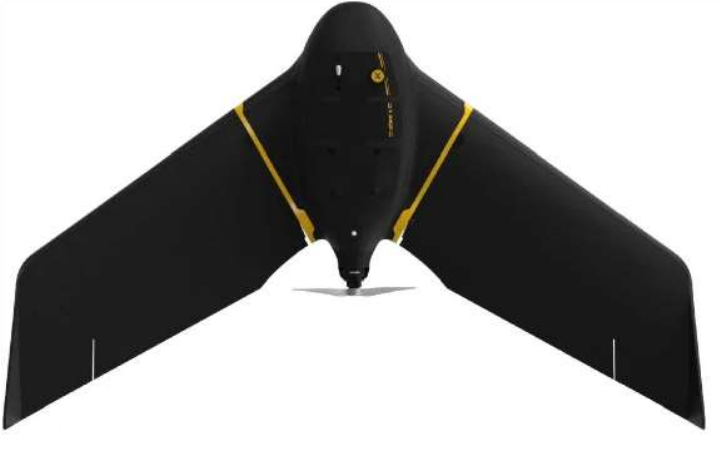}
    }
    \subfigure[Rotary-wing UAV \cite{muller2018sim4cv}.]
    {
        \includegraphics[width=0.33\linewidth]{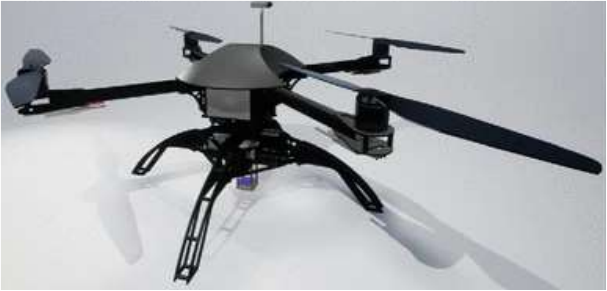}
    }
    \subfigure[Flapping-wing UAV.]
    {
        \includegraphics[width=0.3\linewidth]{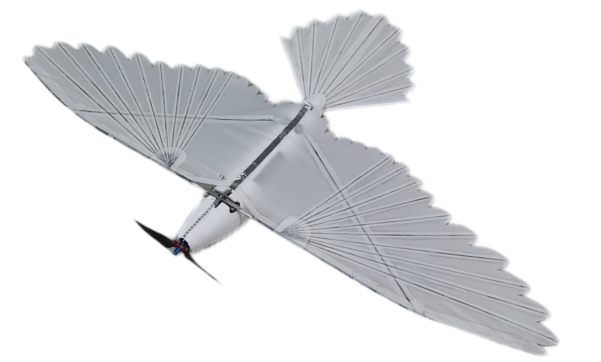}
    }
    \caption{Example of typical UAVs.
    Compared to the other two types of UAVs, flapping-wing UAVs include more challenges due to their bionic mechanical structure.}
    \label{fig:mavs}
    \end{center}
    \end{figure*}

\begin{itemize}
    \item \textbf{Large-scale and high-quality benchmark with robust vision challenges.}
    We take robust vision research as the entry point to construct BioDrone, which includes \textit{600} videos with \textit{304,209} manually labeled frames, and is annotated and reviewed under a precise process. 
    To our knowledge, BioDrone is the first SOT benchmark collected by the bionic-based vision system and the largest UAV-based SOT benchmark.
    Figure~\ref{fig:motivation} qualitatively compares BioDrone to other SOT benchmarks, demonstrating the impact of challenging factors on tracking performance. Most SOTA methods can maintain robust tracking for thousands of frames on generic benchmarks, but easily lose target after tens of frames on BioDrone.
    Figure~\ref{fig:sot-bubble} quantitatively compares BioDrone with others and indicates that \textit{smaller target size} and \textit{more drastic frame changes} between consecutive frames in BioDrone put higher demands on tracking robustness.
    \item \textbf{Videos from Bionic-based UAV.}
    Unlike the existing UAV-based datasets that ignore the flapping-wing UAV structure, our team designs the Large Wingspan bionic flight platform with cutting-edge performance for data collection. Compared with other mechanical structures, the flapping-wing system has broader application prospects due to its lifelike bionic structure. Besides, the flapping-wing design includes additional visual challenges due to more damaging camera shake during the air movements, as shown in Figure~\ref{fig:mavs}.
    \item \textbf{Rich challenging factor annotation.}
    Different from existing UAV-based datasets \cite{UAV123, DTB70, UAVDT, VisDrone} that only provide sequence-level annotation for several challenging factors, BioDrone first provides high-quality fine-grained manual annotations (bounding-box and \textit{occlusion} annotation) and automatically generate frame-level labels for ten challenge attributes, aiming to provide detailed information for further analyses.
    \item \textbf{Effective tracking baseline.} 
    As shown in Figure~\ref{fig:motivation}, challenging factors in BioDrone cause algorithms to fail easily. Thus, we optimize the SOTA method KeepTrack \cite{KeepTrack} and design a new baseline UAV-KT. Besides, we propose a suitable training strategy, and finally achieve a 5\% performance boost in the precision score.
    \item \textbf{Comprehensive experimental analyses.}
    BioDrone contains a complete evaluation mechanism and metrics, compares 20 represent methods and 3 proposed baselines, and analyzes their tracking performance in multiple dimensions, aiming to systematically explore the problems of robust vision brought by flapping-wing UAVs.
\end{itemize}

\section{Related Work}
\label{sec:related_work}

\subsection{Generic SOT Datasets}
\label{sec:sot-dataset}

SOT (\cite{OTB2015}) is a \textit{category-independent} task, which intends to track a moving target without any assumption about the target category. This characteristic allows SOT to be suitable for open-set testing with broad prospects. Since 2013, several generic SOT datasets have been released to support related research.

As one of the earliest benchmarks, \textit{OTB50}\footnote{http://cvlab.hanyang.ac.kr/tracker\_benchmark/index.html}\cite{OTB2013} released in 2013 can be regarded as the earliest SOT benchmark for scientific evaluation. Two years later, \textit{OTB100} \cite{OTB2015} expands the original version for more comprehensive comparisons. Subsequently, the \textit{VOT competition}\footnote{https://votchallenge.net/}\cite{VOT2013, VOT2014,VOT2015,VOT2016,VOT2017,VOT2018,VOT2019,VOT2020,VOT2021} series provide diverse and high-quality datasets to challenge algorithms. 

With the advancement of data-driven trackers, datasets with larger scales are demanded. \textit{GOT-10k}\footnote{http://got-10k.aitestunion.com}\cite{GOT-10k} is a significant high-diversity short-term tracking dataset that comprises 10,000 videos with \textit{one-shot protocol}. Long-term tracking dataset \textit{LaSOT}\footnote{https://cis.temple.edu/lasot/}\cite{LaSOT} has 3.8\textit{m} manually labeled frames with 1,550 videos. It follows the one-shot protocol as well for improving tracking generalization.
Recently, the global instance tracking dataset \textit{VideoCube}\footnote{http://videocube.aitestunion.com}\cite{GIT} is proposed to provide videos with shot-cut and scene-switching. Compared with other SOT datasets, VideoCube not only models the real world comprehensively but also challenges both the perceptual and cognitive components of trackers. 

However, most sequences in these generic benchmarks are collected by fixed cameras, in which the target usually moves smoothly with a notable appearance. The distribution of challenging factors is sparse and usually requires data mining to support robust vision research. 

\subsection{UAV and UAV Vision}
\label{subsec:uav}

%Soaring and overlooking like a bird has been a dream for thousands of years. During the Renaissance, Leonardo da Vinci designed a flapping-wing aerial vehicle by investigating birds. Although the scheme is only a sketch, it provides the original ideas of bionic aircraft for future generations. 
In 1879, French engineer Alphonse Pénaud created a rubber-band-powered aircraft to model the flapping-wing structure, which has been used for toy design due to its straightforward structure.
However, restricted by technology, the research on flapping wing aircraft has progressed slowly. At this stage, the Wright brothers invented plane in 1903, and Paul Kearney prompted helicopter in 1907, causing fixed-wing and rotary-wing aircraft to occupy the sky, and promoting a series of research in the following decades \cite{mcmasters2002airplane,mcmasters2004rethinking,sims1991understanding}. 
Recently, with the development of microcomputers, electrical engineering, and artificial intelligence, UAVs have gradually been favored worldwide, and significantly shortened the gap between enthusiasts and traditional large aircraft. 
UAVs are typically battery-powered, hand-launched, and belly-landed, and can be divided into three types: fixed-wing, rotary-wing, and flapping-wing, as shown in Figure~\ref{fig:mavs}.

The appearances of the first two UAVs are similar to airplanes or helicopters, relying on fixed or rotating wings to provide power for their fuselages, and have been widely used by academia and industry applications, such as intelligent transportation, agricultural procedures, material conveyance, security surveillance, \etc \cite{fraire2015design,barrientos2010rotary}. 
Although the research of fixed-wing and rotary-wing UAVs has become increasingly sophisticated, their structures' shortcomings are also gradually explored.
Defects like large size, insufficient mobility energy, and low efficiency motivate researchers to reconsider designing flapping-wing UAVs -- a kind of bionic aircraft with high lift coefficient and flexible maneuverability for various task situations \cite{lee2018effect,zhang2017review}.
In recent years, flapping-wing UAVs have attracted growing attention due to their flexibility. 
It is worth noting prosperous information obtained by visual sensors installs a pair of "eyes" for the flapping-wing UAVs, enabling a prerequisite for accomplishing various tasks smoothly. 
This section will introduce representative flapping-wing UAVs and their vision system.

In 1988, researchers proposed the first flapping-wing UAV Microbat, which has a 15-20 cm wingspan and 20-30 Hz flapping frequency \cite{pornsin2001microbat}.
In the same year, another flapping-wing UAV, Entomopter, is designed for Mars exploration \cite{rigelsford2004neurotechnology}.
In 2016, DelFly II \cite{de2016delfly}, which contains an airborne stereoscopic perception system (two cameras that can collect visual images simultaneously at 30 Hz), was published for research. 
Flight experiments illustrate that it can successfully detect and avoid walls, but the short battery life and the poor imaging quality ($720 \times 240$ resolution) restrict its application.
Some other researchers modified a commercial flapping-wing UAV and equipped it with a lightweight first-person view (FPV) camera to realize the basic object tracking function \cite{ryu2016autonomous}.
It has a vision algorithm integration system to communicate with the ground control system, which can transfer the captured images to the ground station in real time. 
However, the transmission system has a short communication distance, making it difficult to achieve long-distance tracking. 
Recently, another research group has developed Dove \cite{yang2018dove}, which can transmit color video to the ground station. 
% Experimental verification indicates Dove can further perform visual tasks -- not only the flight time is up to 30 minutes, but also the video transmission distance has reached more than 4 km. 
But its function is mainly limited to aerial photography, and there is still a broad space for development.

Consequently, the visual systems of existing flapping-wing UAVs are all airborne; sensors are mounted on the fuselage and provide environmental information like birds' eyes.
However, specific defects like imaging quality and flight endurance limit the captured visual information. 
% In addition, another type of flapping-wing UAV has an external visual perception system \cite{julian2013cooperative,rosen2016development,festo2018bionicflyingfox}, which uses an external motion capture method to catch the flight trajectories in a specific space. The scope of this article does not cover such external visual systems.
Therefore, although existing research on flapping-wing UAVs has been boosted, it is still difficult to construct high-quality visual datasets like fixed-wing or rotary-wing UAVs.

\subsection{UAV-based Tasks and Datasets}
\label{sec:uav-dataset}

Encouraged by the eye-catching development of UAV-based research, various visual tasks have been applied in UAV systems to process environmental information. 
Since detection and tracking are closely related to UAV vision systems, most UAV-based datasets are constructed to support these two tasks.

\textbf{Object detection} \cite{liu2020deep} aims to accurately determine the category and location of targets, which can be further divided into image object detection (DET \cite{russakovsky2015imagenet}) and video object detection (VID \cite{han2021context}).
It's worth noting that the target category of object detection is generally restricted to pre-defined classes.
\textit{Car Parking Lot (CARPK)}\footnote{https://lafi.github.io/LPN/}\cite{CARPK} is the first large-scale vehicle detection and counting dataset, which is collected by rotary-wing UAVs and covers nearly 90,000 cars in various parking lots. 
\textit{DOTA}\footnote{https://captain-whu.github.io/DOTA/}\cite{DOTA} is another large-scale DET dataset with image resolution ranges from $800 \times 800$ to $20,000 \times 20,000$ pixels. 
% \textit{DroneSURF} \cite{kalra2019dronesurf} \footnote{http://iab-rubric.org/index.php/dronesurf} is a video-based detection dataset, which includes 200 videos with more than 411k frames, intending to facilitate research for face recognition using drones.

\textbf{Object tracking} \cite{VisDrone,wu2021deep} can be further divided into single object tracking (SOT \cite{GIT,LaSOT,GOT-10k}) and multi-object tracking (MOT \cite{luiten2021hota, dendorfer2021motchallenge}). 
MOT usually combines with the VID task -- algorithms should detect objects in the first frame, then calculate the similarity to determine instances with the same ID in consecutive frames. 
Conversely, SOT is a \textit{category-independent} task, which intends to track a moving target without any assumption about the target category. 
\textit{UAV123} and \textit{UAV20L}\footnote{https://cemse.kaust.edu.sa/ivul/uav123}\cite{UAV123} are pioneering works that construct UAV-based SOT datasets from three systems: a rotary-wing UAV, a low-cost UAV, and a UAV simulator (UE4\footnote{https://www.unrealengine.com}). Significant deviation (\eg, target scale and ratio) challenges classical SOT methods and invokes the following research in UAV-based visual tracking. 
\textit{Drone Tracking Benchmark (DTB70)}\footnote{https://github.com/flyers/drone-tracking}\cite{DTB70} includes 70 video sequences to support short-term and long-term tracking. Some sequences are captured by a rotary-wing UAV, while others are collected from YouTube.

Besides, some other UAV datasets are designed to support multiple visual tasks.
\textit{UAV Detection and Tracking (UAVDT)}\footnote{https://sites.google.com/site/daviddo0323/projects/uavdt}\cite{UAVDT} is a large-scale vehicle detection and tracking dataset, which includes 100 video sequences collected by rotary-wing UAVs to support multiple vision tasks like VID, SOT and MOT. 
\textit{VisDrone}\footnote{https://github.com/VisDrone/VisDrone-Dataset}\cite{VisDrone} combines 263 video clips with 179k frames and additional 10k static images to support DET, VID, SOT, and MOT.
Recently, a challenging object detection and tracking dataset\textit{BIRDSAI}\footnote{https://sites.google.com/view/elizabethbondi/dataset}\cite{BIRDSAI} is published. As a multi-modality dataset, it includes 48 real videos collected by a TIR camera mounted on a fixed-wing UAV and 124 synthetic aerial TIR videos generated from AirSim-W simulator \cite{bondi2018airsim}.

Table~\ref{tab:datasets} summarizes the existing generic and UAV-based SOT datasets. Most datasets are collected from websites or simulators, while the limited UAV data comes from rotary-wing or fixed-wing UAVs, lacking visual datasets collected by flapping-wing UAVs. This blank area motivates us to conduct this work and build the first bionic drone-based SOT benchmark to better support robust vision research.

\begin{sidewaystable*}[htbp!]
    \sidewaystablefn%
    \begin{center}
    % \begin{minipage}{\textheight}
    \caption{Summary of existing UAV-based datasets and generic SOT datasets (1\textit{k}=$10^3$, 1\textit{m}=$10^6$). To our knowledge, BioDrone is the first SOT benchmark collected by the bionic-based vision system and the largest UAV-based SOT benchmark.}
    \begin{center}
        \small
        \begin{tabular}{llllllll}
            \toprule
            \multirow{2}{*}{\textbf{Name}} & \multirow{2}{*}{\textbf{Year}} & \multirow{2}{*}{\textbf{Task}} & \multirow{2}{*}{\textbf{\#Frames}} & \multirow{2}{*}{\textbf{\# Videos}} & \multirow{2}{*}{\textbf{Resolution}} & \multicolumn{2}{l}{\textbf{Collection Way}} \\ \cline{7-8} 
            &  &  &  &  &  &  \textbf{UAV-based} & \textbf{Other Sources} \\ \midrule
            CARPK \cite{CARPK} & 2017 & DET & 1.4\textit{k} & -  & 1280*720 &  \tabincell{l}{rotary-wing UAV \\ (DJI Phantom 3 Professional)}  & N \\
            DOTA \cite{DOTA} & 2018 & DET & 2.8\textit{k} & - & various & rotary-wing UAV & \tabincell{l}{ multiple platforms \\ (\eg,Google Earth)} \\
            UAV123 \cite{UAV123} & 2016 & SOT & 110\textit{k} & 123 & 1280*720 & \tabincell{l}{rotary-wing UAV \\ (DJI S1000)} & \tabincell{l}{UAV simulator  \\ (UE4)} \\
            UAV20L \cite{UAV123} & 2016 & SOT & 58.7\textit{k} & 20 & 1280*720 & \tabincell{l}{rotary-wing UAV \\ (DJI S1000)} & \tabincell{l}{UAV simulator  \\ (UE4)} \\
            DTB70 \cite{DTB70} & 2017 & SOT & 15.8\textit{k} & 70 & 1280*720 & \tabincell{l}{rotary-wing UAV \\ (DJI Phantom 2 Vision)} & website \\
            UAVDT \cite{UAVDT} & 2018 & \tabincell{l}{VID, SOT,\\ MOT}  & 80\textit{k} & 100 & 1024*540 & \tabincell{l}{rotary-wing UAV \\ ( DJI Inspire 2) } & N \\
            VisDrone \cite{VisDrone} & 2018 & \tabincell{l}{DET, VID,\\ SOT, MOT} & 179\textit{k}+10\textit{k} & 263 & 3840*2160 & \tabincell{l}{rotary-wing UAV \\ (DJI Mavic and Phantom series) } & N \\
            BIRDSAI \cite{BIRDSAI} & 2020 & VID, MOT & 162\textit{k} & 172 & 640*480 & fixed-wing UAV & \tabincell{l}{UAV simulator \\ (AirSim-W)} \\ \midrule
            \textbf{BioDrone} & \textbf{2022} & \textbf{SOT} & \textbf{304\textit{k}} & \textbf{600} & \textbf{1440*1080} & \textbf{flapping-wing UAV} & \textbf{N} \\ \midrule
            OTB50 \cite{OTB2013} & 2013 & SOT & 29\textit{k} & 59 & various & N & website \\
            OTB100 \cite{OTB2015} & 2015 & SOT & 59\textit{k} & 100 & various & N  & website \\
            VOT2016 \cite{VOT2016} & 2016 & SOT & 21.5\textit{k} & 60 & various & N& website \\
            VOT2017 \cite{VOT2017} & 2017 & SOT & 21.3\textit{k} & 60 & various& N & website \\
            TrackingNet \cite{TrackingNet} & 2018 & SOT & 14.4\textit{m} & 30.6\textit{k} & various & N & website \\
            LaSOT \cite{LaSOT} & 2020 & SOT & 3.87\textit{m} & 1.55\textit{k} & various & N & website \\
            GOT-10k \cite{GOT-10k} & 2021 & SOT & 1.45\textit{m} & 10\textit{k} & various & N & website \\
            VideoCube \cite{GIT} & 2022 & SOT & 7.46\textit{m} & 500 & various & N & website \\ \botrule
            \end{tabular}
    \label{tab:datasets}
    \end{center}
    % \end{minipage}
    \end{center}
    \end{sidewaystable*}

\begin{figure*}[t!]
    \begin{center}
    \subfigure[Schematic diagram of Large Wingspan bionic flight platform and its flight attitudes.]
    {
        \includegraphics[width=\linewidth]{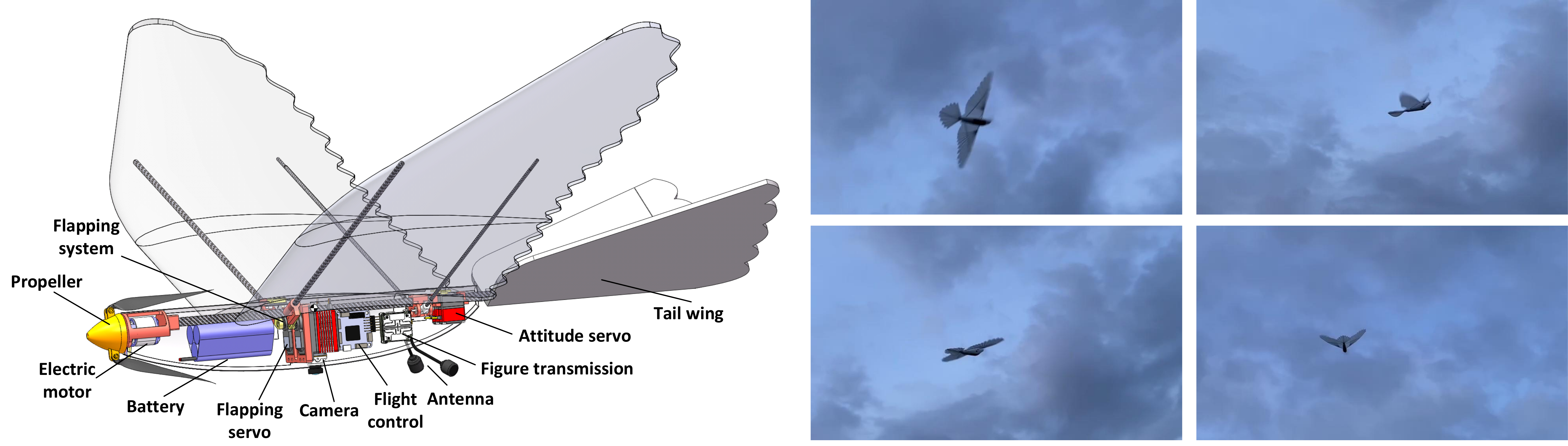}
    }
    \subfigure[The representative data of BioDrone. Each video is strictly collected based on duration, instance classes, main scene categories, and illumination.]
    {
        \includegraphics[width=\linewidth]{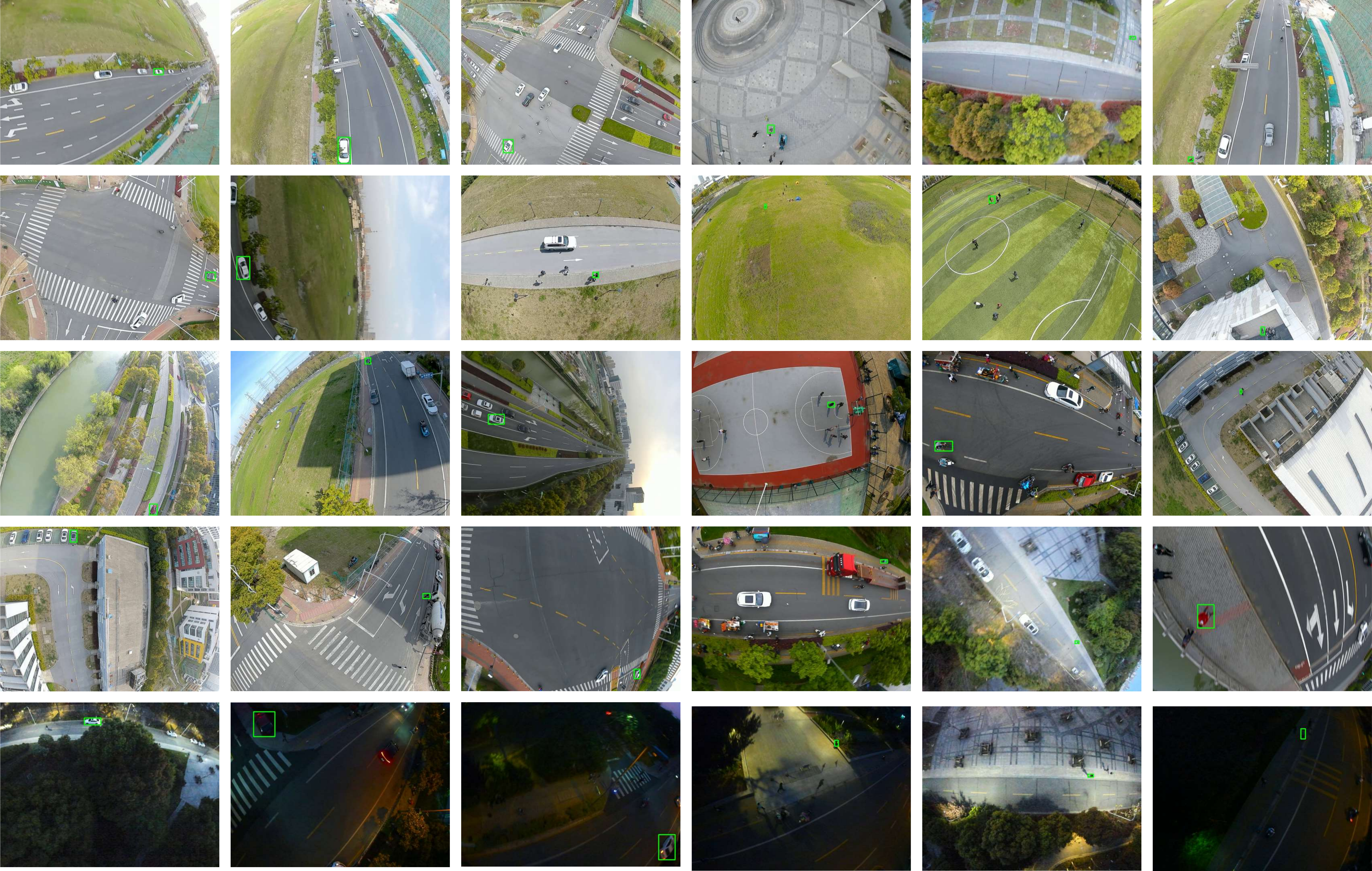}
    }
    \caption{
        Illustrations of the flapping-wing UAV used for data collection and the representative data of BioDrone.
        Different flight attitudes for various scenes under three lighting conditions are included in the data acquisition process, ensuring that BioDrone can fully reflect the robust visual challenges of the flapping-wing UAVs.
    }
    \label{fig:biodrone}
    \end{center}
    \end{figure*}

\begin{figure*}[t!]
    \centering
    \includegraphics[width=\linewidth]{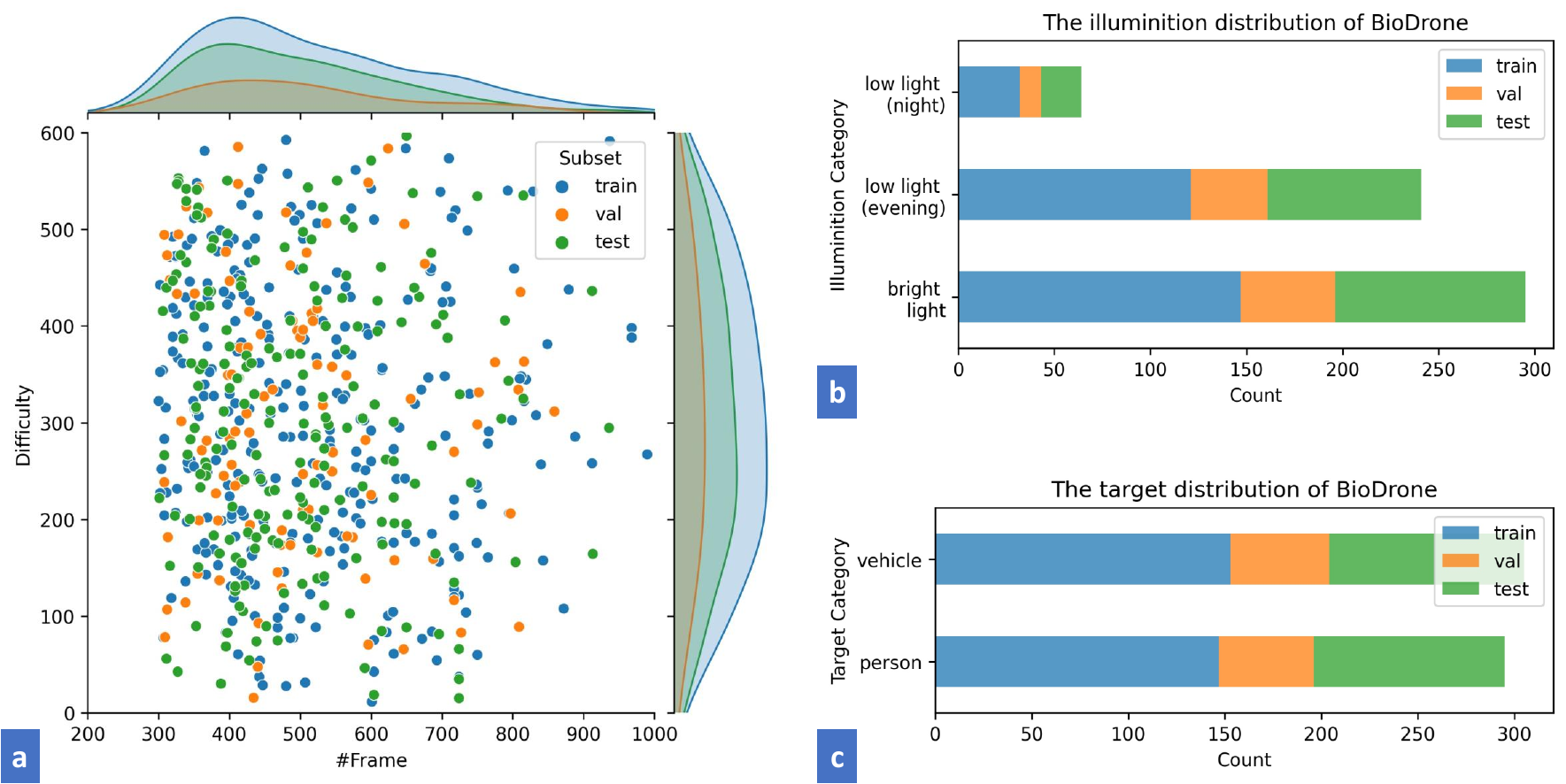}
    \caption{
        Data distribution of BioDrone. The data distribution of different dimensions keeps consistent in each subset. (a) The distribution of sequence lengths and tracking difficulties. (b) The distribution of illumination conditions. (c) The distribution of target categories. 
    }
    \label{fig:statistic}
    \end{figure*}

\begin{figure*}[t!]
    \centering
    \includegraphics[width=\linewidth]{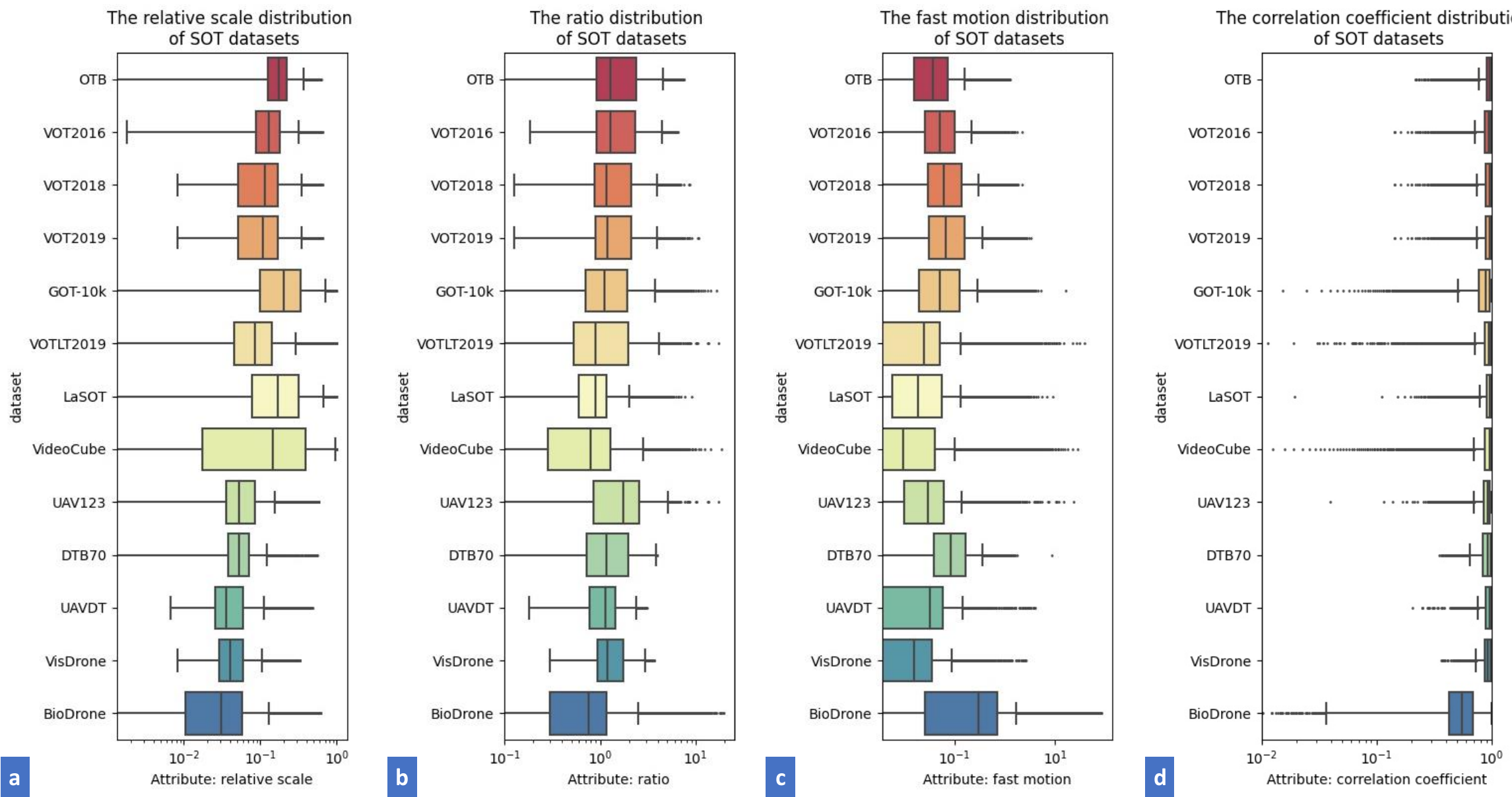}
    \caption{
        Challenging attributes distribution of BioDrone and representative SOT benchmarks. (a) The distribution of relative scale (smaller value means including more tiny targets). (b) The distribution of aspect ratio (smaller or larger value means including more irregular shapes). (c) The distribution of fast motion (larger value means including faster target movement). (d) The distribution of correlation coefficient (smaller value means including more drastic variations between consecutive frames). Clearly, BioDrone includes more \textit{tiny targets} with \textit{more drastic variations} between consecutive frames, and requires more robust methods to accomplish target tracking. 
    }
    \label{fig:dataset-comparison}
    \end{figure*}

\section{BioDrone Benchmark}

A high-quality benchmark labels the target in the video frame and provides criteria for algorithm evaluation. Particularly, benchmarks incorporating multiple challenging factors are critical for training and testing robust trackers.

As summarized in Section~\ref{sec:related_work}, existing benchmarks all ignore collecting data from bionic-based aircraft, motivating us to conduct BioDrone for robust vision research.
BioDrone is collected by a state-of-the-art (SOTA) flapping-wing UAV and annotated under a precise process. It includes \textit{600} videos with \textit{304,209} manually labeled frames. The sequence length varies from \textit{300} to \textit{990} frames, and the average length is around \textit{507}. To our knowledge, BioDrone is the first SOT benchmark collected by a bionic-based aircraft and the largest UAV-based SOT benchmark.

\subsection{Data Collection and Annotation}

\subsubsection{Data Collection}

We use the Large Wingspan bionic flight platform for data acquisition. 
% Large Wingspan is the Suzhou Institute of Nano-tech and Nano-bionics (SINANO), Chinese Academy of Sciences (CAS). 
It is designed with a high degree of biological similarity in appearance and sporty performance, as shown in Figure~\ref{fig:biodrone} (a).
Compared with existing flapping-wing UAVs, Large Wingspan adopts a rotor-flapping composite power arrangement with a single-section wing streamlined aerodynamic layout. 
Its fuselage length is 800mm, wingspan is 1,500mm, biplane flutter frequency is 0-4Hz, and flight altitude is 5-100m. 
Functional loads such as high-definition map transmission and network communication are also deployed in Large Wingspan, ensuring that it can collect visual images from higher altitudes.

In the data acquisition process, we set different flight attitudes for various scenes under three lighting conditions, ensuring that the raw data can fully reflect the robust visual challenges of the flapping-wing UAVs.
In the original date processing process, no post-processing such as frame selection or editing was applied to the collected videos. Therefore, the sequences in the dataset are transformed from real-time recorded videos (30FPS), maintaining a consistent sample rate of 30Hz.

\subsubsection{Data Annotation and Quality Control}

An experienced team precisely labels BioDrone by following two main rules: (1) using the tightest bounding-box to mark the visible part of the user-specified target; and (2) adding an absent label for out-of-view or full-occluded target.
A strict three-round review process is executed to ensure the annotation quality. 
Experienced annotators are trained to conduct the preliminary work and self-inspection, then submit the result to verifiers for second-round verification. Finally, the authors judge whether to accept it in the third-round validation. Any rejection in the above processes will result in the re-annotation to guarantee a high-quality benchmark. The representative data of BioDrone is shown in Figure~\ref{fig:biodrone} (b).

\subsubsection{Subset Division}

We divide BioDrone into the training set (300 videos), the validation set (100 videos), and the test set (200 videos).
The sequence length distribution is illustrated in Figure~\ref{fig:statistic} (a); we ensure that the distribution on the three subsets is essentially the same.
In particular, three representative algorithms (\ie, KeepTrack \cite{KeepTrack}, MixFormer \cite{MixFormer}, and SiamRCNN \cite{SiamRCNN}) are selected to test the 600 videos, and the mean performance of the three trackers is regarded as the score of each sequence. We then organize 600 sequences according to their scores, and finally obtain the difficulty ranking of all data. The distribution of sequence difficulty in each subset is roughly the same.
As shown in Figure~\ref{fig:statistic} (b), BioDrone includes three illumination conditions: bright light (295 videos), low light in the evening (241 videos), and low light at night (64 videos). 
Figure~\ref{fig:statistic} (c) indicates that BioDrone has two main target categories: person (295 videos) and vehicle (305 videos).

\subsection{Challenging Attributes}

The need for robust vision in SOT tasks is primarily from a large number of challenging factors in the environment.
Notably, special collection situations (\eg, lens shake, the unique viewpoint, and the long shooting distance) bring more challenging factors to UAV-based datasets and require more robust algorithms to accomplish tracking tasks.
However, we note that existing UAV-based datasets \cite{UAV123, DTB70, UAVDT, VisDrone} only provide sequence-level annotation for several challenging attributes -- these coarse-grained labels cannot effectively provide detailed information for further analyses.

Therefore, we first provide high-quality frame-by-frame manual annotations (bounding-box and \textit{occlusion} annotation) and automatically generate frame-level labels for ten challenge attributes based on SOTVerse \cite{SOTVerse} and VideoCube \cite{GIT}.

For the $t$-th frame $F_t$ in a sequence $s_i=\{ F_1,F_2,\ldots,F_t,\ldots\}$, BioDrone uses ($x_{t}$, $y_{t}$, $w_{t}$, $h_{t}$) (\ie, the coordinate information of the upper left corner and the shape of the bounding-box) like most classical benchmarks to represent the target bounding-box. 
Challenging attributes in BioDrone are two categories: \textit{static attributes} only relate to the current frame, while \textit{dynamic attributes} record changes between consecutive frames.
The calculation rules for static attributes are as follows:

\begin{itemize}
	\item \textbf{Target aspect ratio and scale.}
    Target \textit{ratio} is defined as $r_{t}={h_{t}}/{w_{t}}$, and target scale is calculated via $s_{t}=\sqrt{w_{t}h_{t}}$. Specifically, we calculate \textit{relative scale} by $s_{t}^{'}=s_{t}/\sqrt{W_{t}H_{t}}$ to eliminate the influence of image resolution, where $W_{t}$ and $H_{t}$ represent the image resolution of $F_{t}$. 
	\item \textbf{Illumination condition.}
    Visual information recorded in special light conditions can be transferred to standard illumination by multiplying a correction matrix $C_{t}$ \cite{ShadeofGray}. Thus, BioDrone quantifies the \textit{illumination} by calculating the Euclidean distance between $C_{t}$ and $\boldsymbol{1}^{1\times 3}$. 
	\item \textbf{Image clarity.}
    BioDrone uses the \textit{blur box} degree to measure the image clarity, which is generated by Laplacian transform \cite{Laplacian}. We convert the RGB bounding-box into gray-scale $G_{t}$, then convolve $G_{t}$ with a Laplacian kernel, and calculate the variance as clarity. 
\end{itemize}

Several dynamic attributes can be directly calculated from static attributes. Correspondingly, the variations of the above static attributes in two sequential frames are defined as \textit{delta ratio}, \textit{delta relative scale}, \textit{delta illumination} and \textit{delta blur box}. Besides, BioDrone also supplies another two dynamic attributes for in-depth analyses:

\begin{itemize}
	\item \textbf{Target motion.}
    \textit{Fast motion} is selected to quantify the target center distance between consecutive frames by $d_{t}={\left\|c_{t}-c_{t-1}\right\|_{2}}/max(s_{i},s_{t-1})$.
    Note that we do not distinguish between the specific causes of the target center distance (\eg, target motion or camera motion), but rather focus on the disruption of the target trajectory due to fast motion. For instance, some SOT algorithms only locate the target position in the next frame within a limited search region near the result of the previous frame. However, fast motion can disrupt the continuity of the target's motion trajectory (\eg, the target's position in the next frame is likely to exceed the search region of the algorithm) and challenge the tracking robustness.
	\item \textbf{Integrated variation between consecutive frames.}
    \textit{Correlation coefficient} is a metric used to measure the similarity between current frame $F_t$ and the previous frame $F_{t-1}$. BioDrone selects the Pearson product-moment correlation coefficient $\rho_t = \frac{\operatorname{cov}(F_t, F_{t-1})}{\sigma_{F_t} \sigma_{F_{t-1}}}$, in which the numerator calculates the covariance of $F_t$ and $F_{t-1}$, and the denominator is the product of the standard deviation. The correlation coefficient reflects the changes between consecutive frames and has been normalized in $[0,1]$.
    % \textit{Correlation coefficient} measures the similarity between progressive frames. Pearson product-moment correlation coefficient $p_i ={\operatorname{cov}(F_i, F_{i-1})}/{\sigma_{F_i} \sigma_{F_{i-1}}}$ is selected to calculate the image covariance, and the denominator is the product of the standard deviation. 
	\end{itemize}
    
To further demonstrate the challenges of BioDrone, we compare the attribute distributions of BioDrone and other SOT benchmarks (frame-level annotations are provided by SOTVerse), then plot the attribute distributions in Figure~\ref{fig:dataset-comparison}. Compared with other SOT benchmarks, BioDrone includes more \textit{tiny targets} (Figure~\ref{fig:dataset-comparison} (a-b)) with \textit{more drastic variations} (Figure~\ref{fig:dataset-comparison} (c-d)) between consecutive frames, which provides a high-quality test bed for further research.

\begin{table*}[t!]
  \begin{center}
      \caption{
        Characteristic of the single object tracing methods in this work (CNN-Convolutional Neural Network. HOG-Histogram of Oriented Gradient.)
        }
      \label{tab:sot-model}
      \small
    \begin{tabular}{lllll}
    \toprule
    \textbf{Tracker} & \textbf{Publish} & \textbf{Feature Representation} & \textbf{Matching Operation} & \textbf{Update} \\ \midrule
    KCF \cite{KCF} & TPAMI'15 & HOG & Correlation Filter & Y \\
    SiamFC \cite{SiamFC} & ECCV'16 & AlexNet & Cross Correlation &  \\
    ECO \cite{ECO} & CVPR'17 & VGG-m & Correlation Filter & Y \\
    SiamRPN \cite{SiamRPN} & CVPR'18 & AlexNet & Cross Correlation &  \\
    DaSiamRPN \cite{DaSiamRPN} & ECCV'18 & AlexNet & Cross Correlation &  \\
    ATOM \cite{ATOM} & CVPR'19 & ResNet-18 & Correlation Filter & Y \\
    SiamRPN++ \cite{SiamRPN++} & CVPR'19 & ResNet-50 & Cross Correlation &  \\
    SiamDW  \cite{SiamDW} & CVPR'19 & ResNet-22 & Cross Correlation &  \\
    DiMP \cite{DiMP} & ICCV'19 & ResNet-50 & Correlation Filter & Y \\
    GlobalTrack \cite{GlobalTrack} & AAAI'20 & ResNet-50 & Hadamard Correlation &  \\
    SiamFC++ \cite{SiamFC++} & AAAI'20 & AlexNet & Cross Correlation &  \\
    Ocean \cite{Ocean} & ECCV'20 & ResNet-50 & Cross Correlation &  \\
    KYS \cite{KYS} & ECCV'20 & ResNet-50 & Correlation Filter & Y \\
    SiamCAR \cite{SiamCAR} & CVPR'20 & ResNet-50 & Cross Correlation &  \\
    PrDiMP \cite{PrDiMP} & CVPR'20 & ResNet-50 & Correlation Filter & Y \\
    SuperDiMP \cite{PrDiMP} & CVPR'20 & ResNet-50 & Correlation Filter & Y \\
    SiamRCNN  \cite{SiamRCNN} & CVPR'20 & ResNet-101 & Concatenate and Re-detection & Y \\
    KeepTrack \cite{KeepTrack} & ICCV'21 & ResNet-50 & Correlation Filter & Y \\
    TCTrack \cite{TCTrack} & CVPR'22 & Temporally Adaptive CNN & Adaptive Temporal Transformer & Y \\
    MixFormer \cite{MixFormer} & CVPR'22 & \multicolumn{2}{c}{Mixed Attention Module} & Y \\ \botrule
\end{tabular}
  \end{center}
  \end{table*}

\section{Trackers}
\label{sec:trackers}

\subsection{Single Object Tracking Methods}
\label{subsec:sot-method}

Table~\ref{tab:sot-model} shows 20 representing SOT algorithms covering both classic and SOTA methods. 
Here, we list the basic information about these trackers.

KCF \cite{KCF} is a classical correlation filter (CF) based method, which balances high speed and tracking accuracy, and becomes a representative tracking framework in the early days. 
ECO \cite{ECO} combines convolutional neural networks (CNN) with CF, aiming to use deep networks to improve feature representation. The feature representation of ECO is a combination of the first and last convolutional layer in the VGG-m \cite{chatfield2014return}, along with histogram of oriented gradient (HOG) \cite{HOG} and color names (CN) \cite{van2009learning}.

As the originator of siamese neural network (SNN) based trackers, SiamFC \cite{SiamFC} achieves satisfactory tracking performance by matching features between the template region and the search region through a simple network structure. It uses AlexNet \cite{AlexNet} for feature representation and matches features via cross-correlation operation.
After that, SiamRPN \cite{SiamRPN} select the region proposal network \cite{FastRCNN} to achieve accurate target regression, 
DaSiamRPN \cite{DaSiamRPN} uses data augmentation to enhance the discriminative ability, 
SiamRPN++ \cite{SiamRPN++} and SiamDW \cite{SiamDW} introduce deeper and wider backbones (ResNet \cite{ResNet}) for feature extraction.
Besides the development of backbone utilization, SiamFC++ \cite{SiamFC++}, Ocean \cite{Ocean}, and SiamCAR \cite{SiamCAR} employ an anchor-free structure (\cite{FCOS}) to eliminate the dependence on anchors. 
Recently, SiamRCNN \cite{SiamRCNN} utilizes a re-detection mechanism (based on FasterRCNN \cite{FasterRCNN}) and proposes a tracklet dynamic programming algorithm to process object disappearance.

Another series of works started by ATOM \cite{ATOM} tries to combine CF and SNN together, and proposes a new framework to combine offline training and online updating. Based on the framework, DiMP \cite{DiMP} optimizes the loss function for stronger discriminative ability, PrDiMP and SuperDiMP \cite{PrDiMP} use probabilistic regression to improve the accuracy.  
KeepTrack \cite{KeepTrack} combines SuperDiMP with a target candidate association network, which is re-trained on hard sequences mined from LaSOT \cite{LaSOT}.

Some other works design custom networks to solve specific problems like target absence or similar instance interference.
GlobalTrack \cite{GlobalTrack} aims to keep tracking performance in long sequences; it does not assume motion consistency and performs a full-image search to eliminate cumulative error. 
KYS \cite{KYS} aims to better use scene information in the tracking process; it represents scene information as state vectors and combines them with the appearance model to locate the object. 
TcTrack \cite{TCTrack} and MixFormer \cite{MixFormer} are the two newest methods based on the transformer structure. 
TcTrack \cite{TCTrack} is designed for object tracking in UAV-based scenes, which aims to fully exploit temporal contexts for aerial tracking.
MixFormer \cite{MixFormer} designs an end-to-end transformer-based framework to simultaneously accomplish feature extraction and target information integration.

% The algorithm evaluation experiments are performed on a server with 4 NVIDIA TITAN RTX GPUs and a 64 Intel(R) Xeon(R) Gold 5218 CPU @ 2.30GHz.

\subsection{New Baselines}
\label{subsec:new-baselines}

\begin{figure*}[t!]
    \centering
    \includegraphics[width=\linewidth]{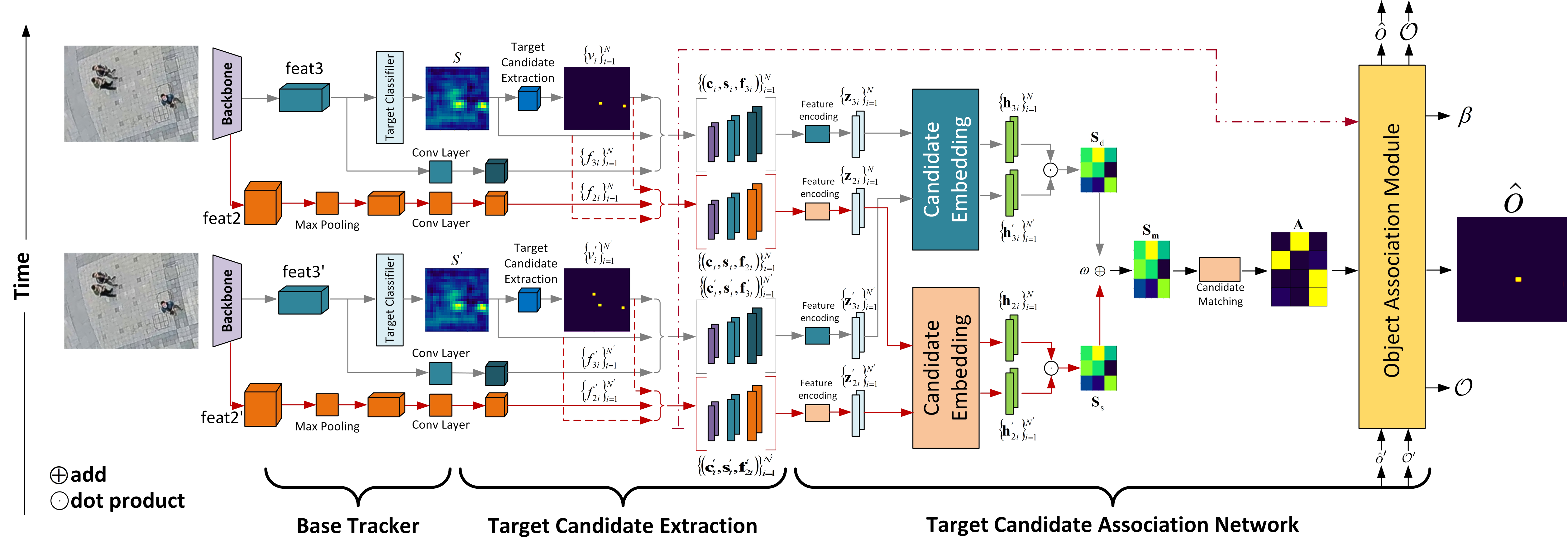}
    \caption
    {
        Overview structure of the proposed new baseline UAV-KT based on KeepTrack\cite{KeepTrack}. The parts connected by red arrows represent our proposed shallow target candidate feature association network module, including target candidate feature extraction, production, embedding, and other operations. The parts connected by gray arrows are the original modules of KeepTrack. The score matrices obtained from different depth features are summed by a learnable coefficient $w$ and perform matching and association operations (the parts connected by black arrows). Since the improvements are closely related and parallel to the original structure of KeepTrack, we draw UAV-KT based on KeepTrack to show the similarities and differences between these two methods clearly.
    }
    \label{fig:UAV-KT}
\end{figure*}

As we analyzed in Section~\ref{sec:introduction}, challenging factors such as \textit{tiny target} and \textit{fast motion} cause algorithms to lose the target easily. Although some methods have combined a re-detection mechanism, fast motion makes it difficult to relocate the target via continuous trajectories, while the small object size significantly limits available appearance information. Thus, it is easy for trackers to relocate interferers rather than the target. Based on the above analyses, we optimize the SOTA method KeepTrack \cite{KeepTrack}, which employs a learned target candidate association network to track both the target and distractor objects, and design a new baseline UAV-KT for BioDrone (Figure~\ref{fig:UAV-KT}).

\subsubsection{Base Model: KeepTrack}

To improve the robust tracking ability when facing similar object interference, KeepTrack \cite{KeepTrack} designs a mechanism to keep track of distractor objects. 
It chooses SuperDiMP \cite{PrDiMP} as the baseline, and adds a learnable correlation network to propagate the identity of all candidate targets in the tracking process.
KeepTrack contains a classification branch and a bounding-box regression branch. The classification branch first obtains the score map through the SuperDiMP network, then generates the coordinates of candidates by selecting points that satisfy the requirements (\ie, the score is a local maximum and should exceed the threshold). 
Afterward, candidates' features are extracted and sent to the target candidate association network for candidate matching and location information generation.
The regression branch follows the IoUNet \cite{IoU-Net} utilized in ATOM \cite{ATOM} to precisely regress the bounding-box, and the target position information obtained from the classification branch is used to obtain and refine its position. Please refer to the original paper for more detailed information on the above two branches. 
Since our improvements are mainly concentrated in the candidate target matching network, here we briefly describe its structure in KeepTrack as follows.

\textbf{Problem formulation.} 
KeepTrack defines the set of target candidates corresponding to the previous frame and the current frame, including distractors and targets, as $V^{'}$ and $V$. $V=\{v_{i}\}^{N}_i$, where N denotes the number of candidates appearing in each frame.
The target candidate association problem for two subsequent frames is also formulated as finding the assignment matrix $A$ between the two sets $V^{'}$ and $V$.

\textbf{Target candidate extraction.} 
KeepTrack first processes the score map by selecting points that meet the requirements as candidate locations and extracts their features.
After that, KeepTrack uses the candidate location $c_{i}$ as a strong cue, then selects the candidate score $s(c_{i})$ and the feature $f_{i}=f(c_{i})$ obtained after a learnable convolutional layer as the other two complementary cues. 
Finally, a feature tuple is created for each candidate and is combined in the following way:

\begin{equation}
z_{i}=f_{i}+\varphi(s_{i},c_{i}),{\forall} v_{i} \in V 
\label{eq:tcf}
\end{equation}

\noindent
where $\varphi$ denotes a multilayer perceptron that maps $s$ and $c$ to the same dimensional space as $f_{i}$. 
% add the meaning of $v_i$ and V

\textbf{Candidate embedding network.} To get more representative candidate features, KeepTrack uses sparse feature matching to exchange $z_{i}$ with bilateral information and self-information. Finally, a new more robust feature representation $h_{i}$ is obtained.

\textbf{Candidate matching.} The similarity matrix $S$, which is obtained by the dot product operation of $S_{i,j}=<h^{'}_{i},h_{j}>$, is used to represent the similarity of candidates in $V^{'}$ and $V$. 
Due to situations like occlusion, disappearance, new appearance, or reappearance, the candidate targets do not necessarily have a definite correspondence within $V^{'}$ and $V$. However, the candidates must have a definite correspondence result to support the following process. 
Therefore, KeepTrack designs a dustbin to match candidates without correspondence \cite{detone2018superpoint,sarlin2020superglue}.
Finally, an augmented assignment matrix $A$ is obtained, in which an additional row and column are added to represent the dustbin. 
Note that a dustbin is a virtual candidate without any feature representation, and a candidate corresponds only if its similarity to all other candidates is low to a dustbin.

\textbf{Object association.} A library $O$ is used to keep track of each object that appears in the scene over time, in which each entry is an object that is visible in the current frame. When tracked online, the estimated assignment matrix $A$ is used to determine the situation of objects (\ie, disappear, newly appear, or remain visible), and the visible objects can be explicitly associated and help in reasoning the target object $\hat{O}$.

Besides, KeepTrack also allows online updating. It describes a memory sample confidence score to decide whether to keep a sample in memory or not, and old samples will be replaced when a fixed memory size is used.

\subsubsection{A New Baseline: UAV-KT}

KeepTrack performs well among the representative SOT trackers in Section~\ref{subsec:sot-method}. However, due to the robust vision challenges introduced by the BioDrone benchmark, the original KeepTrack still has some limitations, motivating us to make appropriate modifications to obtain a more suitable model architecture.

Compared with generic object tracking, the tiny target in BioDrone not only lacks appearance information, but also needs wider receptive fields of deeper-level features to locate its position. 
On the one hand, deep features can obtain rich high-level semantic information, but cannot compensate for the lost pixel information for tiny targets. 
On the other hand, the smaller receptive field of low-level features can avoid the information loss problem, but it mainly extracts spatial information and ignores important semantic information (\eg, assumed as high-level features like temporal and spatial relationships, forward and backward scenes logical relationships, \etc). 
Based on the above analyses, a proper feature fusion module is added in KeepTrack to generate a new baseline named UAV-KT, which aims to improve the capability of tracking tiny targets in BioDrone.

\begin{algorithm}[t!]
  % \footnotesize
  \caption{Target candidate association algorithm}
  \SetKwInput{KwInput}{Input}             
\SetKwInput{KwOutput}{Output}          
\DontPrintSemicolon
\KwInput{\\
$V$: Set of target candidates; \\
${Z'}_{2}(V_i)$: Set of embedded features of the previous frame; \\
$S$:  Depth target candidate feature matching score matrix \\
}
\KwOutput{\\
$\hat{O}$: Target candidate association and matching module \\
}
$N = \lvert V \rvert $ \tcp{Initialize} \leavevmode
\For{$i \leftarrow 1$ \KwTo $N$}{ \leavevmode \tcp{Extraction via id} 
${f_2}(V_i) \leftarrow $ extract from $feat_{backbone}$ \\ \tcp{Extract backbone features} 
    ${f_2}(V_i) \leftarrow \mathsf{MAXPOOL}(\mathsf{CONV}({f_2}(V_i)))$ \\ \tcp{Produce target candidate features} 
    ${z_2}(V_i) = \mathsf{ADD}(\Phi (c(V_i)+s(V_i)),{f_2}(V_i))$ \\ \tcp{Feature integration}
    ${h_2}(V_i), {h'}_{2}(V_i) \leftarrow \mathsf{EMBED}({z_2}(V_i), {z'}_{2}(V_i))$ \\
}
$S_s \leftarrow \{ {h_2}(V_i) \}^{N}_{i=1} \odot \{ {h'}_{2}(V_i) \}^{N}_{i=1}$ \leavevmode\\ \tcp{Obtain score matrix} 
$S_m = \mathsf{ADD}(w[0]*S_d, w[1]*S_s)$ \\ \tcp{Fusion score matrix} 
$\hat{O} \leftarrow $ match and associate by $S_m$ \\ \tcp{Target candidate association and matching module}
return $\hat{O}$
\label{alg:algorithm}
\end{algorithm}

\textbf{Design of target candidate matching network based on different depth backbone features.}
As shown in Figure~\ref{fig:UAV-KT}, the red arrows represent operations of the new target candidate matching network proposed by UAV-KT, in which the feature map in the shallow block of the backbone is selected as a new cue, aiming to enhance the candidate target features and facilitate the ability of target candidate matching.

Unlike the original KeepTrack, we extend the target candidate matching network into two parallel networks for processing backbone features of different depths. The results of these processes are fused to obtain the final matching results. The operation on the shallow features and the information fusion method are described as follows:

\begin{itemize}
\item \textbf{Step 1.} The shallow features $feat_{2}$ of the target candidates extracted from the backbone are fed into a maximum pooling layer and a learnable convolution layer to obtain a more discriminative appearance $f_{2i}$ of the same size as $f_{3i}$.
\item \textbf{Step 2.} $f_{2i}$ is encoded respectively with the target candidate coordinates and scores according to Equation~\ref{eq:tcf} to obtain the shallow target candidate features $z_{2i}$. 
\item \textbf{Step 3.} The shallow target candidate features of the current frame and past frame are fed into the target candidate embedding network for information exchange and extraction, and finally generate richer and more robust features $h_{2i}$, $h'_{2i}$. The dot-product operation is performed on them to obtain the score map $S_{s}$.
\item \textbf{Step 4.} Here, the fusion operation is performed to obtain the final score matrix $S_{m}$. Notably, we introduce a learnable weight $w$ to control the effect of different depth features, which is borrowed from BiFpn \cite{tan2020efficientdet}. The final score matrix is calculated by:
\begin{equation} 
    \label{equ:s_merge}
	\begin{aligned}
	S_{m}&=w[0]*S_{d} + w[1]*S_{s} \\
        w[i]&=\frac{w[i]} {\sum\limits_{i=0}^1 {w[i]}+\varepsilon}\\
    \end{aligned}
\end{equation}
where $w[i]$ denotes the learnable weight set in the net, $\varepsilon$ is a constant, generally set to $1 \times 10^{-4}$.
\item \textbf{Step 5.} Finally, the fused score matrix is used for subsequent operations such as candidate association and object association.
\end{itemize}

\subsubsection{Training Strategies}
\label{subsec:training-strategies}

Unlike large-scale general benchmarks, BioDrone is designed for robust vision research based on the flapping-wing UAV scenario, which contains multiple challenging factors. Therefore, a reasonable training strategy can help trackers enhance robustness in facing challenging factors such as tiny targets, fast motion, and interfering objects.
In this section, we illustrate the detailed training strategies for the BioDrone benchmark and propose the re-trained baselines named KeepTrack* and UAV-KT*.

Generic SOT benchmarks include LaSOT \cite{LaSOT}, GoT-10k \cite{GOT-10k}, and the proposed BioDrone are selected to re-train the base tracker (the left part in Figure~\ref{fig:UAV-KT}), which makes the tracker more robust in tracking tiny targets with fast motion in the UAV-based tasks. 
We sample multiple training and test frames from a video sequence to form training sub-sequences. 
40k sub-sequences with a weight of 1:1:1 for each dataset are obtained for training the base tracker.
The training and testing processes are conducted in a server with 4 NVIDIA TITAN RTX GPUs and a 64 Intel(R) Xeon(R) Gold 5218 CPU @ 2.30GHz. We use adaptive moment estimation (Adam) with a batch size of 32 to train our model, in which the learning rate decay by 0.2 every 20th epoch with a learning rate of $2 \times 10^{-4}$. 
We train 30 epochs and freeze the first half of the weights of the backbone network during the training period. 

The original KeepTrack and the proposed UAV-KT are trained based on the above training strategy to generate KeepTrack* and UAV-KT*. 
Furthermore, we notice that a proper training strategy is important -- training different parts of the module (\eg, the target candidate association network) by BioDrone may decrease the performance of the original versions.
Please refer to Section~\ref{subsubsection:training-strategies} for detailed results and analyses.

\section{Evaluation and Experiments}

\subsection{Evaluation Protocol}

\subsubsection{Mechanisms}

\begin{figure*}[t!]
    \begin{center}
    \subfigure[One pass evaluation (OPE) mechanism by OTB benchmark \cite{OTB2013}.]
    {
        \includegraphics[width=\linewidth]{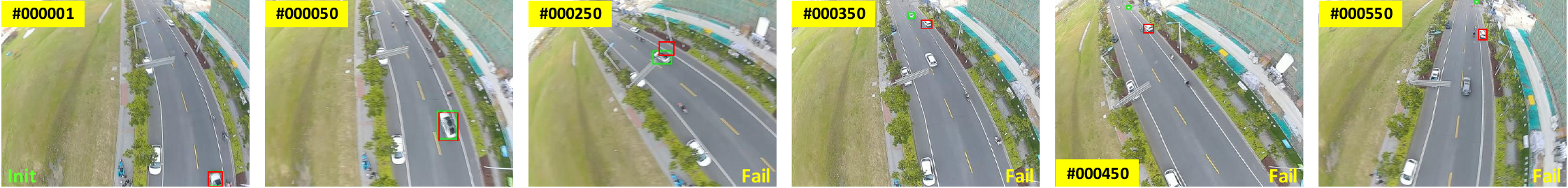}
    }
    \subfigure[OPE system with re-initialization (R-OPE) mechanism by VideoCube benchmark \cite{GIT}. ]
    {
        \includegraphics[width=\linewidth]{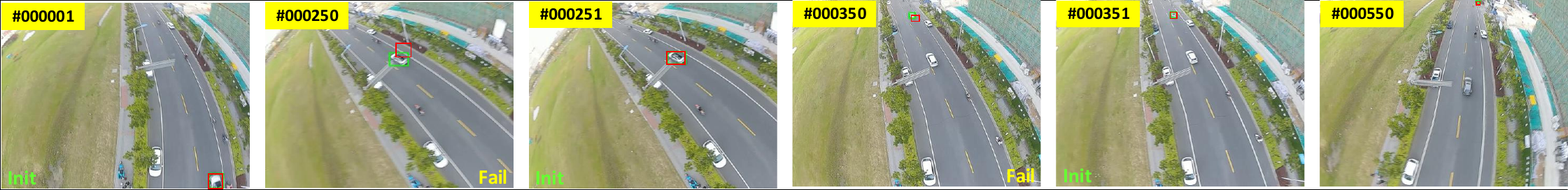}
    }
    \caption{Execution process of two evaluation mechanisms. (a) The traditional OPE mechanism proposed by the OTB benchmark, in which the trackers keep tracking during the whole sequence. (b) The R-OPE mechanism proposed by VideoCube, in which trackers will be re-initialized in the next frame when tracking failure (\ie, the IoU of predicted result $p_t$ and ground-truth $g_t$ $\frac{p_t\bigcap g_t}{p_t\bigcup g_t} < 0.5$) occurs.}
    \label{fig:evaluation-mechanism}
    \end{center}
    \end{figure*}

\begin{sidewaystable*}[htbp!]
    \sidewaystablefn%
    \begin{center}
    % \begin{minipage}{\textheight}
    \caption{
    % \textcolor{red}{$\blacksquare$}$\blacksquare$
    Performance of generic SOT trackers and the proposed baselines based on OPE and R-OPE mechanisms. The top-4 trackers are highlighted by \textcolor{red}{red}, \textcolor{violet}{violet}, \textcolor{blue}{blue}, and \textcolor{teal}{teal}.
    Clearly, the proposed UAV-KT* baseline performs better in different evaluation mechanisms and metrics, demonstrating its robustness improvement compared to the baseline KeepTrack \cite{KeepTrack}.
    }
    \begin{center}
        \small
        \begin{tabular}{llllllllll}
        \toprule
        \multirow{2}{*}{\textbf{Tracker}} & \multicolumn{3}{l}{\textbf{OPE Mechanism}} & & \multicolumn{5}{l}{\textbf{R-OPE Mechanism}} \\ \cline{2-4} \cline{6-10} 
         & \textbf{$P_{score}$$\uparrow$ } & \textbf{$P^{'}_{score}$$\uparrow$} & \textbf{$S_{score}$$\uparrow$} & & \textbf{$P_{score}$$\uparrow$} & \textbf{$P^{'}_{score}$$\uparrow$} & \textbf{$S_{score}$$\uparrow$} & \textbf{$L_{max}$$\uparrow$} & \textbf{$R_{count}$$\downarrow$} \\ \midrule
        KCF \cite{KCF} & 0.052 & 0.077 & 0.047 & & 0.363 & 0.438 & 0.311 & 2.880 & 13.980 \\
        SiamFC \cite{SiamFC} & 0.131 & 0.161 & 0.104 & & 0.535 & 0.583 & 0.414 & 44.775 & 9.015 \\
        SiamDW \cite{SiamDW} & 0.151 & 0.210 & 0.126 & & 0.550 & 0.626 & 0.434 & 50.340 & 8.640 \\
        Ocean \cite{Ocean} & 0.158 & 0.167 & 0.134 & & 0.625 & 0.636 & 0.507 & 73.740 & 7.800 \\
        SiamFC++ \cite{SiamFC++} & 0.162 & 0.180 & 0.139 & & 0.568 & 0.594 & 0.482 & 71.175 & 8.595 \\
        DaSiamRPN \cite{DaSiamRPN} & 0.163 & 0.188 & 0.133 & & 0.551 & 0.605 & 0.448 & 69.870 & 8.395 \\
        SiamRPN \cite{SiamRPN} & 0.173 & 0.199 & 0.139 & & 0.557 & 0.606 & 0.448 & 69.115 & 8.245 \\
        SiamCAR \cite{SiamCAR} & 0.213 & 0.235 & 0.178 & & 0.655 & 0.672 & 0.530 & 94.000 & 6.480 \\
        TCTrack \cite{TCTrack} & 0.231 & 0.255 & 0.192 & & 0.644 & 0.671 & 0.529 & 90.095 & 6.725 \\
        GlobalTrack \cite{GlobalTrack} & 0.237 & 0.249 & 0.183 & & 0.560 & 0.570 & 0.451 & 55.515 & 6.865 \\
        ECO \cite{ECO} & 0.243 & 0.299 & 0.184 & & 0.678 & 0.739 & 0.510 & 85.330 & 6.130 \\
        SiamRPN++ \cite{SiamRPN++} & 0.315 & 0.337 & 0.241 & & 0.685 & 0.703 & 0.528 & 116.620 & 5.030 \\
        ATOM \cite{ATOM} & 0.341 & 0.385 & 0.285 & & 0.754 & 0.787 & 0.623 & 141.420 & 4.180 \\
        DiMP \cite{DiMP} & 0.379 & 0.412 & 0.318 & & 0.763 & 0.788 & 0.635 & 151.330 & 3.960 \\
        KYS \cite{KYS} & 0.380 & 0.411 & 0.315 & & 0.771 & 0.798 & 0.642 & 158.735 & 3.770 \\
        PrDiMP \cite{PrDiMP} & 0.409 & 0.433 & 0.341 & & 0.777 & 0.796 & 0.652 & 156.905 & 3.540 \\
        SuperDiMP \cite{PrDiMP} & 0.426 & 0.447 & 0.361 & & 0.784 & 0.796 & 0.658 & 163.785 & 3.565 \\
        MixFormer \cite{MixFormer} & 0.458 & 0.466 & 0.399 & & 0.782 & 0.786 & \textcolor{blue}{0.675} & 159.110 & 3.33 \\
        SiamRCNN \cite{SiamRCNN} & 0.468 & 0.474 & 0.394 & & 0.720 & 0.726 & 0.616 & 119.335 & 4.455 \\
        \textbf{KeepTrack} \cite{KeepTrack} & \textcolor{teal}{0.504} & \textcolor{teal}{0.523} & \textcolor{teal}{0.424} & & \textcolor{blue}{0.803} & \textcolor{blue}{0.817} & \textcolor{teal}{0.673} & \textcolor{teal}{170.000} & \textcolor{teal}{3.075} \\ \midrule
        \textbf{UAV-KT} & \textcolor{blue}{\tabincell{l}{0.513 \\ (0.009$\uparrow$)}} & \textcolor{blue}{\tabincell{l}{0.537 \\ (0.014$\uparrow$)}} & \textcolor{blue}{\tabincell{l}{0.428 \\ (0.003$\uparrow$)}} & & \textcolor{teal}{0.797} & \textcolor{teal}{0.814} & 0.663 & \textcolor{blue}{172.660} & \textcolor{blue}{\tabincell{l}{2.930 \\ (0.145$\downarrow$)}} \\
        \textbf{KeepTrack*} & \textcolor{violet}{\tabincell{l}{0.538 \\ (0.034$\uparrow$)}} & \textcolor{violet}{\tabincell{l}{0.551 \\ (0.028$\uparrow$)}} & \textcolor{violet}{\tabincell{l}{0.457 \\ (0.033$\uparrow$)}} & & \textcolor{red}{0.832} & \textcolor{red}{0.838} & \textcolor{red}{0.703} & \textcolor{red}{181.185} & \textcolor{violet}{\tabincell{l}{2.660 \\ (0.415$\downarrow$)}} \\
        \textbf{UAV-KT*} & \textcolor{red}{\tabincell{l}{0.554 \\ (0.050$\uparrow$)}}  & \textcolor{red}{\tabincell{l}{0.568 \\ (0.045$\uparrow$)}} & \textcolor{red}{\tabincell{l}{0.466 \\ (0.041$\uparrow$)}} & & \textcolor{violet}{0.822} & \textcolor{violet}{0.832} & \textcolor{violet}{0.691} & \textcolor{violet}{180.595} & \textcolor{red}{\tabincell{l}{2.605 \\ (0.470$\downarrow$)}} \\ \botrule
        \end{tabular}
    \label{tab:results}
    \end{center}
    % \end{minipage}
    \end{center}
    \end{sidewaystable*}

SOT tasks use two evaluation systems -- OPE and the re-initialization mechanism (R-OPE). 
OPE mechanism initializes a tracker in the first frame and continuously records the results, which has been widely used by classical benchmarks \cite{OTB2015,LaSOT,GOT-10k}.
Recently, VideoCube \cite{GIT} provides the R-OPE mechanism, which re-initializes the tracker when it fails in ten consecutive frames.
BioDrone provides the above two mechanisms for performance evaluation, as shown in Figure~\ref{fig:evaluation-mechanism}.

\subsubsection{Metrics}

For the $t$-th frame $F_t$ in a sequence $s_i=\{ F_1,F_2,\ldots,F_t,\ldots\}$, the positional relationship (\eg, intersection over union (IoU) and center distance) between predicted result $p_t$ and ground-truth $g_t$ is usually selected to calculate tracking performance. Like other SOT benchmarks, all evaluation indicators in BioDrone are based on the relationship between two bounding-boxes and their center points (\ie, the predicted center point $c_{p}$ and the actual center point $c_{g}$).
Note that target absent is regarded as an empty set (\ie, $g_t = \phi $). 

\textbf{Precision (PRE).}
Traditional \textit{precision} score is calculated by: 

% \begin{footnotesize}
\begin{equation} 
    \label{equ:pre}
	\begin{aligned}
        d_c &= {\left\|c_{p}-c_{g}\right\|_{2}} \\
        \mathcal{P} (\theta_{d}) &=  \frac{1}{\lvert \mathcal{G} \rvert} \sum_{s_i \in \mathcal{G} } \frac{1}{\lvert s_i \rvert } \lvert \left \{ F_t: d_c \le \theta_{d} \right \} \rvert \\
        P_{score} &=  \frac{1}{\lvert \mathcal{G} \rvert} \sum_{s_i \in \mathcal{G} } \frac{1}{\lvert s_i \rvert } \lvert \left \{ F_t: d_c \le 20 \right \} \rvert \\
	\end{aligned}
	\end{equation}
% \end{footnotesize}

\noindent
where $\lvert \cdot \rvert$ is the cardinality, $\theta_{d}$ is a threshold to judge whether the tracking result is precise. 
The precision score of $s_i$ is defined as the proportion of frames whose center distance $d_c  \le \theta_{d}$. Calculating the mean value of each sequence $s_i$ under video group $\mathcal{G}$ can generate the final precision score $\mathcal{P} (\mathcal{G})$.
Previous works \cite{OTB2015, LaSOT,TrackingNet} usually draw the statistical results based on different $\theta_{d}$ into a curve named \textit{precision plot}. 
Typically, $\theta_{d}=20$ is widelyused to rank trackers ($P_{score}$).

\textbf{Normalized precision (N-PRE).} 
Recent work \cite{GIT} indicates that the PRE score ignores the influence of the target scale, and provides a normalized precision score named N-PRE to solve this problem. Trackers with a predicted center outside the ground-truth rectangle will add a penalty item ${d_c}^{p}$ (\ie, the shortest distance between center point $c_{p}$ and the ground-truth edge). For trackers whose center point falls into the ground-truth rectangle, the center distance ${d_c}^{'}$ equals the original precision $d_c$ (\ie, ${d_c}^{p}=0$). 
Besides, to exclude the influence of target size and frame resolution, N-PRE selects the maximum value in frame $F_t$ to normalize the result. The calculation can be summarized as:

% \begin{footnotesize}
\begin{equation} 
  \label{equ:npre}
	\begin{aligned}
        \mathcal{N}({d_c}^{'}) &= \frac{{d_c}^{'}} {\max ( \{{d_i}^{'} \mid i \in F_t \} )}\\
        \mathcal{P^{'}} (\theta_{d}^{'}) &= \frac{1}{\lvert \mathcal{G} \rvert} \sum_{s_i \in \mathcal{G} } \frac{1}{\lvert s_i \rvert } \lvert \left \{ F_t: \mathcal{N}({d_c}^{'}) \le {\theta_{d}^{'}} \right \} \rvert \\
        P^{'}_{score} &= \frac{1}{\lvert \mathcal{G} \rvert} \sum_{s_i \in \mathcal{G} } \frac{1}{\lvert s_i \rvert } \lvert \left \{ F_t: c_p \in g_t \right \} \rvert \\
	\end{aligned}
	\end{equation}
% \end{footnotesize}

\noindent
Draw statistical results based on different ${\theta_{d}}^{'} \in [0,1] $ into a curve generates the \textit{normalized precision plot}. 
Particularly to overcome the influence of threshold selection, the proportion of frames whose predicted results successfully fall in the ground-truth rectangle is used to rank trackers ($P^{'}_{score}$).

\textbf{Success.} 
Like the calculation process in the precision plot, traditional \textit{success} score of frame $F_t$ is calculated by: 

% \begin{footnotesize}
\begin{equation} \label{equ:success}
	\begin{aligned}
	s_t &= \Omega (p_t,g_t) = \frac{p_t\bigcap g_t}{p_t\bigcup g_t} \\
        \mathcal{S} (\theta_{s}) &=  \frac{1}{\lvert \mathcal{G} \rvert} \sum_{s_i \in \mathcal{G} } \frac{1}{\lvert s_i \rvert } \lvert \left \{ F_t: s_t \le \theta_{s} \right \} \rvert \\
        S_{score} &=  \frac{1}{\lvert \Theta_{s} \rvert} \sum_{\theta_{s} \in \Theta_{s} } \mathcal{S} (\theta_{s}) \\
	\end{aligned}
	\end{equation}
% \end{footnotesize}

\noindent
where $\Omega (\cdot )$ is the intersection over union.
Recent work \cite{GIT} also implements two more success scores based on generalized IoU (GIoU \cite{GIoU}) and distance IoU (DIoU \cite{DIoU}) for calculation.
Frames with overlap $s_t \ge \theta_{s}$ are defined as successful tracking. 
Draw the results based on various overlap threshold $\theta_{s}$ into a curve is the \textit{success plot}, where the mAO (mean average overlap) is widelyused to rank trackers ($S_{score}$).

\textbf{Robustness in R-OPE.} 
The \textit{robust plot} aims to exhibit the performance of trackers in the R-OPE mechanism. Each sequence is divided into several segments by the tracker's re-initialization points, thus the longest sub-sequence that a tracker successfully runs and the re-initialization points can be used to represent the robustness of the tracking process. Taking the number of restarts ($R_{count}$) and the average value of the longest sub-sequence $L_{max}$ as abscissa and ordinate can generate a robust plot. Trackers closer to the upper left corner perform better (indicating successful tracking in longer sequences with rare re-initializations).
Note that we do not limit the number of restarts under the R-OPE mechanism. Thus, we cannot only evaluate an algorithm by the above three metrics, since the high scores may be generated by frequent re-initializations. Therefore, the most reasonable metric for the R-OPE mechanism is the robustness plot and the number of restarts ($R_{count}$).

\begin{figure}[htbp!]
    \begin{center}
    \subfigure[Precision plot.]
    {
        \includegraphics[width=\linewidth]{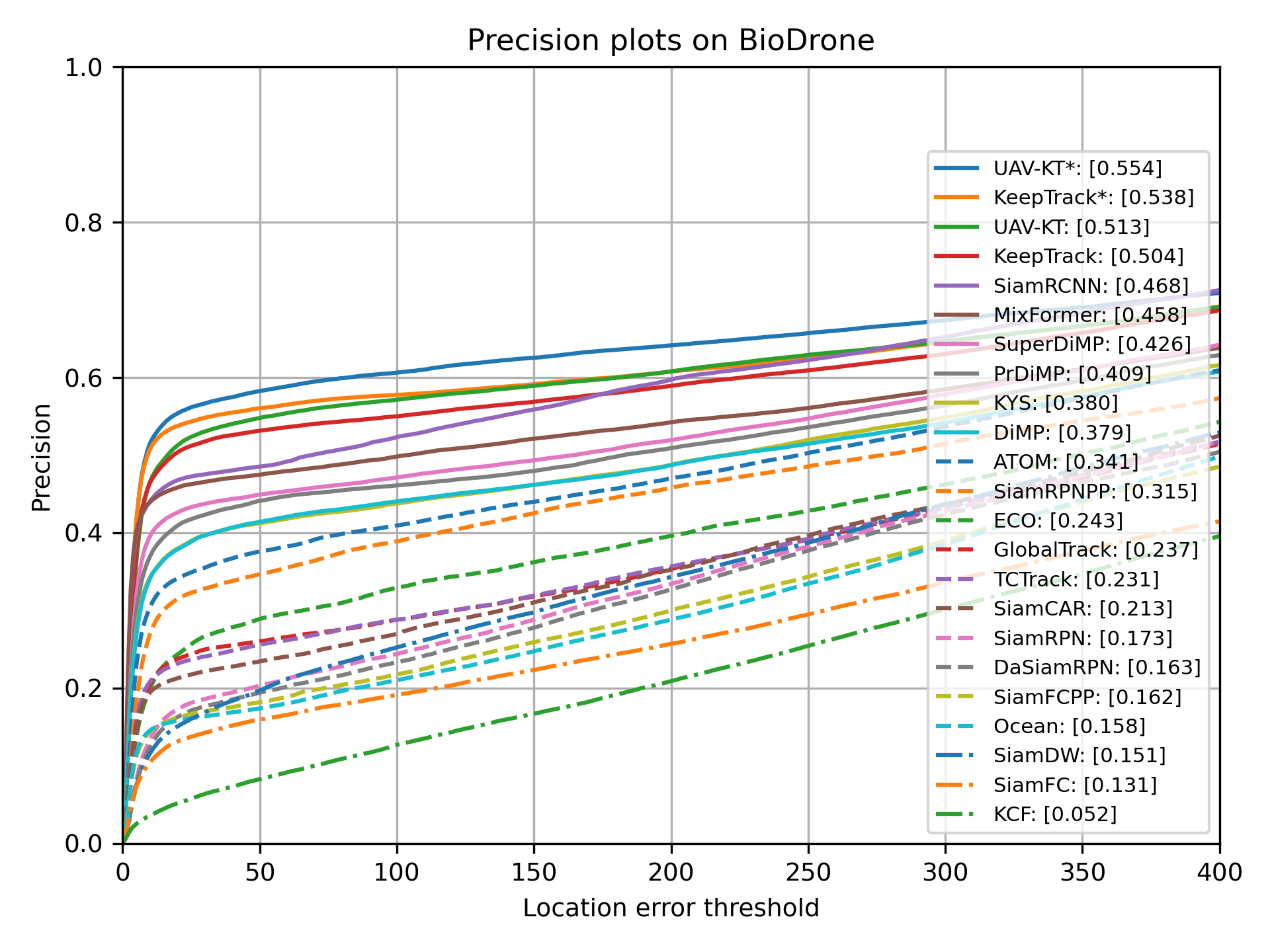}
    }
    \subfigure[Normalized precision plot. ]
    {
        \includegraphics[width=\linewidth]{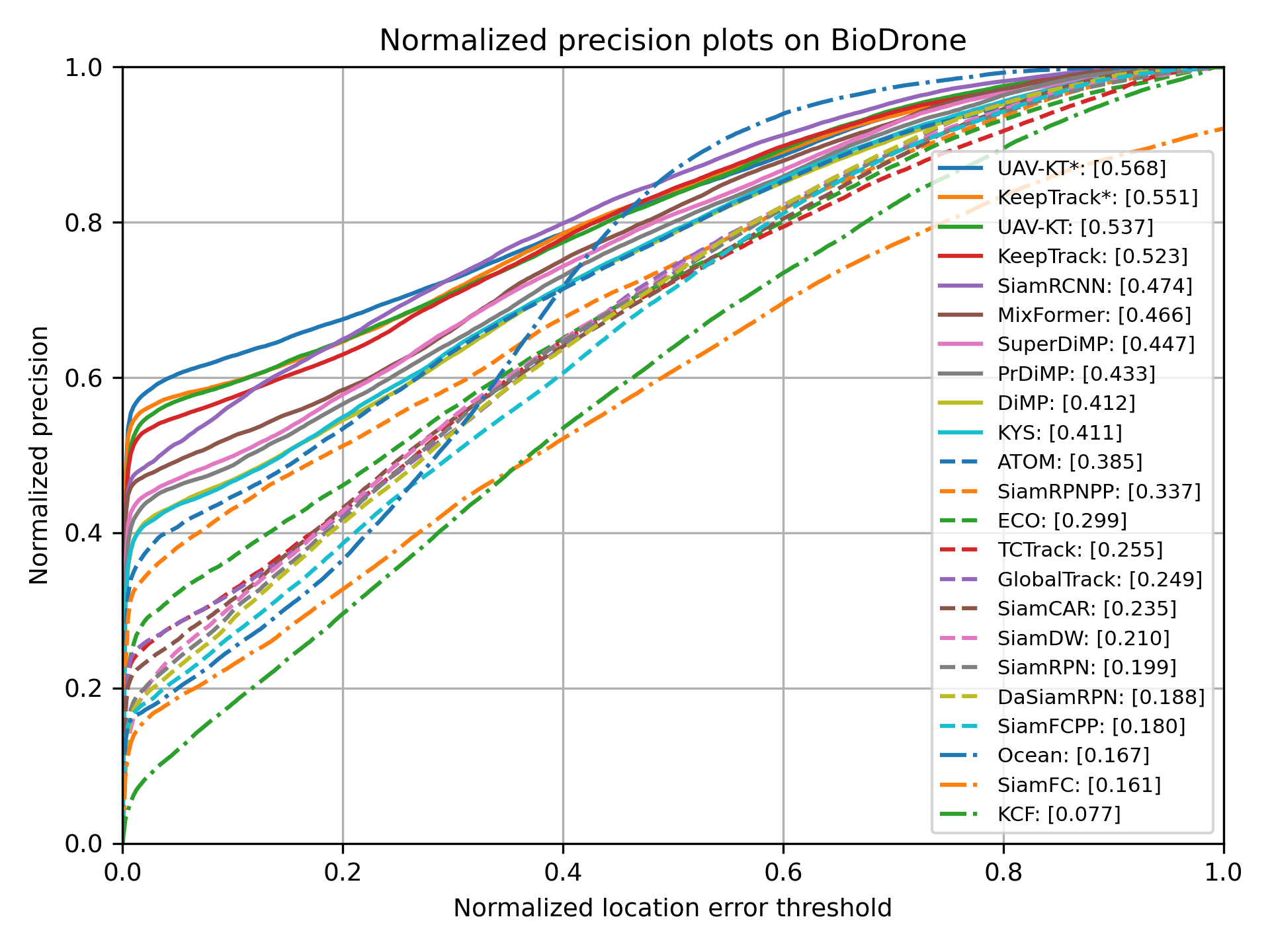}
    }
    \subfigure[Success plot. ]
    {
        \includegraphics[width=\linewidth]{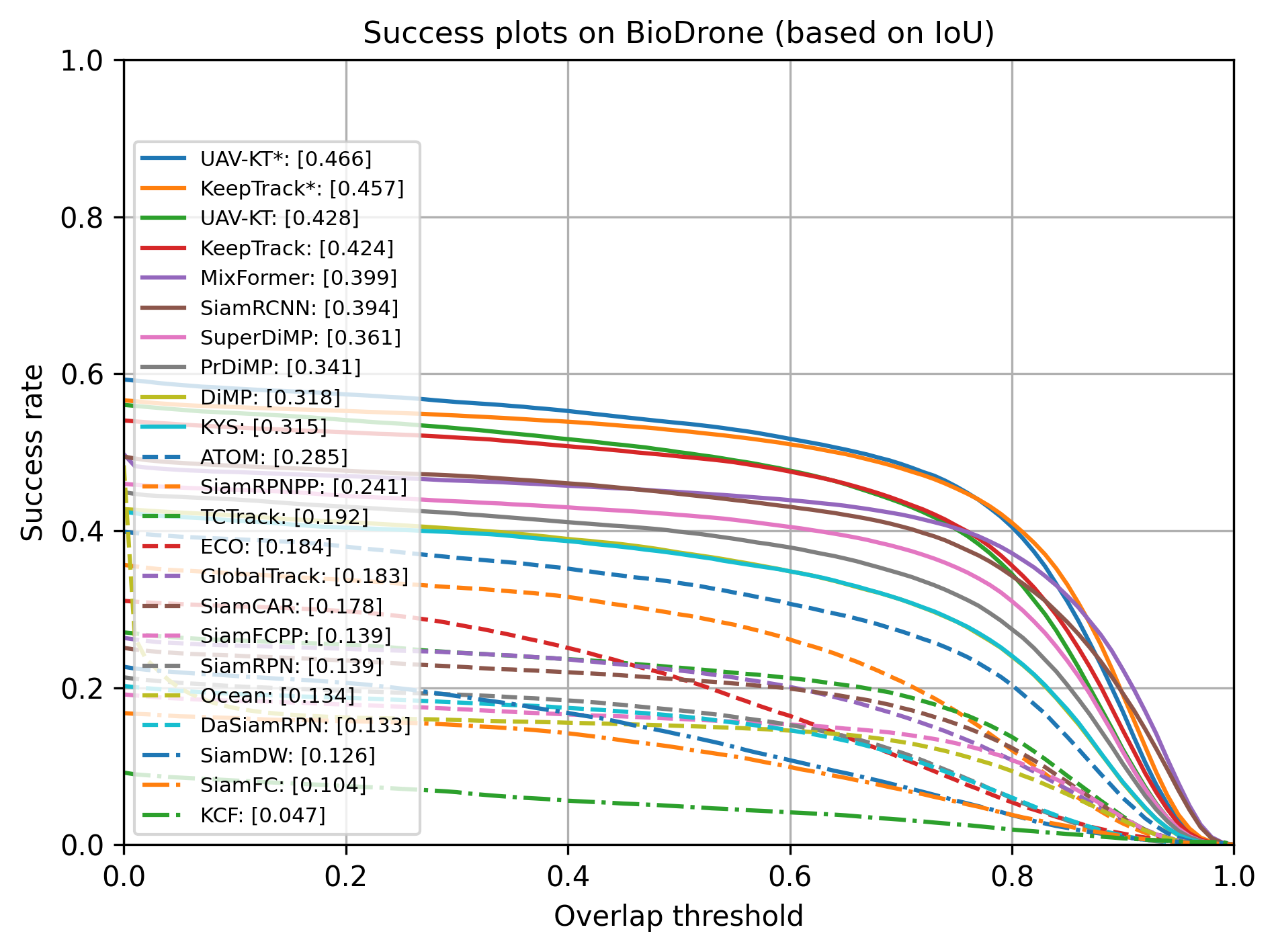}
    }
    \caption{General experiments of BioDrone based on OPE mechanism, evaluated by precision plot (a), normalized precision plot (b), and success plot (c). In brackets, we rank trackers by $P_{score}$, $P^{'}_{score}$, and $S_{score}$.}
    \label{fig:ope-result}
    \end{center}
    \end{figure}

\begin{figure*}[t!]
    \centering
    \includegraphics[width=\linewidth]{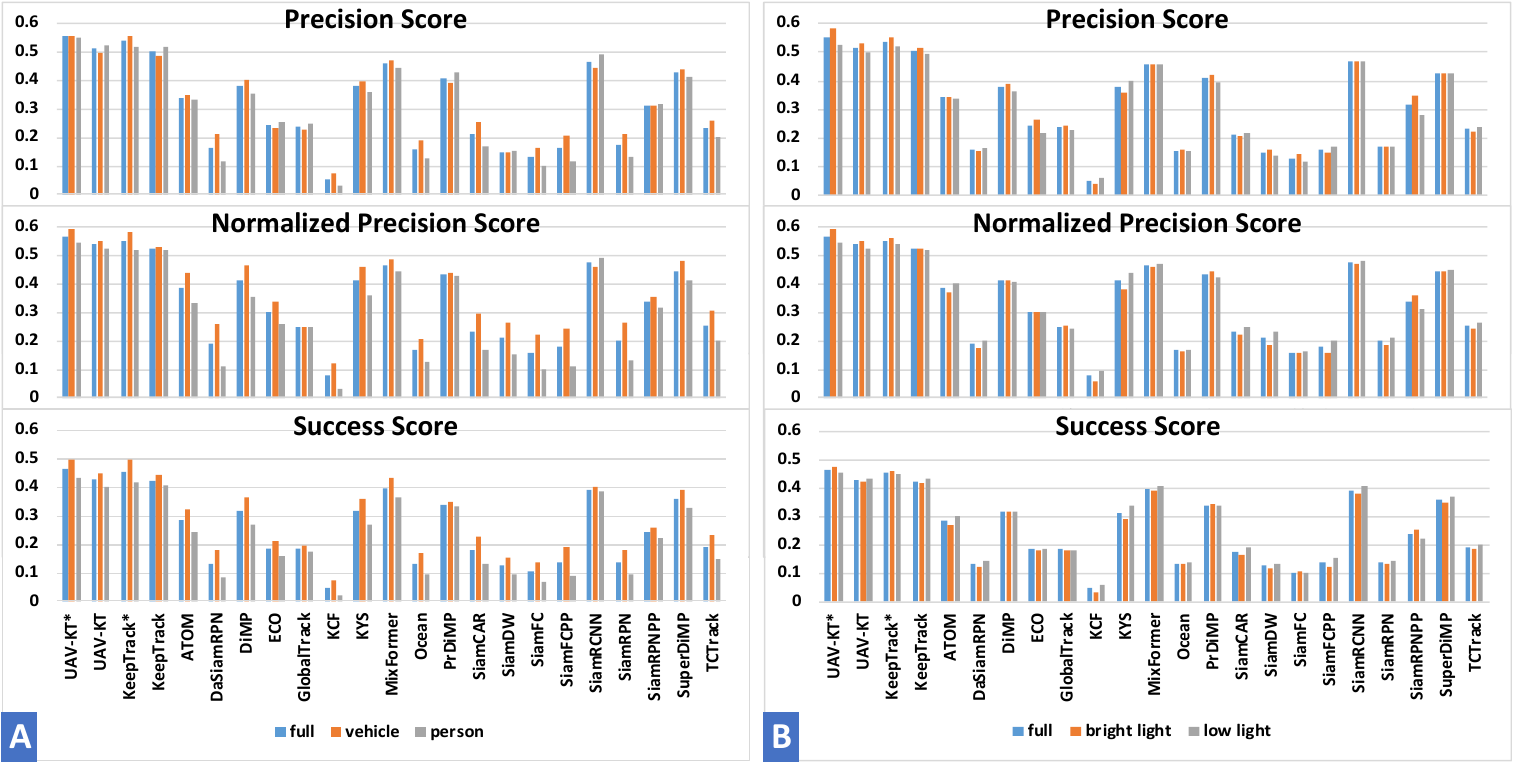}
    \caption
    {
    General experiments of BioDrone based on OPE mechanism, evaluated in different target categories (A) and different light condition (B).
    }
    \label{fig:attribute}
\end{figure*}  

\begin{figure*}[htbp!]
    \begin{center}
    \subfigure[Precision plot.]
    {
        \includegraphics[width=0.47\linewidth]{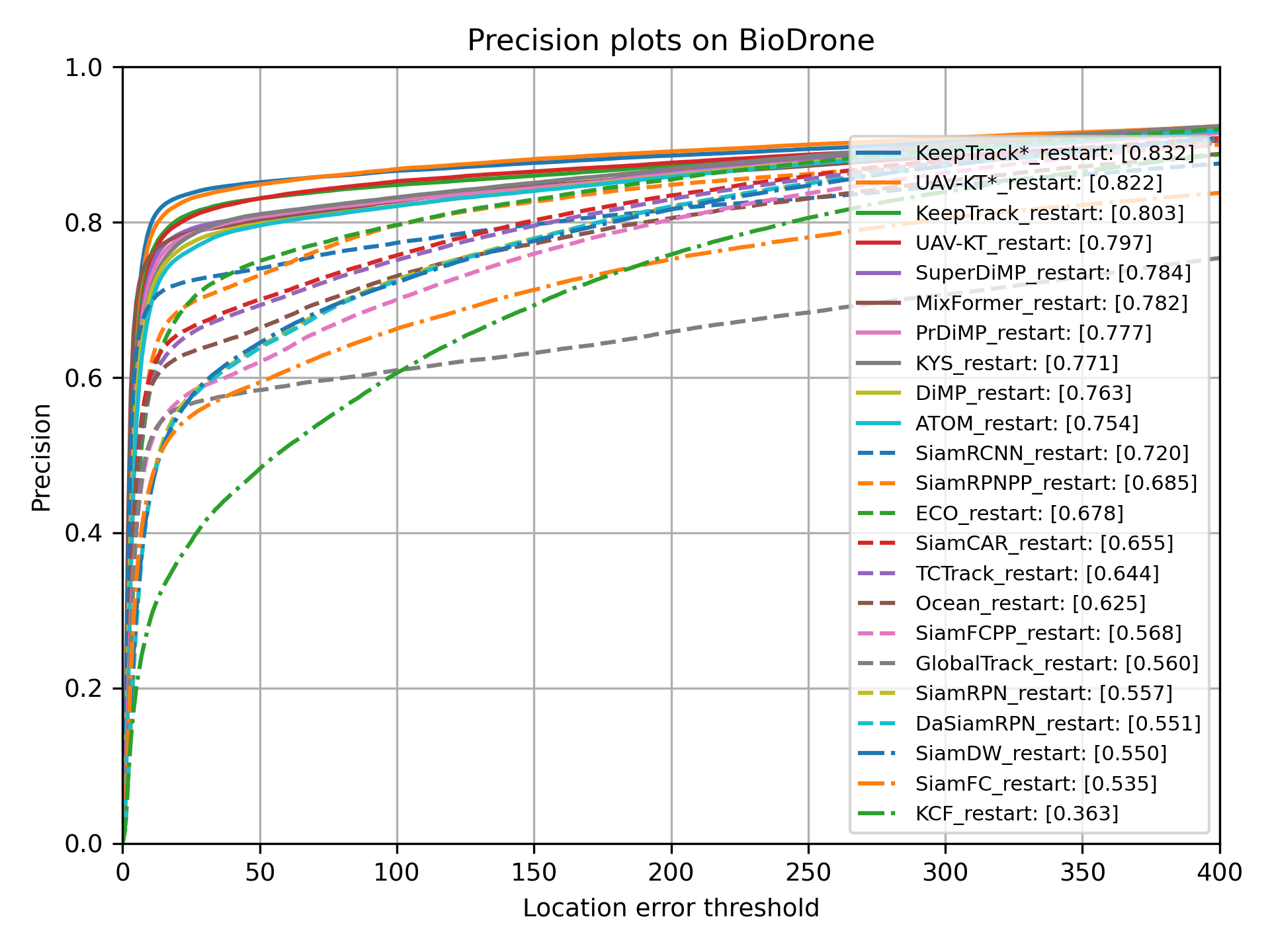}
    }
    \subfigure[Normalized precision plot. ]
    {
        \includegraphics[width=0.47\linewidth]{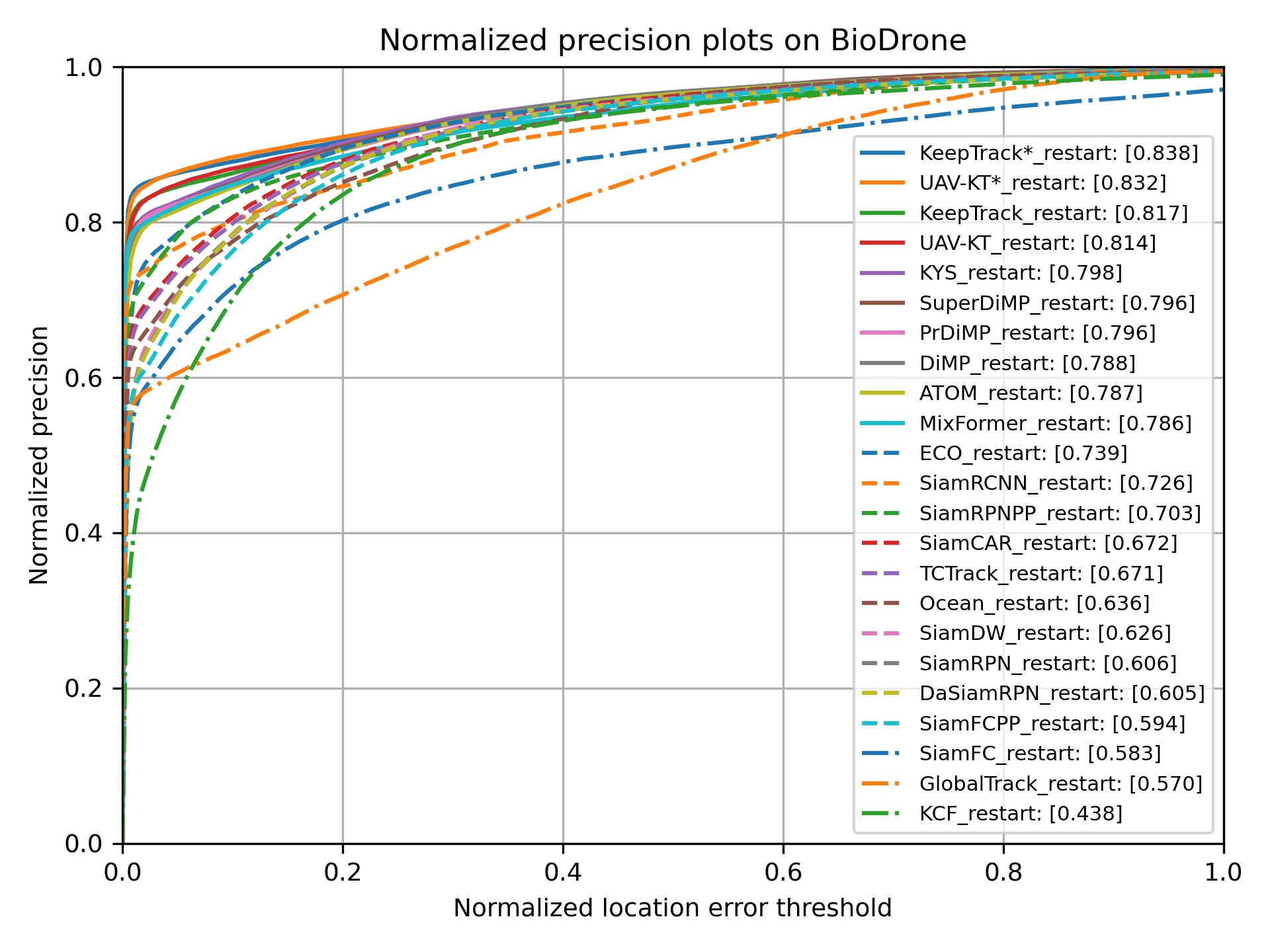}
    }
    \subfigure[Success plot. ]
    {
        \includegraphics[width=0.47\linewidth]{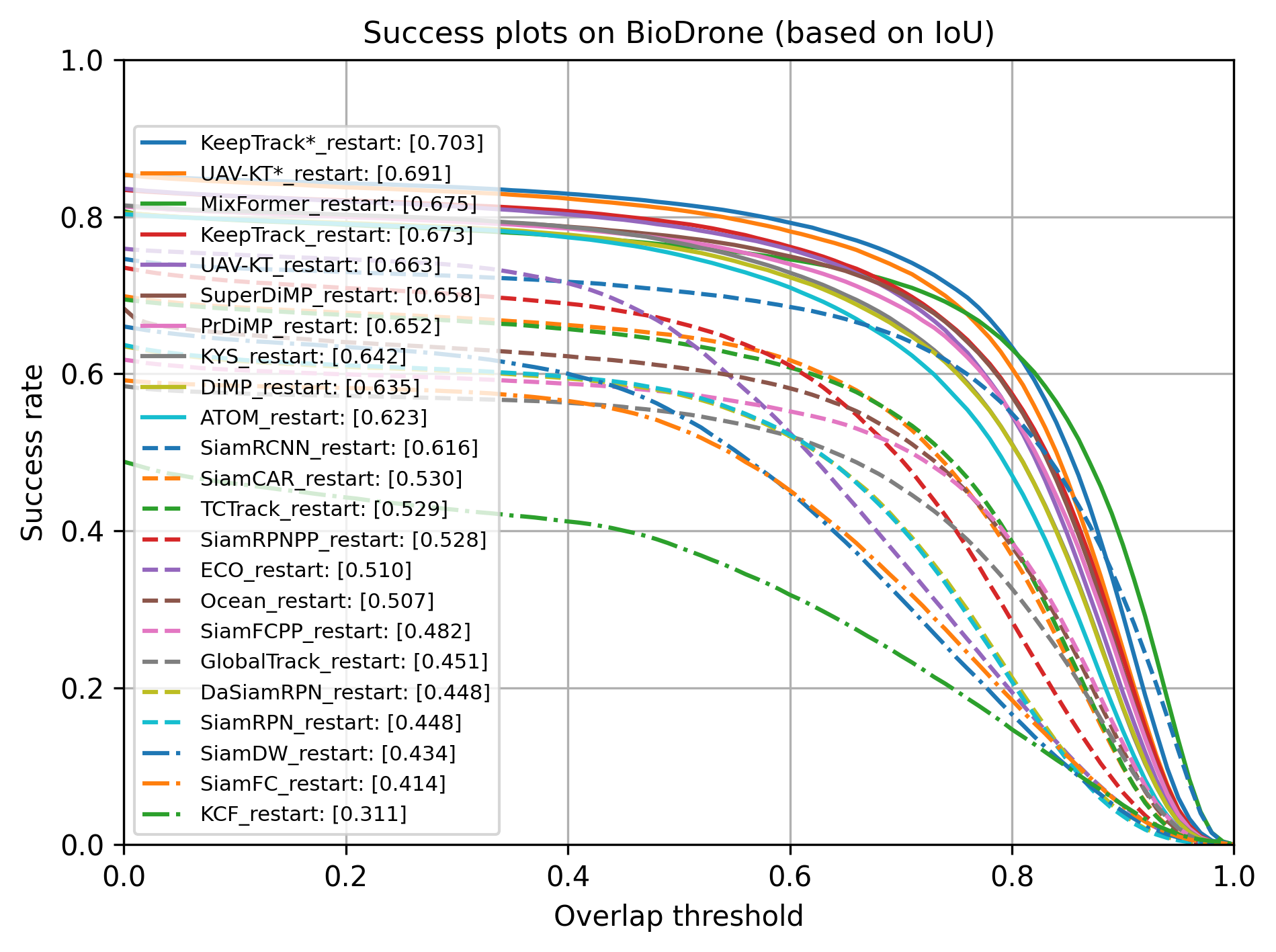}
    }
    \subfigure[Robust tracking plot. ]
    {
        \includegraphics[width=0.47\linewidth]{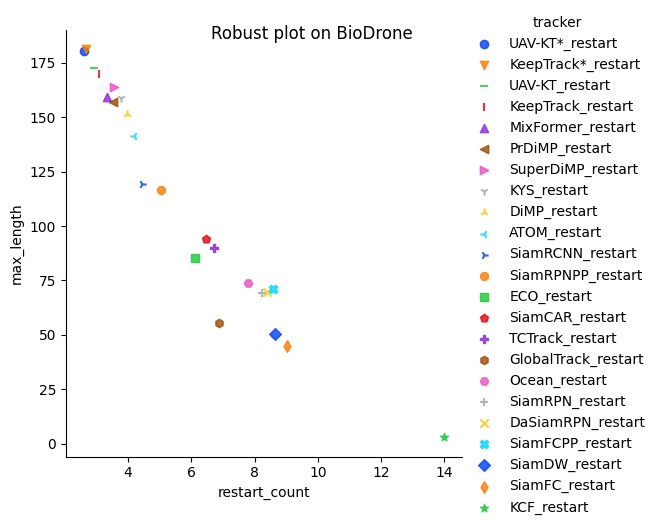}
    }
    \caption{
    General experiments of BioDrone based on the R-OPE mechanism, evaluated by precision plot (a), normalized precision plot (b), and success plot (c). In brackets, we rank trackers by $P_{score}$, $P^{'}_{score}$, and $S_{score}$.
    (d) Besides, BioDrone counts the number of restarts for each video, divides the entire video into several segments based on the restart point, and returns the longest sub-sequence that the algorithm successfully runs. Taking the number of restarts and the mean value of the longest sub-sequence as abscissa and ordinate can generate a robust plot. Trackers closer to the upper left corner perform better (indicating successful tracking in longer sequences with rare re-initializations).
    }
    \label{fig:rope-result}
    \end{center}
    \end{figure*}

\subsection{Performance of Generic SOT Trackers}

We first compare the 20 represent trackers (Section~\ref{subsec:sot-method}) with the proposed baselines (Section~\ref{subsec:new-baselines}) based on OPE and R-OPE evaluation mechanism, as shown in Table~\ref{tab:results}.

For OPE mechanism, precision plot, normalized precision plot, and success plot are selected for evaluation, as shown in Figure~\ref{fig:ope-result}.
Except for the top-4 trackers which are all based on KeepTrack architecture (KeepTrack \cite{KeepTrack} and three proposed new baselines), we note that two other trackers with different model architectures also perform well. 
MixFormer \cite{MixFormer}, a simple end-to-end model based on transformer structure, performs well in all evaluation metrics, indicating that the Mixed Attention Module (MAM) and a straightforward detection head can provide powerful tracking ability. 
Another re-detection-based model SiamRCNN \cite{SiamRCNN} combines a two-stage scheme with a new trajectory-based dynamic planning algorithm and also achieves suitable tracking scores.

We also test trackers on two categories of targets (\ie, vehicles and persons) and three illumination conditions (\ie, bright light, low light (evening), and low light (night)).
We combine low light (evening) and low light (night) into a single category and represented the test results in the above figure. 
In relation to different categories of moving targets (Figure~\ref{fig:attribute} (A)), most algorithms exhibit better tracking performance on vehicles compared to persons. 
One possible explanation is that, from the perspective of a flapping-wing UAV, the size of a person is smaller than that of a vehicle, leading to a reduced number of available visual features and decreased robustness of the trackers.
In various lighting conditions (Figure~\ref{fig:attribute} (B)), most algorithms demonstrate superior tracking performance under bright light compared to low light. 
This indicates that inadequate lighting conditions diminish the visual features of moving targets and present challenges to the robustness of tracking.

Distinguished from the OPE mechanism, the R-OPE mechanism measures robust tracking capability mainly by the number of restarts. 
As shown in Figure~\ref{fig:rope-result} and Table~\ref{tab:results}, all trackers perform better than the original OPE mechanism thanks to the re-initialization. However, all generic SOT trackers need more than 3 times re-initialization in tracking one BioDrone sequence, which means their robust tracking performances are limited in a very short period.

Moreover, we note that the series of methods based on combining CF and SNN (\eg, KeepTrack \cite{KeepTrack}, SuperDiMP \cite{PrDiMP}, PrDiMP \cite{PrDiMP}, DiMP \cite{DiMP}, ATOM \cite{ATOM}) are superior to the SNN-based algorithms (\eg, SiamRPN++ \cite{SiamRPN++}, SiamCAR \cite{SiamCAR}, SiamFC++ \cite{SiamFC++}, DaSiamRPN \cite{DaSiamRPN}, SiamRPN \cite{SiamRPN}, SiamDW \cite{SiamDW}, SiamFC \cite{SiamFC}) of the same period in both OPE and R-OPE mechanisms.
A possible reason is that most SNN-based methods exclude the update mechanism, and highly rely on the integrity of appearance and motion information. The tracking process is executed by matching features between the template region and the search region, while \textit{tiny target} and \textit{fast motion} can decrease the available target information, causing the SNN-based trackers to lose the target easily.
On the contrary, the CF and SNN combination can take advantage of offline training and online updating, helping trackers to suit the appearance variations in the tracking process, and that is why we select the best CF-SNN combination tracker KeepTrack \cite{KeepTrack} as our base model.

\begin{figure*}[t!]
    \centering
    \includegraphics[width=\linewidth]{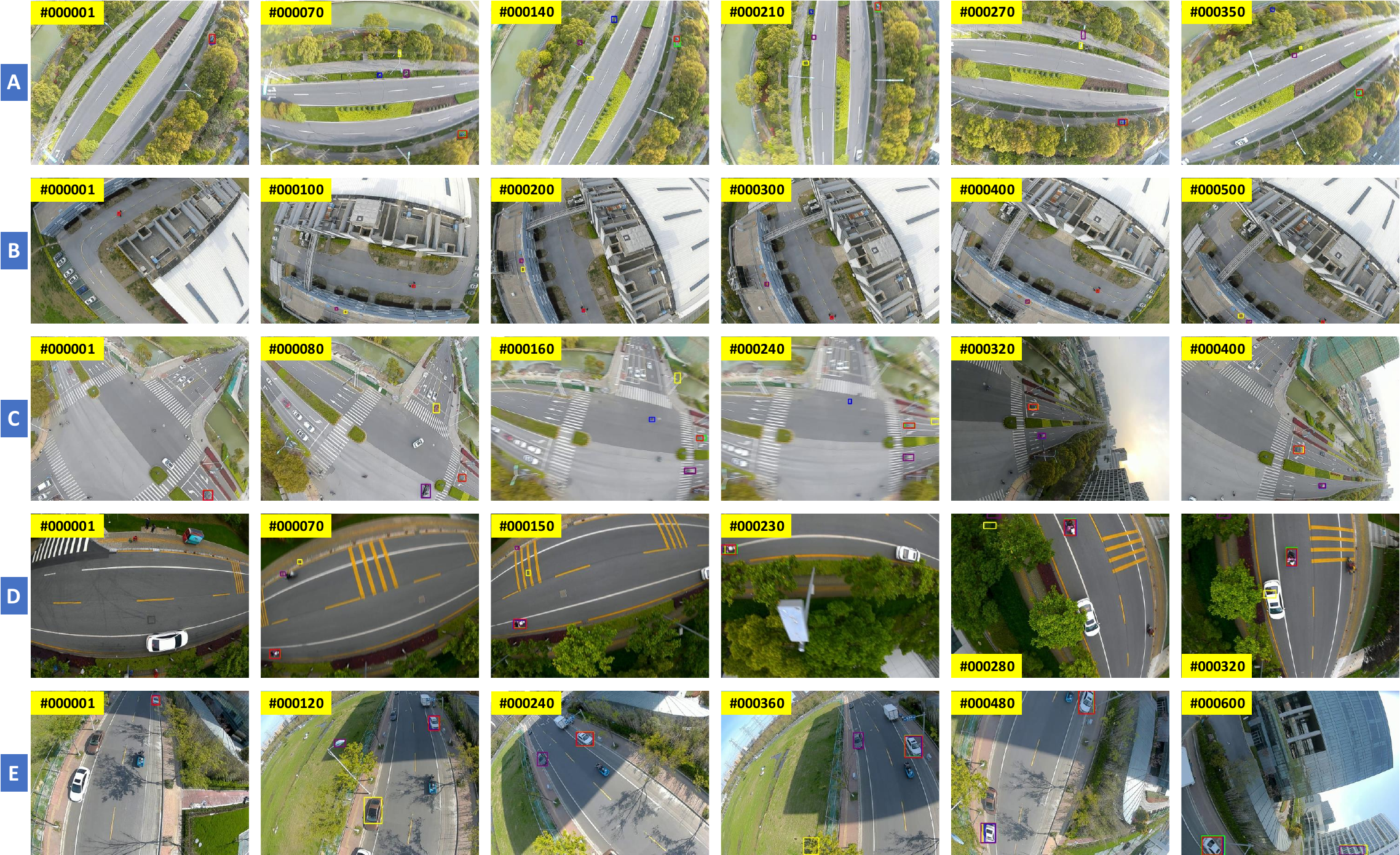}
    \caption{
        Qualitative results of KeepTrack \cite{KeepTrack} and the proposed baselines on BioDrone under the OPE mechanism (\textcolor{green}{$\blacksquare$} green bounding-box represents ground-truth, \textcolor{yellow}{$\blacksquare$} yellow bounding-box  represents KeepTrack \cite{KeepTrack}, \textcolor{blue}{$\blacksquare$} blue bounding-box represents UAV-KT, \textcolor{violet}{$\blacksquare$} violet bounding-box represents KeepTrack*, \textcolor{red}{$\blacksquare$} red bounding-box represents UAV-KT*). 
        Compared to the base model, UAV-KT* performs better when facing challenges in BioDrone.
    }
    \label{fig:better}
    \end{figure*}

\begin{table}[t!]
\caption{Ablation experiments of the proposed new baseline UAV-KT, based on the OPE mechanism.}
\centering
    \subtable[Performance of the new target candidate matching module.]{
      \small
        \begin{tabular}{llll}
        \toprule
        Tracker & \textbf{$P_{score}$$\uparrow$} & \textbf{$P^{'}_{score}$$\uparrow$} & \textbf{$S_{score}$$\uparrow$}  \\\midrule
        KeepTrack \cite{KeepTrack} & 0.504 & 0.523 & 0.424 \\ \midrule
        UAV-KT & \tabincell{l}{0.513 \\ (0.009 $\uparrow$)} & \tabincell{l}{0.537 \\ (0.014 $\uparrow$)} & \tabincell{l}{0.428 \\ (0.004 $\uparrow$)} \\\botrule
        \end{tabular}
        }
    \subtable[Performance of different training strategies.]{        
    \begin{tabular}{llll}
        \toprule
        Tracker & \textbf{$P_{score}$$\uparrow$} & \textbf{$P^{'}_{score}$$\uparrow$} & \textbf{$S_{score}$$\uparrow$}  \\\midrule
        KeepTrack \cite{KeepTrack} & 0.504 & 0.523 & 0.424 \\ \midrule
        KeepTrack* & \tabincell{l}{0.538 \\ (0.034 $\uparrow$)} & \tabincell{l}{0.551 \\ (0.028 $\uparrow$)} & \tabincell{l}{0.457 \\ (0.033 $\uparrow$)} \\
        KeepTrack\# & \tabincell{l}{0.496 \\ (0.008 $\downarrow$)} & \tabincell{l}{0.520 \\ (0.003 $\downarrow$)} & \tabincell{l}{0.417 \\ (0.007 $\downarrow$)} \\\botrule
        \end{tabular}
    }
    \subtable[Performance of the combination results.]{        
    \begin{tabular}{llll}
        \toprule
        Tracker & \textbf{$P_{score}$$\uparrow$} & \textbf{$P^{'}_{score}$$\uparrow$} & \textbf{$S_{score}$$\uparrow$}  \\\midrule
        KeepTrack \cite{KeepTrack} & 0.504 & 0.523 & 0.424 \\ \midrule
        KeepTrack* & \tabincell{l}{0.538 \\ (0.034 $\uparrow$)} & \tabincell{l}{0.551 \\ (0.028 $\uparrow$)} & \tabincell{l}{0.457 \\ (0.033 $\uparrow$)} \\
        UAV-KT* & \tabincell{l}{0.554 \\ (0.050 $\uparrow$)} & \tabincell{l}{0.568 \\ (0.045 $\uparrow$)} & \tabincell{l}{0.466 \\ (0.042 $\uparrow$)} \\\botrule
        \end{tabular}
    }
    \label{tab:ablation-exp}
\end{table}

\subsection{Performance of the Proposed Baselines}

Obviously, UAV-KT* and KeepTrack*, the two trackers which have been re-trained on the BioDrone benchmark, achieve the best two performances in both OPE (Figure~\ref{fig:ope-result}) and R-OPE mechanisms (Figure~\ref{fig:rope-result}). 
For all trackers that have not been re-trained on BioDrone (we use the parameters and confirmations provided by the original authors), the proposed new baseline UAV-KT performs well.
Here we design several ablation experiments to better exhibit the performance of the proposed new baseline UAV-KT and the training strategies.

\subsubsection{Target Candidate Matching Network}

The proposed UAV-KT utilizes some shallow features, which is especially effective for tiny targets, to obtain more meaningful features at the candidate embedding module. The score matrices are summed through the learned weights by the candidate matching module. 
Here, the weights are finally learned as $[0.4929,0.5070]$, in which the former is the shallow score matrix summing coefficient.
Table~\ref{tab:ablation-exp} (a) illustrates the performance of the original KeepTrack \cite{KeepTrack} and the proposed UAV-KT. Note that neither of the two trackers is re-trained on BioDrone. Obviously, based on the target candidate matching network, UAV-KT improves its robustness by perceiving targets of different scales.

\begin{figure*}[htbp!]
    \begin{center}
    \subfigure[Performance in tracking tiny target (smaller value in horizontal coordinate means including more tiny targets).]
    {
        \includegraphics[width=0.9\linewidth]{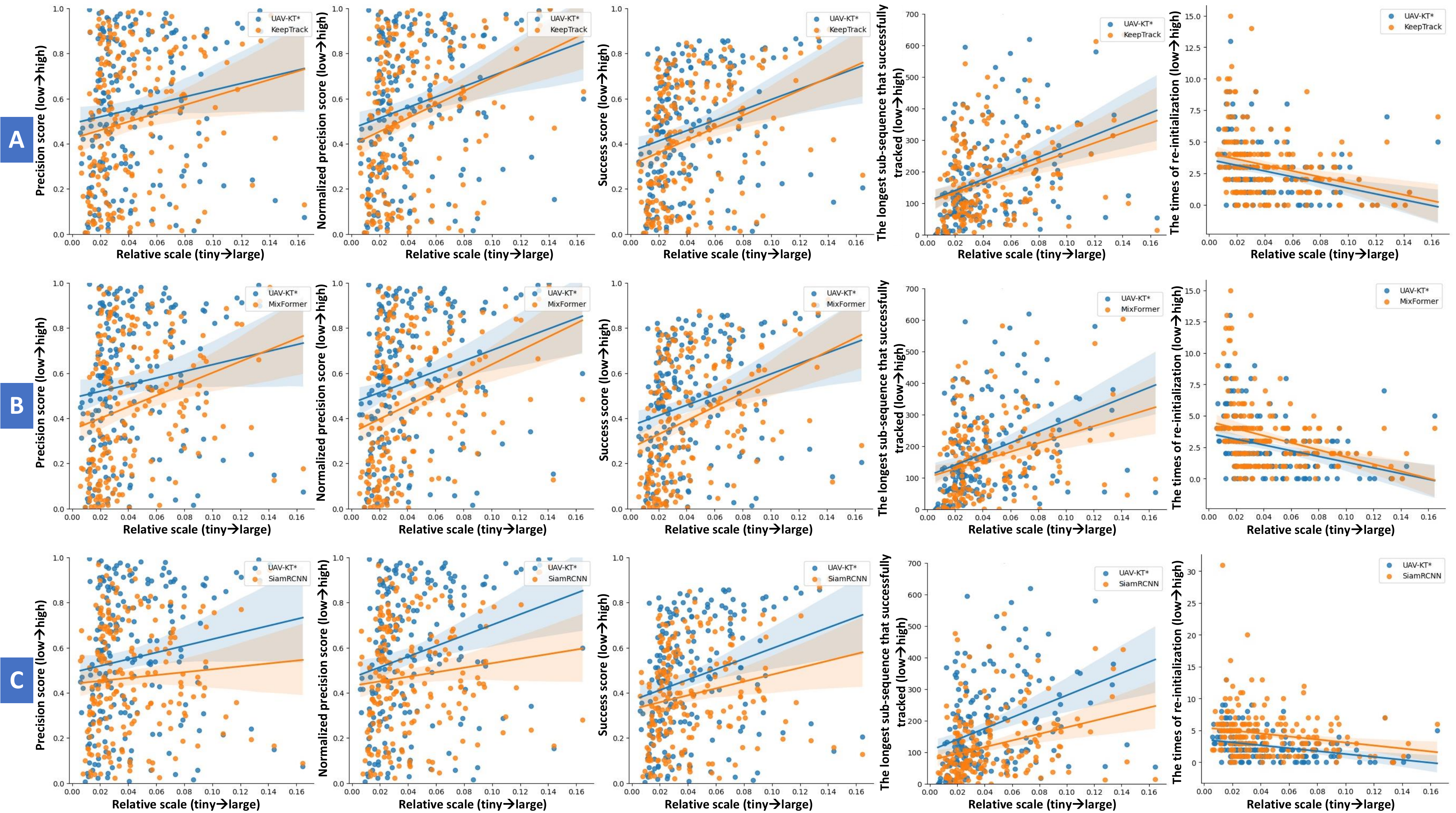}
    }
    \subfigure[Performance in tracking fast motion target (larger value in horizontal coordinate means including faster motion).]
    {
        \includegraphics[width=0.9\linewidth]{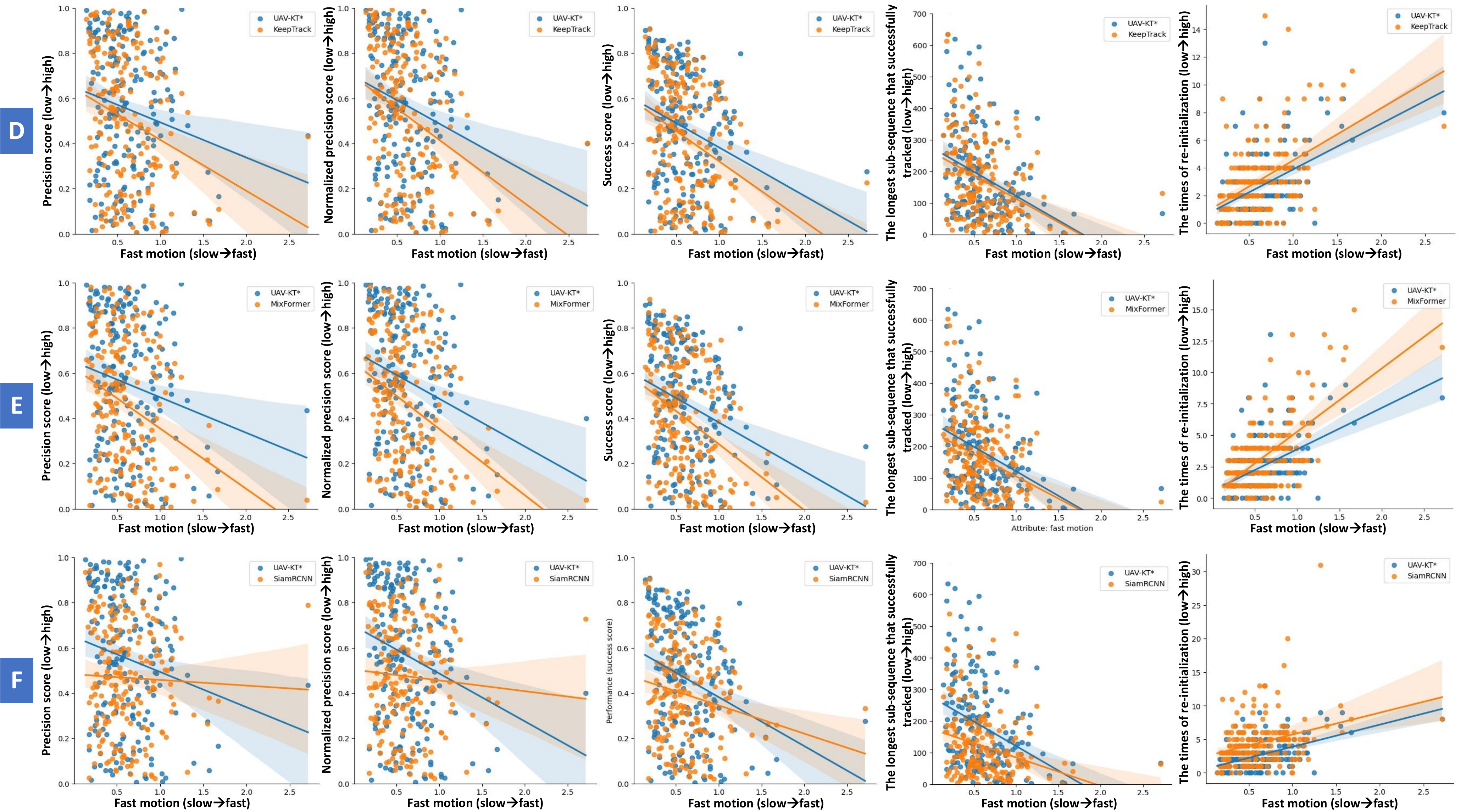}
    }
    \caption{
    Performance of the proposed UAV-KT* and represent generic SOT methods on challenging attributes.
    The scores of each algorithm in the test set (200 videos) are plotted as scatter plots. Where the vertical coordinates represent the scores of the algorithms (from left to right: precision score $P_{score}$, normalized precision score $P^{'}_{score}$, and success score $S_{score}$ in OPE mechanism; the average value of the longest sub-sequence $L_{max}$ and the number of restarts $R_{count}$ in R-OPE mechanism).
    The horizontal coordinates of (a) represent the average relative target scale, and the horizontal coordinates of (b) represent the average target motion in a video.
    Clearly, UAV-KT* performs better than KeepTrack \cite{KeepTrack}, MixFormer \cite{MixFormer}, and SiamRCNN \cite{SiamRCNN} in both tiny target and fast motion challenges.
    }
    \label{fig:challenge-result}
    \end{center}
    \end{figure*}

\begin{figure*}[htbp!]
    \centering
    \includegraphics[width=\linewidth]{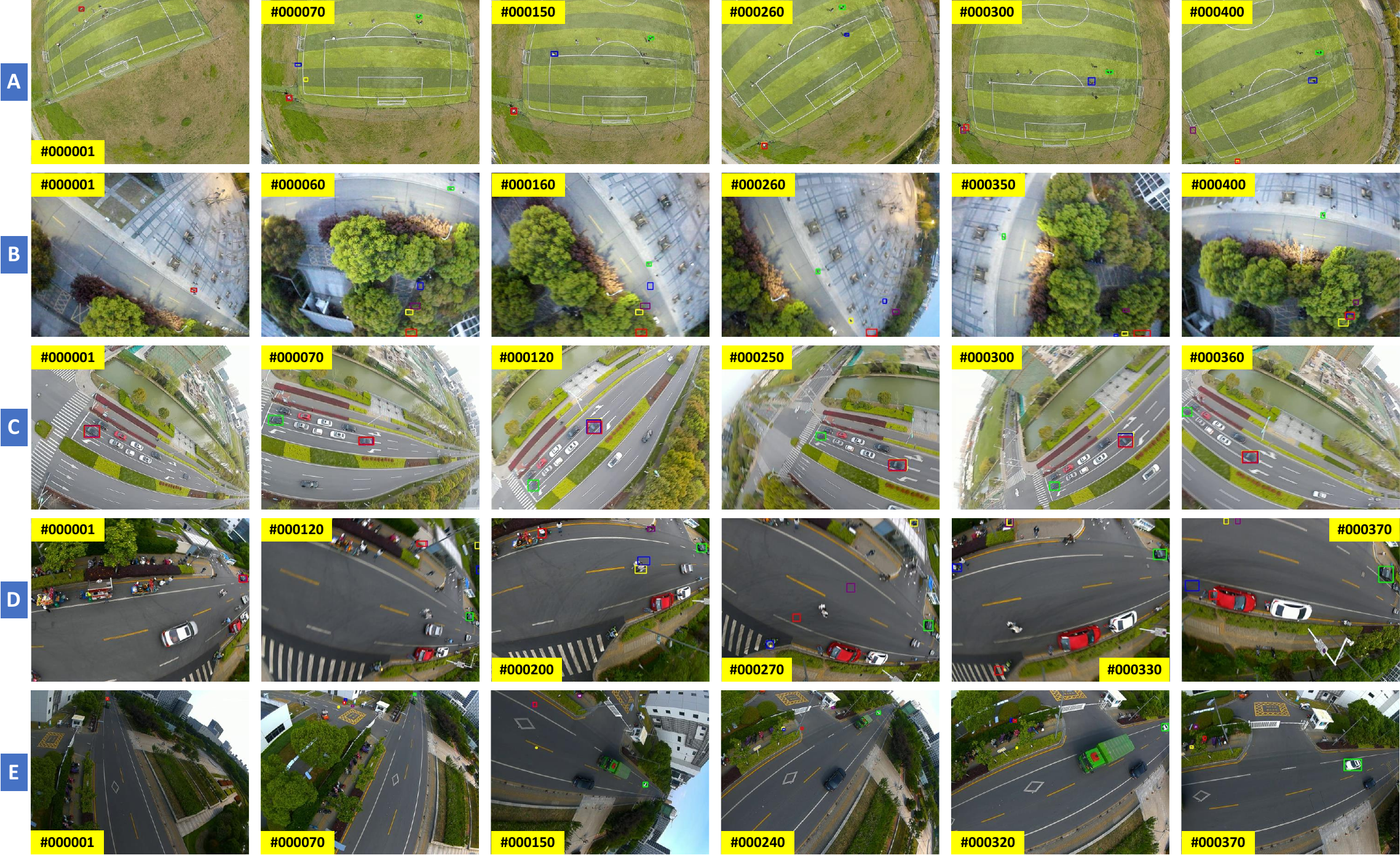}
    \caption{
        Qualitative results of some bad cases for the represent trackers on OPE mechanism (\textcolor{green}{$\blacksquare$} green bounding-box represents ground-truth, \textcolor{yellow}{$\blacksquare$} yellow bounding-box  represents KeepTrack \cite{KeepTrack}, \textcolor{blue}{$\blacksquare$} blue bounding-box represents UAV-KT, \textcolor{violet}{$\blacksquare$} violet bounding-box represents KeepTrack*, \textcolor{red}{$\blacksquare$} red bounding-box represents UAV-KT*). 
    }
    \label{fig:bad-case}
    \end{figure*}

\subsubsection{Different Training Strategies}
\label{subsubsection:training-strategies}

As shown in Figure~\ref{fig:UAV-KT}, the original KeepTrack and UAV-KT include several parts (\ie, the base tracker, the target candidate extraction, and the target candidate association network). We notice that end-to-end training is not an appropriate strategy. Thus, to find a better training method, We design several strategies to explore the optimal parameters.

\begin{itemize}
\item \textbf{Strategy-1.} re-train on the base tracker (KeepTrack*).
\item \textbf{Strategy-2.} Train target candidate association network with data from LaSOT and BioDrone training sets that meet the candidate conditions (KeepTrack\#).
\end{itemize}

Table~\ref{tab:ablation-exp} (b) shows that using BioDrone to re-train the base tracker  improves the performance of KeepTrack (KeepTrack*), while the candidate association network performs poorly after re-training by the supplementary dataset (KeepTrack\#).

We believe that this difference occurs because the two modules are designed for different tasks.
(1) The task of the base tracker (SuperDiMP in KeepTrack) is target classification. A discriminative target predictor weight is obtained from template features, then it performs a cross-correlation operation with the frame features to be detected, and finally a score map is obtained. 
(2) Target candidate association network uses the score map from the base tracker to select target candidates, then extracts target candidate features for target candidate matching to finally identify the target.

Thus, using Strategy-1 for the base tracker can effectively improve the model's discriminative ability between forward and backward information, making it locate the target more robustly.

On the contrary, when Strategy-2 is applied to the target candidate association network, we first run the base tracker on all sequences of the BioDrone train-set to obtain tracking results, and then set the train-set into two parts: a \textit{train-train} and a \textit{train-val} set.  
These datasets contain several tracking situations: (1) The correct candidate object is selected as the target. (2) It is no longer possible to track the target because the target classifier score of the corresponding candidate object is below a threshold. (3) Tracking fails, which includes the correct target existing but not selected or there is no correct target and none of the target candidates is selected. 
The task of the target candidate association module includes learning how to distinguish between the target with distractors, and how to remediate wrong results when the base tracker fails. 
However, due to the tiny target challenge, the appearance information on targets and distractors in BioDrone is not obvious. Thus, this training strategy may cause even a negative impact on trackers (please refer to the worse performance of KeepTrack\# in Table~\ref{tab:ablation-exp} (b)). 

Based on the above analyses, Strategy-1 is selected as the final training strategy.

\subsubsection{Results of Our New Baseline}

Our new baseline UAV-KT* employs the proposed target candidate association module and the training Strategy-1 based on the BioDrone. Table~\ref{tab:ablation-exp} (c) illustrates that the combination improves the tracking performance effectively, which provides a novel  direction for the following research.

\subsection{Performance on Challenging Attributes}

Different from tracking the target in generic scenarios, the UAV-based SOT task requires more visual robustness. In this section, we compare the proposed UAV-DT* baseline and three SOTA methods in challenging situations, to further analyze their robustness.
Figure~\ref{fig:challenge-result} illustrates the performance of trackers in tracking \textit{tiny target} with \textit{fast motion}. The above two factors reduce the available appearance information and abrupt the trajectories, causing trackers to fail easily. 

Although SOTA methods like KeepTrack  \cite{KeepTrack}, MixFormer \cite{MixFormer}, and SiamRCNN \cite{SiamRCNN} perform well in generic situations (Figure~\ref{fig:motivation}), they are easily failed in facing \textit{tiny target}. 
Figure~\ref{fig:challenge-result} (a) shows that with the decrease in target size, performances of all trackers based on different mechanisms and metrics all drop quickly.
For example, SiamRCNN \cite{SiamRCNN} even fails more than 30 times in a sequence (the rightmost sub-figure in Figure~\ref{fig:challenge-result} (c)), which shows that it is completely unable to handle this task, regardless of what strategies it has enabled.
This phenomenon can also be observed in \textit{fast motion} situation. As exhibited in Figure~\ref{fig:challenge-result} (b), the faster motion in two continuous frames, the poorer performance that trackers have. 

Thus, the BioDrone benchmark introduces new challenging factors in the visual object tracking task and provides a comprehensive experimental environment for robust vision.
Although existing methods perform poorly on this dataset, the proposed UAV-KT* gives a preliminary solution by optimizing the model structure and training strategies. 
However, some bad cases presented in Figure~\ref{fig:bad-case} demonstrate that our base can be further improved, and multiple robust vision problems on BioDrone still deserve further research.
The challenges brought by the \textit{tiny target} and \textit{fast motion} are highlighted in these examples. In contrast to tracking tasks in general scenes, pedestrians and vehicles appear significantly smaller in the drone's field of view. Additionally, the shaking and rotation of the camera during flapping flight can disturb the motion trajectory of the target, thus presenting significant challenges for algorithms that depend on visual features and motion information.

\section{Conclusion}
In this paper, a bionic drone-based single object tracking benchmark BioDrone is proposed for robust vision research.
Unlike existing benchmarks that are mainly based on fixed-wing or rotary-wing UAVs, the flapping-wing system selected by BioDrone includes additional visual challenges due to its serious camera shake. Compared with existing works, BioDrone is the largest UAV-based SOT benchmark with a \textit{smaller target size} and \textit{more drastic appearance changes} between consecutive frames. It includes \textit{600} videos with \textit{304,209} manually labeled frames, and automatically generates frame-level labels for ten challenge attributes, which provides a high-quality and challenging experimental environment for robust vision research.
Besides, We further optimize the SOTA method KeepTrack \cite{KeepTrack} and design a new baseline UAV-KT with a suitable training strategy, aiming to propose a preliminary baseline for challenging factors in BioDrone.  
Finally, we test our method and 20 representative methods by comprehensive evaluation mechanisms and metrics in BioDrone, and experimental results indicate that the proposed method achieves 5\% performance boost in the precision score.
However, several failure cases and systematic analyses indicate that BioDrone still contains many unresolved challenges and deserves further attention in robust vision research.

In the future, we believe that the proposed BioDrone benchmark can provide a high-quality experimental environment for further research, and help researchers to design new robust tracking methods.
Besides, this work also represents a broader range of SOT problems, such as those in high-speed autonomous driving, and egocentric vision. While BioDrone mainly focuses on bionic UAVs, the results and findings in this paper might transfer to those more comprehensive problems. 

\section*{Declarations}

\begin{itemize}
% \item \textbf{Funding.}
\item \textbf{Conflict of Interest.} All authors declare no conflicts of interest.
% \item \textbf{Ethics approval.} 
% \item \textbf{Consent to participate.}
% \item \textbf{Consent for publication.}
\item \textbf{Availability of data and materials.} All data will be made available on reasonable request.
\item \textbf{Code availability.} The toolkit and experimental results will be made publicly available. 
% \item \textbf{Authors' contributions.}
\end{itemize}

% \begin{appendices}
% \section{Section title of first appendix}\label{secA1}
% \end{appendices}

\bibliography{sn-bibliography}


\begin{thebibliography}{84}
\ifx \bisbn   \undefined \def \bisbn  #1{ISBN #1}\fi
\ifx \binits  \undefined \def \binits#1{#1}\fi
\ifx \bauthor  \undefined \def \bauthor#1{#1}\fi
\ifx \batitle  \undefined \def \batitle#1{#1}\fi
\ifx \bjtitle  \undefined \def \bjtitle#1{#1}\fi
\ifx \bvolume  \undefined \def \bvolume#1{\textbf{#1}}\fi
\ifx \byear  \undefined \def \byear#1{#1}\fi
\ifx \bissue  \undefined \def \bissue#1{#1}\fi
\ifx \bfpage  \undefined \def \bfpage#1{#1}\fi
\ifx \blpage  \undefined \def \blpage #1{#1}\fi
\ifx \burl  \undefined \def \burl#1{\textsf{#1}}\fi
\ifx \doiurl  \undefined \def \doiurl#1{\url{https://doi.org/#1}}\fi
\ifx \betal  \undefined \def \betal{\textit{et al.}}\fi
\ifx \binstitute  \undefined \def \binstitute#1{#1}\fi
\ifx \binstitutionaled  \undefined \def \binstitutionaled#1{#1}\fi
\ifx \bctitle  \undefined \def \bctitle#1{#1}\fi
\ifx \beditor  \undefined \def \beditor#1{#1}\fi
\ifx \bpublisher  \undefined \def \bpublisher#1{#1}\fi
\ifx \bbtitle  \undefined \def \bbtitle#1{#1}\fi
\ifx \bedition  \undefined \def \bedition#1{#1}\fi
\ifx \bseriesno  \undefined \def \bseriesno#1{#1}\fi
\ifx \blocation  \undefined \def \blocation#1{#1}\fi
\ifx \bsertitle  \undefined \def \bsertitle#1{#1}\fi
\ifx \bsnm \undefined \def \bsnm#1{#1}\fi
\ifx \bsuffix \undefined \def \bsuffix#1{#1}\fi
\ifx \bparticle \undefined \def \bparticle#1{#1}\fi
\ifx \barticle \undefined \def \barticle#1{#1}\fi
\bibcommenthead
\ifx \bconfdate \undefined \def \bconfdate #1{#1}\fi
\ifx \botherref \undefined \def \botherref #1{#1}\fi
\ifx \url \undefined \def \url#1{\textsf{#1}}\fi
\ifx \bchapter \undefined \def \bchapter#1{#1}\fi
\ifx \bbook \undefined \def \bbook#1{#1}\fi
\ifx \bcomment \undefined \def \bcomment#1{#1}\fi
\ifx \oauthor \undefined \def \oauthor#1{#1}\fi
\ifx \citeauthoryear \undefined \def \citeauthoryear#1{#1}\fi
\ifx \endbibitem  \undefined \def \endbibitem {}\fi
\ifx \bconflocation  \undefined \def \bconflocation#1{#1}\fi
\ifx \arxivurl  \undefined \def \arxivurl#1{\textsf{#1}}\fi
\csname PreBibitemsHook\endcsname

%%% 1
\bibitem{KeepTrack}
\begin{bchapter}
\bauthor{\bsnm{Mayer}, \binits{C.}},
\bauthor{\bsnm{Danelljan}, \binits{M.}},
\bauthor{\bsnm{Paudel}, \binits{D.P.}},
\bauthor{\bsnm{Van~Gool}, \binits{L.}}:
\bctitle{Learning target candidate association to keep track of what not to track}.
In: \bbtitle{Proceedings of the IEEE/CVF International Conference on Computer Vision},
pp. \bfpage{13444}--\blpage{13454}
(\byear{2021})
\end{bchapter}
\endbibitem

%%% 2
\bibitem{VOT2018}
\begin{bchapter}
\bauthor{\bsnm{Kristan}, \binits{M.}},
\bauthor{\bsnm{Matas}, \binits{J.}},
\bauthor{\bsnm{Leonardis}, \binits{A.}},
\bauthor{\bsnm{Felsberg}, \binits{M.}},
\bauthor{\bsnm{Pflugfelder}, \binits{R.}},
\bauthor{\bsnm{K{\"a}m{\"a}r{\"a}inen}, \binits{J.-K.}},
\bauthor{\bsnm{Cehovin~Zajc}, \binits{L.}},
\bauthor{\bsnm{Drbohlav}, \binits{O.}},
\bauthor{\bsnm{Lukezic}, \binits{A.}},
\bauthor{\bsnm{Berg}, \binits{A.}},
\bauthor{\bsnm{Eldesokey}, \binits{A.}},
\bauthor{\bsnm{K{\"a}pyl{\"a}}, \binits{J.}},
\bauthor{\bsnm{Fern{\'a}ndez}, \binits{G.}},
\bauthor{\bsnm{{Gonzalez-Garcia}}, \binits{A.}},
\bauthor{\bsnm{Memarmoghadam}, \binits{A.}},
\bauthor{\bsnm{others.}}:
\bctitle{The {{Seventh Visual Object Tracking VOT2019 Challenge Results}}}.
In: \bbtitle{Proceedings of 2019 {{IEEE}}/{{CVF International Conference}} on {{Computer Vision Workshop}} ({{ICCVW}})},
pp. \bfpage{2206}--\blpage{2241}.
\bpublisher{{IEEE}},
\blocation{{Seoul, Korea (South)}}
(\byear{2019})
\end{bchapter}
\endbibitem

%%% 3
\bibitem{VOT2019}
\begin{bchapter}
\bauthor{\bsnm{Kristan}, \binits{M.}},
\bauthor{\bsnm{Leonardis}, \binits{A.}},
\bauthor{\bsnm{Matas}, \binits{J.}},
\bauthor{\bsnm{Felsberg}, \binits{M.}},
\bauthor{\bsnm{Pflugfelder}, \binits{R.}},
\bauthor{\bsnm{Zajc}, \binits{L.{\v C}.}},
\bauthor{\bsnm{Voj{\'i}{\~r}}, \binits{T.}},
\bauthor{\bsnm{Bhat}, \binits{G.}},
\bauthor{\bsnm{Luke{\v z}i{\v c}}, \binits{A.}},
\bauthor{\bsnm{Eldesokey}, \binits{A.}},
\bauthor{\bsnm{Fern{\'a}ndez}, \binits{G.}},
\bauthor{\bsnm{others.}}:
\bctitle{The {{Sixth Visual Object Tracking VOT2018 Challenge Results}}}.
In: \bbtitle{Computer {{Vision}} \textendash{} {{ECCV}} 2018 {{Workshops}}},
pp. \bfpage{3}--\blpage{53}.
\bpublisher{{Springer}},
\blocation{{Munich, Germany}}
(\byear{2019})
\end{bchapter}
\endbibitem

%%% 4
\bibitem{LaSOT}
\begin{barticle}
\bauthor{\bsnm{Fan}, \binits{H.}},
\bauthor{\bsnm{Bai}, \binits{H.}},
\bauthor{\bsnm{Lin}, \binits{L.}},
\bauthor{\bsnm{Yang}, \binits{F.}},
\bauthor{\bsnm{Chu}, \binits{P.}},
\bauthor{\bsnm{Deng}, \binits{G.}},
\bauthor{\bsnm{Yu}, \binits{S.}},
\bauthor{\bsnm{Huang}, \binits{M.}},
\bauthor{\bsnm{Liu}, \binits{J.}},
\bauthor{\bsnm{Xu}, \binits{Y.}}, \betal:
\batitle{Lasot: A high-quality large-scale single object tracking benchmark}.
\bjtitle{International Journal of Computer Vision}
\bvolume{129}(\bissue{2}),
\bfpage{439}--\blpage{461}
(\byear{2021})
\end{barticle}
\endbibitem

%%% 5
\bibitem{GIT}
\begin{barticle}
\bauthor{\bsnm{Hu}, \binits{S.}},
\bauthor{\bsnm{Zhao}, \binits{X.}},
\bauthor{\bsnm{Huang}, \binits{L.}},
\bauthor{\bsnm{Huang}, \binits{K.}}:
\batitle{Global {{Instance Tracking}}: {{Locating Target More Like Humans}}}.
\bjtitle{IEEE Transactions on Pattern Analysis and Machine Intelligence}
\bvolume{45}(\bissue{1}),
\bfpage{576}--\blpage{592}
(\byear{2023})
\end{barticle}
\endbibitem

%%% 6
\bibitem{MixFormer}
\begin{bchapter}
\bauthor{\bsnm{Cui}, \binits{Y.}},
\bauthor{\bsnm{Jiang}, \binits{C.}},
\bauthor{\bsnm{Wang}, \binits{L.}},
\bauthor{\bsnm{Wu}, \binits{G.}}:
\bctitle{Mixformer: End-to-end tracking with iterative mixed attention}.
In: \bbtitle{Proceedings of the IEEE/CVF Conference on Computer Vision and Pattern Recognition},
pp. \bfpage{13608}--\blpage{13618}
(\byear{2022})
\end{bchapter}
\endbibitem

%%% 7
\bibitem{SiamRCNN}
\begin{bchapter}
\bauthor{\bsnm{Voigtlaender}, \binits{P.}},
\bauthor{\bsnm{Luiten}, \binits{J.}},
\bauthor{\bsnm{Torr}, \binits{P.H.}},
\bauthor{\bsnm{Leibe}, \binits{B.}}:
\bctitle{Siam r-cnn: Visual tracking by re-detection}.
In: \bbtitle{Proceedings of the IEEE/CVF Conference on Computer Vision and Pattern Recognition},
pp. \bfpage{6578}--\blpage{6588}
(\byear{2020})
\end{bchapter}
\endbibitem

%%% 8
\bibitem{OTB2015}
\begin{barticle}
\bauthor{\bsnm{{Wu}}, \binits{Y.}},
\bauthor{\bsnm{{Lim}}, \binits{J.}},
\bauthor{\bsnm{{Yang}}, \binits{M.-H.}}:
\batitle{Object tracking benchmark}.
\bjtitle{IEEE Transactions on Pattern Analysis and Machine Intelligence}
\bvolume{37}(\bissue{9}),
\bfpage{1834}--\blpage{1848}
(\byear{2015})
\end{barticle}
\endbibitem

%%% 9
\bibitem{VOT2016}
\begin{bchapter}
\bauthor{\bsnm{Kristan}, \binits{M.}},
\bauthor{\bsnm{Leonardis}, \binits{A.}},
\bauthor{\bsnm{Matas}, \binits{J.}},
\bauthor{\bsnm{Felsberg}, \binits{M.}},
\bauthor{\bsnm{Pflugfelder}, \binits{R.}},
\bauthor{\bsnm{{\v C}ehovin}, \binits{L.}},
\bauthor{\bsnm{Voj{\'i}{\~r}}, \binits{T.}},
\bauthor{\bsnm{H{\"a}ger}, \binits{G.}},
\bauthor{\bsnm{Luke{\v z}i{\v c}}, \binits{A.}},
\bauthor{\bsnm{Fern{\'a}ndez}, \binits{G.}},
\bauthor{\bsnm{Gupta}, \binits{A.}},
\bauthor{\bsnm{Petrosino}, \binits{A.}},
\bauthor{\bsnm{Memarmoghadam}, \binits{A.}},
\bauthor{\bsnm{{Garcia-Martin}}, \binits{A.}},
\bauthor{\bsnm{Sol{\'i}s~Montero}, \binits{A.}},
\bauthor{\bsnm{others.}}:
\bctitle{The {{Visual Object Tracking VOT2016 Challenge Results}}}.
In: \bbtitle{Computer {{Vision}} \textendash{} {{ECCV}} 2016 {{Workshops}}},
pp. \bfpage{777}--\blpage{823}.
\bpublisher{{Springer}},
\blocation{{Amsterdam, The Netherlands}}
(\byear{2016})
\end{bchapter}
\endbibitem

%%% 10
\bibitem{GOT-10k}
\begin{barticle}
\bauthor{\bsnm{Huang}, \binits{L.}},
\bauthor{\bsnm{Zhao}, \binits{X.}},
\bauthor{\bsnm{Huang}, \binits{K.}}:
\batitle{Got-10k: A large high-diversity benchmark for generic object tracking in the wild}.
\bjtitle{IEEE Transactions on Pattern Analysis and Machine Intelligence}
\bvolume{43}(\bissue{5}),
\bfpage{1562}--\blpage{1577}
(\byear{2021})
\end{barticle}
\endbibitem

%%% 11
\bibitem{UAV123}
\begin{bchapter}
\bauthor{\bsnm{Mueller}, \binits{M.}},
\bauthor{\bsnm{Smith}, \binits{N.}},
\bauthor{\bsnm{Ghanem}, \binits{B.}}:
\bctitle{A benchmark and simulator for uav tracking}.
In: \bbtitle{European Conference on Computer Vision},
pp. \bfpage{445}--\blpage{461}
(\byear{2016}).
\bcomment{Springer}
\end{bchapter}
\endbibitem

%%% 12
\bibitem{UAVDT}
\begin{barticle}
\bauthor{\bsnm{Yu}, \binits{H.}},
\bauthor{\bsnm{Li}, \binits{G.}},
\bauthor{\bsnm{Zhang}, \binits{W.}},
\bauthor{\bsnm{Huang}, \binits{Q.}},
\bauthor{\bsnm{Du}, \binits{D.}},
\bauthor{\bsnm{Tian}, \binits{Q.}},
\bauthor{\bsnm{Sebe}, \binits{N.}}:
\batitle{The unmanned aerial vehicle benchmark: Object detection, tracking and baseline}.
\bjtitle{International Journal of Computer Vision}
\bvolume{128}(\bissue{5}),
\bfpage{1141}--\blpage{1159}
(\byear{2020})
\end{barticle}
\endbibitem

%%% 13
\bibitem{DTB70}
\begin{bchapter}
\bauthor{\bsnm{Li}, \binits{S.}},
\bauthor{\bsnm{Yeung}, \binits{D.-Y.}}:
\bctitle{Visual object tracking for unmanned aerial vehicles: A benchmark and new motion models}.
In: \bbtitle{Thirty-first AAAI Conference on Artificial Intelligence}
(\byear{2017})
\end{bchapter}
\endbibitem

%%% 14
\bibitem{VisDrone}
\begin{barticle}
\bauthor{\bsnm{Zhu}, \binits{P.}},
\bauthor{\bsnm{Wen}, \binits{L.}},
\bauthor{\bsnm{Du}, \binits{D.}},
\bauthor{\bsnm{Bian}, \binits{X.}},
\bauthor{\bsnm{Fan}, \binits{H.}},
\bauthor{\bsnm{Hu}, \binits{Q.}},
\bauthor{\bsnm{Ling}, \binits{H.}}:
\batitle{Detection and tracking meet drones challenge}.
\bjtitle{IEEE Transactions on Pattern Analysis and Machine Intelligence}
\bvolume{44}(\bissue{11}),
\bfpage{7380}--\blpage{7399}
(\byear{2021})
\end{barticle}
\endbibitem

%%% 15
\bibitem{KCF}
\begin{barticle}
\bauthor{\bsnm{Henriques}, \binits{J.F.}},
\bauthor{\bsnm{Caseiro}, \binits{R.}},
\bauthor{\bsnm{Martins}, \binits{P.}},
\bauthor{\bsnm{Batista}, \binits{J.}}:
\batitle{High-speed tracking with kernelized correlation filters}.
\bjtitle{IEEE transactions on pattern analysis and machine intelligence}
\bvolume{37}(\bissue{3}),
\bfpage{583}--\blpage{596}
(\byear{2014})
\end{barticle}
\endbibitem

%%% 16
\bibitem{SiamFC}
\begin{bchapter}
\bauthor{\bsnm{Bertinetto}, \binits{L.}},
\bauthor{\bsnm{Valmadre}, \binits{J.}},
\bauthor{\bsnm{Henriques}, \binits{J.F.}},
\bauthor{\bsnm{Vedaldi}, \binits{A.}},
\bauthor{\bsnm{Torr}, \binits{P.H.}}:
\bctitle{Fully-convolutional siamese networks for object tracking}.
In: \bbtitle{European Conference on Computer Vision},
pp. \bfpage{850}--\blpage{865}
(\byear{2016}).
\bcomment{Springer}
\end{bchapter}
\endbibitem

%%% 17
\bibitem{SiamRPN}
\begin{bchapter}
\bauthor{\bsnm{Li}, \binits{B.}},
\bauthor{\bsnm{Yan}, \binits{J.}},
\bauthor{\bsnm{Wu}, \binits{W.}},
\bauthor{\bsnm{Zhu}, \binits{Z.}},
\bauthor{\bsnm{Hu}, \binits{X.}}:
\bctitle{High performance visual tracking with siamese region proposal network}.
In: \bbtitle{The IEEE Conference on Computer Vision and Pattern Recognition (CVPR)}
(\byear{2018})
\end{bchapter}
\endbibitem

%%% 18
\bibitem{kong2022human}
\begin{barticle}
\bauthor{\bsnm{Kong}, \binits{Y.}},
\bauthor{\bsnm{Fu}, \binits{Y.}}:
\batitle{Human action recognition and prediction: A survey}.
\bjtitle{International Journal of Computer Vision}
\bvolume{130}(\bissue{5}),
\bfpage{1366}--\blpage{1401}
(\byear{2022})
\end{barticle}
\endbibitem

%%% 19
\bibitem{dendorfer2021motchallenge}
\begin{barticle}
\bauthor{\bsnm{Dendorfer}, \binits{P.}},
\bauthor{\bsnm{Osep}, \binits{A.}},
\bauthor{\bsnm{Milan}, \binits{A.}},
\bauthor{\bsnm{Schindler}, \binits{K.}},
\bauthor{\bsnm{Cremers}, \binits{D.}},
\bauthor{\bsnm{Reid}, \binits{I.}},
\bauthor{\bsnm{Roth}, \binits{S.}},
\bauthor{\bsnm{Leal-Taix{\'e}}, \binits{L.}}:
\batitle{Motchallenge: A benchmark for single-camera multiple target tracking}.
\bjtitle{International Journal of Computer Vision}
\bvolume{129}(\bissue{4}),
\bfpage{845}--\blpage{881}
(\byear{2021})
\end{barticle}
\endbibitem

%%% 20
\bibitem{abu2018augmented}
\begin{barticle}
\bauthor{\bsnm{Abu~Alhaija}, \binits{H.}},
\bauthor{\bsnm{Mustikovela}, \binits{S.K.}},
\bauthor{\bsnm{Mescheder}, \binits{L.}},
\bauthor{\bsnm{Geiger}, \binits{A.}},
\bauthor{\bsnm{Rother}, \binits{C.}}:
\batitle{Augmented reality meets computer vision: Efficient data generation for urban driving scenes}.
\bjtitle{International Journal of Computer Vision}
\bvolume{126}(\bissue{9}),
\bfpage{961}--\blpage{972}
(\byear{2018})
\end{barticle}
\endbibitem

%%% 21
\bibitem{gauglitz2011evaluation}
\begin{barticle}
\bauthor{\bsnm{Gauglitz}, \binits{S.}},
\bauthor{\bsnm{H{\"o}llerer}, \binits{T.}},
\bauthor{\bsnm{Turk}, \binits{M.}}:
\batitle{Evaluation of interest point detectors and feature descriptors for visual tracking}.
\bjtitle{International journal of computer vision}
\bvolume{94}(\bissue{3}),
\bfpage{335}--\blpage{360}
(\byear{2011})
\end{barticle}
\endbibitem

%%% 22
\bibitem{robot-navigation}
\begin{botherref}
\oauthor{\bsnm{{Dupeyroux}}, \binits{J.}},
\oauthor{\bsnm{{Serres}}, \binits{J.R.}},
\oauthor{\bsnm{{Viollet}}, \binits{S.}}:
Antbot: A six-legged walking robot able to home like desert ants in outdoor environments.
Science Robotics
\textbf{4}(27)
(2019)
\end{botherref}
\endbibitem

%%% 23
\bibitem{ramakrishnan2021exploration}
\begin{barticle}
\bauthor{\bsnm{Ramakrishnan}, \binits{S.K.}},
\bauthor{\bsnm{Jayaraman}, \binits{D.}},
\bauthor{\bsnm{Grauman}, \binits{K.}}:
\batitle{An exploration of embodied visual exploration}.
\bjtitle{International Journal of Computer Vision}
\bvolume{129}(\bissue{5}),
\bfpage{1616}--\blpage{1649}
(\byear{2021})
\end{barticle}
\endbibitem

%%% 24
\bibitem{CARPK}
\begin{bchapter}
\bauthor{\bsnm{Hsieh}, \binits{M.-R.}},
\bauthor{\bsnm{Lin}, \binits{Y.-L.}},
\bauthor{\bsnm{Hsu}, \binits{W.H.}}:
\bctitle{Drone-based object counting by spatially regularized regional proposal network}.
In: \bbtitle{Proceedings of the IEEE International Conference on Computer Vision},
pp. \bfpage{4145}--\blpage{4153}
(\byear{2017})
\end{bchapter}
\endbibitem

%%% 25
\bibitem{DOTA}
\begin{bchapter}
\bauthor{\bsnm{Xia}, \binits{G.-S.}},
\bauthor{\bsnm{Bai}, \binits{X.}},
\bauthor{\bsnm{Ding}, \binits{J.}},
\bauthor{\bsnm{Zhu}, \binits{Z.}},
\bauthor{\bsnm{Belongie}, \binits{S.}},
\bauthor{\bsnm{Luo}, \binits{J.}},
\bauthor{\bsnm{Datcu}, \binits{M.}},
\bauthor{\bsnm{Pelillo}, \binits{M.}},
\bauthor{\bsnm{Zhang}, \binits{L.}}:
\bctitle{Dota: A large-scale dataset for object detection in aerial images}.
In: \bbtitle{Proceedings of the IEEE Conference on Computer Vision and Pattern Recognition},
pp. \bfpage{3974}--\blpage{3983}
(\byear{2018})
\end{bchapter}
\endbibitem

%%% 26
\bibitem{BIRDSAI}
\begin{bchapter}
\bauthor{\bsnm{Bondi}, \binits{E.}},
\bauthor{\bsnm{Jain}, \binits{R.}},
\bauthor{\bsnm{Aggrawal}, \binits{P.}},
\bauthor{\bsnm{Anand}, \binits{S.}},
\bauthor{\bsnm{Hannaford}, \binits{R.}},
\bauthor{\bsnm{Kapoor}, \binits{A.}},
\bauthor{\bsnm{Piavis}, \binits{J.}},
\bauthor{\bsnm{Shah}, \binits{S.}},
\bauthor{\bsnm{Joppa}, \binits{L.}},
\bauthor{\bsnm{Dilkina}, \binits{B.}}, \betal:
\bctitle{Birdsai: A dataset for detection and tracking in aerial thermal infrared videos}.
In: \bbtitle{Proceedings of the IEEE/CVF Winter Conference on Applications of Computer Vision},
pp. \bfpage{1747}--\blpage{1756}
(\byear{2020})
\end{bchapter}
\endbibitem

%%% 27
\bibitem{hu2022sensaturban}
\begin{barticle}
\bauthor{\bsnm{Hu}, \binits{Q.}},
\bauthor{\bsnm{Yang}, \binits{B.}},
\bauthor{\bsnm{Khalid}, \binits{S.}},
\bauthor{\bsnm{Xiao}, \binits{W.}},
\bauthor{\bsnm{Trigoni}, \binits{N.}},
\bauthor{\bsnm{Markham}, \binits{A.}}:
\batitle{Sensaturban: Learning semantics from urban-scale photogrammetric point clouds}.
\bjtitle{International Journal of Computer Vision}
\bvolume{130}(\bissue{2}),
\bfpage{316}--\blpage{343}
(\byear{2022})
\end{barticle}
\endbibitem

%%% 28
\bibitem{muller2018sim4cv}
\begin{barticle}
\bauthor{\bsnm{M{\"u}ller}, \binits{M.}},
\bauthor{\bsnm{Casser}, \binits{V.}},
\bauthor{\bsnm{Lahoud}, \binits{J.}},
\bauthor{\bsnm{Smith}, \binits{N.}},
\bauthor{\bsnm{Ghanem}, \binits{B.}}:
\batitle{Sim4cv: A photo-realistic simulator for computer vision applications}.
\bjtitle{International Journal of Computer Vision}
\bvolume{126}(\bissue{9}),
\bfpage{902}--\blpage{919}
(\byear{2018})
\end{barticle}
\endbibitem

%%% 29
\bibitem{OTB2013}
\begin{bchapter}
\bauthor{\bsnm{{Wu}}, \binits{Y.}},
\bauthor{\bsnm{{Lim}}, \binits{J.}},
\bauthor{\bsnm{{Yang}}, \binits{M.-H.}}:
\bctitle{Online object tracking: A benchmark}.
In: \bbtitle{2013 IEEE Conference on Computer Vision and Pattern Recognition},
pp. \bfpage{2411}--\blpage{2418}
(\byear{2013})
\end{bchapter}
\endbibitem

%%% 30
\bibitem{VOT2013}
\begin{bchapter}
\bauthor{\bsnm{Kristan}, \binits{M.}},
\bauthor{\bsnm{Pflugfelder}, \binits{R.}},
\bauthor{\bsnm{Leonardis}, \binits{A.}},
\bauthor{\bsnm{Matas}, \binits{J.}},
\bauthor{\bsnm{Porikli}, \binits{F.}},
\bauthor{\bsnm{Cehovin}, \binits{L.}},
\bauthor{\bsnm{Nebehay}, \binits{G.}},
\bauthor{\bsnm{Fernandez}, \binits{G.}},
\bauthor{\bsnm{Vojir}, \binits{T.}},
\bauthor{\bsnm{Gatt}, \binits{A.}},
\bauthor{\bsnm{Khajenezhad}, \binits{A.}},
\bauthor{\bsnm{Salahledin}, \binits{A.}},
\bauthor{\bsnm{{Soltani-Farani}}, \binits{A.}},
\bauthor{\bsnm{others.}}:
\bctitle{The {{Visual Object Tracking VOT2013 Challenge Results}}}.
In: \bbtitle{Proceedings of 2013 {{IEEE International Conference}} on {{Computer Vision Workshops}} ({{ICCVW}})},
pp. \bfpage{98}--\blpage{111}.
\bpublisher{{IEEE}},
\blocation{{Sydney, Australia}}
(\byear{2013})
\end{bchapter}
\endbibitem

%%% 31
\bibitem{VOT2014}
\begin{bchapter}
\bauthor{\bsnm{Kristan}, \binits{M.}},
\bauthor{\bsnm{Pflugfelder}, \binits{R.P.}},
\bauthor{\bsnm{Leonardis}, \binits{A.}},
\bauthor{\bsnm{Matas}, \binits{J.}},
\bauthor{\bsnm{Cehovin}, \binits{L.}},
\bauthor{\bsnm{Nebehay}, \binits{G.}},
\bauthor{\bsnm{Voj{\'i}r}, \binits{T.}},
\bauthor{\bsnm{Fern{\'a}ndez}, \binits{G.}},
\bauthor{\bsnm{Lukezic}, \binits{A.}},
\bauthor{\bsnm{Dimitriev}, \binits{A.}},
\bauthor{\bsnm{Petrosino}, \binits{A.}},
\bauthor{\bsnm{Saffari}, \binits{A.a.}},
\bauthor{\bsnm{others.}}:
\bctitle{The {{Visual Object Tracking VOT2014 Challenge Results}}}.
In: \beditor{\bsnm{Agapito}, \binits{L.}},
\beditor{\bsnm{Bronstein}, \binits{M.M.}},
\beditor{\bsnm{Rother}, \binits{C.}} (eds.)
\bbtitle{Computer {{Vision}} - {{ECCV}} 2014 {{Workshops}}},
vol. \bseriesno{8926},
pp. \bfpage{191}--\blpage{217}.
\bpublisher{{Springer}},
\blocation{{Zurich, Switzerland}}
(\byear{2014})
\end{bchapter}
\endbibitem

%%% 32
\bibitem{VOT2015}
\begin{bchapter}
\bauthor{\bsnm{Kristan}, \binits{M.}},
\bauthor{\bsnm{Matas}, \binits{J.}},
\bauthor{\bsnm{Leonardis}, \binits{A.}},
\bauthor{\bsnm{Felsberg}, \binits{M.}},
\bauthor{\bsnm{Cehovin}, \binits{L.}},
\bauthor{\bsnm{Fernandez}, \binits{G.}},
\bauthor{\bsnm{Vojir}, \binits{T.}},
\bauthor{\bsnm{Hager}, \binits{G.}},
\bauthor{\bsnm{Nebehay}, \binits{G.}},
\bauthor{\bsnm{Pflugfelder}, \binits{R.}},
\bauthor{\bsnm{Gupta}, \binits{A.}},
\bauthor{\bsnm{Bibi}, \binits{A.}},
\bauthor{\bsnm{Lukezic}, \binits{A.}},
\bauthor{\bsnm{{Garcia-Martin}}, \binits{A.}},
\bauthor{\bsnm{Saffari}, \binits{A.}},
\bauthor{\bsnm{Petrosino}, \binits{A.}},
\bauthor{\bsnm{Solis~Montero}, \binits{A.}}:
\bctitle{The {{Visual Object Tracking VOT2015 Challenge Results}}}.
In: \bbtitle{Proceedings of 2015 {{IEEE International Conference}} on {{Computer Vision Workshop}} ({{ICCVW}})},
pp. \bfpage{564}--\blpage{586}.
\bpublisher{{IEEE}},
\blocation{{Santiago, Chile}}
(\byear{2015})
\end{bchapter}
\endbibitem

%%% 33
\bibitem{VOT2017}
\begin{bchapter}
\bauthor{\bsnm{Kristan}, \binits{M.}},
\bauthor{\bsnm{Leonardis}, \binits{A.}},
\bauthor{\bsnm{Matas}, \binits{J.}},
\bauthor{\bsnm{Felsberg}, \binits{M.}},
\bauthor{\bsnm{Pflugfelder}, \binits{R.}},
\bauthor{\bsnm{Zajc}, \binits{L.C.}},
\bauthor{\bsnm{Voj{\'i}r}, \binits{T.}},
\bauthor{\bsnm{H{\"a}ger}, \binits{G.}},
\bauthor{\bsnm{Luke{\v z}ic}, \binits{A.}},
\bauthor{\bsnm{Eldesokey}, \binits{A.}},
\bauthor{\bsnm{Fern{\'a}ndez}, \binits{G.}},
\bauthor{\bsnm{{Garc{\'i}a-Mart{\'i}n}}, \binits{{\'A}.}},
\bauthor{\bsnm{Muhic}, \binits{A.}},
\bauthor{\bsnm{Petrosino}, \binits{A.}},
\bauthor{\bsnm{Memarmoghadam}, \binits{A.}},
\bauthor{\bsnm{others.}}:
\bctitle{The {{Visual Object Tracking VOT2017 Challenge Results}}}.
In: \bbtitle{Proceedings of 2017 {{IEEE International Conference}} on {{Computer Vision Workshops}} ({{ICCVW}})},
pp. \bfpage{1949}--\blpage{1972}.
\bpublisher{{IEEE}},
\blocation{{Venice, Italy}}
(\byear{2017})
\end{bchapter}
\endbibitem

%%% 34
\bibitem{VOT2020}
\begin{bchapter}
\bauthor{\bsnm{Kristan}, \binits{M.}},
\bauthor{\bsnm{Leonardis}, \binits{A.}},
\bauthor{\bsnm{Matas}, \binits{J.}},
\bauthor{\bsnm{Felsberg}, \binits{M.}},
\bauthor{\bsnm{Pflugfelder}, \binits{R.}},
\bauthor{\bsnm{K{\"a}m{\"a}r{\"a}inen}, \binits{J.-K.}},
\bauthor{\bsnm{Danelljan}, \binits{M.}},
\bauthor{\bsnm{Zajc}, \binits{L.{\v C}.}},
\bauthor{\bsnm{Luke{\v z}i{\v c}}, \binits{A.}},
\bauthor{\bsnm{Drbohlav}, \binits{O.}},
\bauthor{\bsnm{He}, \binits{L.}},
\bauthor{\bsnm{others.}}:
\bctitle{The {{Eighth Visual Object Tracking VOT2020 Challenge Results}}}.
In: \bbtitle{Computer {{Vision}} \textendash{} {{ECCV}} 2020 {{Workshops}}},
pp. \bfpage{547}--\blpage{601}.
\bpublisher{{Springer}},
\blocation{{Glasgow, UK}}
(\byear{2020})
\end{bchapter}
\endbibitem

%%% 35
\bibitem{VOT2021}
\begin{bchapter}
\bauthor{\bsnm{Kristan}, \binits{M.}},
\bauthor{\bsnm{Matas}, \binits{J.}},
\bauthor{\bsnm{Leonardis}, \binits{A.}},
\bauthor{\bsnm{Felsberg}, \binits{M.}},
\bauthor{\bsnm{Pflugfelder}, \binits{R.}},
\bauthor{\bsnm{K{\"a}m{\"a}r{\"a}inen}, \binits{J.-K.}},
\bauthor{\bsnm{Chang}, \binits{H.J.}},
\bauthor{\bsnm{Danelljan}, \binits{M.}},
\bauthor{\bsnm{Zajc}, \binits{L.{\v C}.}},
\bauthor{\bsnm{Luke{\v z}i{\v c}}, \binits{A.}},
\bauthor{\bsnm{Drbohlav}, \binits{O.}},
\bauthor{\bsnm{others.}}:
\bctitle{The {{Ninth Visual Object Tracking VOT2021 Challenge Results}}}.
In: \bbtitle{Proceedings of 2021 {{IEEE}}/{{CVF International Conference}} on {{Computer Vision Workshops}} ({{ICCVW}})},
pp. \bfpage{2711}--\blpage{2738}.
\bpublisher{{IEEE}},
\blocation{{Montreal, BC, Canada}}
(\byear{2021})
\end{bchapter}
\endbibitem

%%% 36
\bibitem{mcmasters2002airplane}
\begin{barticle}
\bauthor{\bsnm{McMasters}, \binits{J.H.}},
\bauthor{\bsnm{Cummings}, \binits{R.M.}}:
\batitle{Airplane design-past, present, and future}.
\bjtitle{Journal of Aircraft}
\bvolume{39}(\bissue{1}),
\bfpage{10}--\blpage{17}
(\byear{2002})
\end{barticle}
\endbibitem

%%% 37
\bibitem{mcmasters2004rethinking}
\begin{bchapter}
\bauthor{\bsnm{McMasters}, \binits{J.}},
\bauthor{\bsnm{Cummings}, \binits{R.}}:
\bctitle{Rethinking the airplane design process-an early 21st century perspective}.
In: \bbtitle{42nd AIAA Aerospace Sciences Meeting and Exhibit},
p. \bfpage{693}
(\byear{2004})
\end{bchapter}
\endbibitem

%%% 38
\bibitem{sims1991understanding}
\begin{botherref}
\oauthor{\bsnm{Sims}, \binits{C.A.}},
\oauthor{\bsnm{Uhlig}, \binits{H.}}:
Understanding unit rooters: A helicopter tour.
Econometrica: Journal of the Econometric Society,
1591--1599
(1991)
\end{botherref}
\endbibitem

%%% 39
\bibitem{fraire2015design}
\begin{bchapter}
\bauthor{\bsnm{Fraire}, \binits{A.E.}},
\bauthor{\bsnm{Morado}, \binits{R.P.}},
\bauthor{\bsnm{L{\'o}pez}, \binits{A.D.}},
\bauthor{\bsnm{Leal}, \binits{R.L.}}:
\bctitle{Design and implementation of fixed-wing mav controllers}.
In: \bbtitle{2015 Workshop on Research, Education and Development of Unmanned Aerial Systems (RED-UAS)},
pp. \bfpage{172}--\blpage{179}
(\byear{2015}).
\bcomment{IEEE}
\end{bchapter}
\endbibitem

%%% 40
\bibitem{barrientos2010rotary}
\begin{bchapter}
\bauthor{\bsnm{Barrientos}, \binits{A.}},
\bauthor{\bsnm{Colorado}, \binits{J.}},
\bauthor{\bsnm{Martinez}, \binits{A.}},
\bauthor{\bsnm{Valente}, \binits{J.}}:
\bctitle{Rotary-wing mav modeling \& control for indoor scenarios}.
In: \bbtitle{2010 IEEE International Conference on Industrial Technology},
pp. \bfpage{1475}--\blpage{1480}
(\byear{2010}).
\bcomment{IEEE}
\end{bchapter}
\endbibitem

%%% 41
\bibitem{lee2018effect}
\begin{barticle}
\bauthor{\bsnm{Lee}, \binits{N.}},
\bauthor{\bsnm{Lee}, \binits{S.}},
\bauthor{\bsnm{Cho}, \binits{H.}},
\bauthor{\bsnm{Shin}, \binits{S.}}:
\batitle{Effect of flexibility on flapping wing characteristics in hover and forward flight}.
\bjtitle{Computers \& Fluids}
\bvolume{173},
\bfpage{111}--\blpage{117}
(\byear{2018})
\end{barticle}
\endbibitem

%%% 42
\bibitem{zhang2017review}
\begin{barticle}
\bauthor{\bsnm{Zhang}, \binits{C.}},
\bauthor{\bsnm{Rossi}, \binits{C.}}:
\batitle{A review of compliant transmission mechanisms for bio-inspired flapping-wing micro air vehicles}.
\bjtitle{Bioinspiration \& biomimetics}
\bvolume{12}(\bissue{2}),
\bfpage{025005}
(\byear{2017})
\end{barticle}
\endbibitem

%%% 43
\bibitem{pornsin2001microbat}
\begin{bchapter}
\bauthor{\bsnm{Pornsin-Sirirak}, \binits{T.N.}},
\bauthor{\bsnm{Tai}, \binits{Y.-C.}},
\bauthor{\bsnm{Ho}, \binits{C.-M.}},
\bauthor{\bsnm{Keennon}, \binits{M.}}:
\bctitle{Microbat: A palm-sized electrically powered ornithopter}.
In: \bbtitle{Proceedings of NASA/JPL Workshop on Biomorphic Robotics},
vol. \bseriesno{14},
p. \bfpage{17}
(\byear{2001}).
\bcomment{Citeseer}
\end{bchapter}
\endbibitem

%%% 44
\bibitem{rigelsford2004neurotechnology}
\begin{barticle}
\bauthor{\bsnm{Rigelsford}, \binits{J.}}:
\batitle{Neurotechnology for biomimetic robots}.
\bjtitle{Industrial Robot: An International Journal}
\bvolume{31}(\bissue{6}),
\bfpage{534}--\blpage{534}
(\byear{2004})
\end{barticle}
\endbibitem

%%% 45
\bibitem{de2016delfly}
\begin{barticle}
\bauthor{\bsnm{De~Croon}, \binits{G.}},
\bauthor{\bsnm{Per{\c{c}}in}, \binits{M.}},
\bauthor{\bsnm{Remes}, \binits{B.}},
\bauthor{\bsnm{Ruijsink}, \binits{R.}},
\bauthor{\bsnm{De~Wagter}, \binits{C.}}:
\batitle{The delfly}.
\bjtitle{Dordrecht: Springer Netherlands. doi}
\bvolume{10},
\bfpage{978}--\blpage{94}
(\byear{2016})
\end{barticle}
\endbibitem

%%% 46
\bibitem{ryu2016autonomous}
\begin{bchapter}
\bauthor{\bsnm{Ryu}, \binits{S.}},
\bauthor{\bsnm{Kwon}, \binits{U.}},
\bauthor{\bsnm{Kim}, \binits{H.J.}}:
\bctitle{Autonomous flight and vision-based target tracking for a flapping-wing mav}.
In: \bbtitle{2016 IEEE/RSJ International Conference on Intelligent Robots and Systems (IROS)},
pp. \bfpage{5645}--\blpage{5650}
(\byear{2016}).
\bcomment{IEEE}
\end{bchapter}
\endbibitem

%%% 47
\bibitem{yang2018dove}
\begin{barticle}
\bauthor{\bsnm{Yang}, \binits{W.}},
\bauthor{\bsnm{Wang}, \binits{L.}},
\bauthor{\bsnm{Song}, \binits{B.}}:
\batitle{Dove: A biomimetic flapping-wing micro air vehicle}.
\bjtitle{International Journal of Micro Air Vehicles}
\bvolume{10}(\bissue{1}),
\bfpage{70}--\blpage{84}
(\byear{2018})
\end{barticle}
\endbibitem

%%% 48
\bibitem{liu2020deep}
\begin{barticle}
\bauthor{\bsnm{Liu}, \binits{L.}},
\bauthor{\bsnm{Ouyang}, \binits{W.}},
\bauthor{\bsnm{Wang}, \binits{X.}},
\bauthor{\bsnm{Fieguth}, \binits{P.}},
\bauthor{\bsnm{Chen}, \binits{J.}},
\bauthor{\bsnm{Liu}, \binits{X.}},
\bauthor{\bsnm{Pietik{\"a}inen}, \binits{M.}}:
\batitle{Deep learning for generic object detection: A survey}.
\bjtitle{International journal of computer vision}
\bvolume{128}(\bissue{2}),
\bfpage{261}--\blpage{318}
(\byear{2020})
\end{barticle}
\endbibitem

%%% 49
\bibitem{russakovsky2015imagenet}
\begin{barticle}
\bauthor{\bsnm{Russakovsky}, \binits{O.}},
\bauthor{\bsnm{Deng}, \binits{J.}},
\bauthor{\bsnm{Su}, \binits{H.}},
\bauthor{\bsnm{Krause}, \binits{J.}},
\bauthor{\bsnm{Satheesh}, \binits{S.}},
\bauthor{\bsnm{Ma}, \binits{S.}},
\bauthor{\bsnm{Huang}, \binits{Z.}},
\bauthor{\bsnm{Karpathy}, \binits{A.}},
\bauthor{\bsnm{Khosla}, \binits{A.}},
\bauthor{\bsnm{Bernstein}, \binits{M.}}, \betal:
\batitle{Imagenet large scale visual recognition challenge}.
\bjtitle{International journal of computer vision}
\bvolume{115}(\bissue{3}),
\bfpage{211}--\blpage{252}
(\byear{2015})
\end{barticle}
\endbibitem

%%% 50
\bibitem{han2021context}
\begin{barticle}
\bauthor{\bsnm{Han}, \binits{L.}},
\bauthor{\bsnm{Wang}, \binits{P.}},
\bauthor{\bsnm{Yin}, \binits{Z.}},
\bauthor{\bsnm{Wang}, \binits{F.}},
\bauthor{\bsnm{Li}, \binits{H.}}:
\batitle{Context and structure mining network for video object detection}.
\bjtitle{International Journal of Computer Vision}
\bvolume{129}(\bissue{10}),
\bfpage{2927}--\blpage{2946}
(\byear{2021})
\end{barticle}
\endbibitem

%%% 51
\bibitem{wu2021deep}
\begin{barticle}
\bauthor{\bsnm{Wu}, \binits{X.}},
\bauthor{\bsnm{Li}, \binits{W.}},
\bauthor{\bsnm{Hong}, \binits{D.}},
\bauthor{\bsnm{Tao}, \binits{R.}},
\bauthor{\bsnm{Du}, \binits{Q.}}:
\batitle{Deep learning for unmanned aerial vehicle-based object detection and tracking: a survey}.
\bjtitle{IEEE Geoscience and Remote Sensing Magazine}
\bvolume{10}(\bissue{1}),
\bfpage{91}--\blpage{124}
(\byear{2021})
\end{barticle}
\endbibitem

%%% 52
\bibitem{luiten2021hota}
\begin{barticle}
\bauthor{\bsnm{Luiten}, \binits{J.}},
\bauthor{\bsnm{Osep}, \binits{A.}},
\bauthor{\bsnm{Dendorfer}, \binits{P.}},
\bauthor{\bsnm{Torr}, \binits{P.}},
\bauthor{\bsnm{Geiger}, \binits{A.}},
\bauthor{\bsnm{Leal-Taix{\'e}}, \binits{L.}},
\bauthor{\bsnm{Leibe}, \binits{B.}}:
\batitle{Hota: A higher order metric for evaluating multi-object tracking}.
\bjtitle{International journal of computer vision}
\bvolume{129}(\bissue{2}),
\bfpage{548}--\blpage{578}
(\byear{2021})
\end{barticle}
\endbibitem

%%% 53
\bibitem{bondi2018airsim}
\begin{bchapter}
\bauthor{\bsnm{Bondi}, \binits{E.}},
\bauthor{\bsnm{Dey}, \binits{D.}},
\bauthor{\bsnm{Kapoor}, \binits{A.}},
\bauthor{\bsnm{Piavis}, \binits{J.}},
\bauthor{\bsnm{Shah}, \binits{S.}},
\bauthor{\bsnm{Fang}, \binits{F.}},
\bauthor{\bsnm{Dilkina}, \binits{B.}},
\bauthor{\bsnm{Hannaford}, \binits{R.}},
\bauthor{\bsnm{Iyer}, \binits{A.}},
\bauthor{\bsnm{Joppa}, \binits{L.}}, \betal:
\bctitle{Airsim-w: A simulation environment for wildlife conservation with uavs}.
In: \bbtitle{Proceedings of the 1st ACM SIGCAS Conference on Computing and Sustainable Societies},
pp. \bfpage{1}--\blpage{12}
(\byear{2018})
\end{bchapter}
\endbibitem

%%% 54
\bibitem{TrackingNet}
\begin{bchapter}
\bauthor{\bsnm{Muller}, \binits{M.}},
\bauthor{\bsnm{Bibi}, \binits{A.}},
\bauthor{\bsnm{Giancola}, \binits{S.}},
\bauthor{\bsnm{Alsubaihi}, \binits{S.}},
\bauthor{\bsnm{Ghanem}, \binits{B.}}:
\bctitle{Trackingnet: A large-scale dataset and benchmark for object tracking in the wild}.
In: \bbtitle{Proceedings of the European Conference on Computer Vision (ECCV)},
pp. \bfpage{300}--\blpage{317}
(\byear{2018})
\end{bchapter}
\endbibitem

%%% 55
\bibitem{SOTVerse}
\begin{botherref}
\oauthor{\bsnm{Hu}, \binits{S.}},
\oauthor{\bsnm{Zhao}, \binits{X.}},
\oauthor{\bsnm{Huang}, \binits{K.}}:
Sotverse: A user-defined task space of single object tracking.
arXiv preprint arXiv:2204.07414
(2022)
\end{botherref}
\endbibitem

%%% 56
\bibitem{ShadeofGray}
\begin{bchapter}
\bauthor{\bsnm{{Finlayson}}, \binits{G.D.}},
\bauthor{\bsnm{{Trezzi}}, \binits{E.}}:
\bctitle{Shades of gray and colour constancy}.
In: \bbtitle{The Twelfth Color Imaging Conference 2004},
pp. \bfpage{37}--\blpage{41}
(\byear{2004})
\end{bchapter}
\endbibitem

%%% 57
\bibitem{Laplacian}
\begin{bchapter}
\bauthor{\bsnm{{Pech-Pacheco}}, \binits{J.L.}},
\bauthor{\bsnm{{Cristobal}}, \binits{G.}},
\bauthor{\bsnm{{Chamorro-Martinez}}, \binits{J.}},
\bauthor{\bsnm{{Fernandez-Valdivia}}, \binits{J.}}:
\bctitle{Diatom autofocusing in brightfield microscopy: a comparative study}.
In: \bbtitle{Proceedings 15th International Conference on Pattern Recognition. ICPR-2000},
vol. \bseriesno{3},
pp. \bfpage{314}--\blpage{317}
(\byear{2000})
\end{bchapter}
\endbibitem

%%% 58
\bibitem{ECO}
\begin{bchapter}
\bauthor{\bsnm{Danelljan}, \binits{M.}},
\bauthor{\bsnm{Bhat}, \binits{G.}},
\bauthor{\bsnm{Shahbaz~Khan}, \binits{F.}},
\bauthor{\bsnm{Felsberg}, \binits{M.}}:
\bctitle{Eco: Efficient convolution operators for tracking}.
In: \bbtitle{Proceedings of the IEEE Conference on Computer Vision and Pattern Recognition},
pp. \bfpage{6638}--\blpage{6646}
(\byear{2017})
\end{bchapter}
\endbibitem

%%% 59
\bibitem{DaSiamRPN}
\begin{bchapter}
\bauthor{\bsnm{Zhu}, \binits{Z.}},
\bauthor{\bsnm{Wang}, \binits{Q.}},
\bauthor{\bsnm{Li}, \binits{B.}},
\bauthor{\bsnm{Wu}, \binits{W.}},
\bauthor{\bsnm{Yan}, \binits{J.}},
\bauthor{\bsnm{Hu}, \binits{W.}}:
\bctitle{Distractor-aware siamese networks for visual object tracking}.
In: \bbtitle{Proceedings of the European Conference on Computer Vision (ECCV)},
pp. \bfpage{101}--\blpage{117}
(\byear{2018})
\end{bchapter}
\endbibitem

%%% 60
\bibitem{ATOM}
\begin{bchapter}
\bauthor{\bsnm{Danelljan}, \binits{M.}},
\bauthor{\bsnm{Bhat}, \binits{G.}},
\bauthor{\bsnm{Khan}, \binits{F.S.}},
\bauthor{\bsnm{Felsberg}, \binits{M.}}:
\bctitle{Atom: Accurate tracking by overlap maximization}.
In: \bbtitle{Proceedings of the IEEE/CVF Conference on Computer Vision and Pattern Recognition},
pp. \bfpage{4660}--\blpage{4669}
(\byear{2019})
\end{bchapter}
\endbibitem

%%% 61
\bibitem{SiamRPN++}
\begin{bchapter}
\bauthor{\bsnm{Li}, \binits{B.}},
\bauthor{\bsnm{Wu}, \binits{W.}},
\bauthor{\bsnm{Wang}, \binits{Q.}},
\bauthor{\bsnm{Zhang}, \binits{F.}},
\bauthor{\bsnm{Xing}, \binits{J.}},
\bauthor{\bsnm{Yan}, \binits{J.}}:
\bctitle{Siamrpn++: Evolution of siamese visual tracking with very deep networks}.
In: \bbtitle{Proceedings of the IEEE/CVF Conference on Computer Vision and Pattern Recognition},
pp. \bfpage{4282}--\blpage{4291}
(\byear{2019})
\end{bchapter}
\endbibitem

%%% 62
\bibitem{SiamDW}
\begin{bchapter}
\bauthor{\bsnm{Zhang}, \binits{Z.}},
\bauthor{\bsnm{Peng}, \binits{H.}}:
\bctitle{Deeper and wider siamese networks for real-time visual tracking}.
In: \bbtitle{Proceedings of the IEEE/CVF Conference on Computer Vision and Pattern Recognition},
pp. \bfpage{4591}--\blpage{4600}
(\byear{2019})
\end{bchapter}
\endbibitem

%%% 63
\bibitem{DiMP}
\begin{bchapter}
\bauthor{\bsnm{Bhat}, \binits{G.}},
\bauthor{\bsnm{Danelljan}, \binits{M.}},
\bauthor{\bsnm{Gool}, \binits{L.V.}},
\bauthor{\bsnm{Timofte}, \binits{R.}}:
\bctitle{Learning discriminative model prediction for tracking}.
In: \bbtitle{Proceedings of the IEEE/CVF International Conference on Computer Vision},
pp. \bfpage{6182}--\blpage{6191}
(\byear{2019})
\end{bchapter}
\endbibitem

%%% 64
\bibitem{GlobalTrack}
\begin{bchapter}
\bauthor{\bsnm{Huang}, \binits{L.}},
\bauthor{\bsnm{Zhao}, \binits{X.}},
\bauthor{\bsnm{Huang}, \binits{K.}}:
\bctitle{Globaltrack: A simple and strong baseline for long-term tracking}.
In: \bbtitle{Proceedings of the AAAI Conference on Artificial Intelligence},
vol. \bseriesno{34},
pp. \bfpage{11037}--\blpage{11044}
(\byear{2020})
\end{bchapter}
\endbibitem

%%% 65
\bibitem{SiamFC++}
\begin{bchapter}
\bauthor{\bsnm{Xu}, \binits{Y.}},
\bauthor{\bsnm{Wang}, \binits{Z.}},
\bauthor{\bsnm{Li}, \binits{Z.}},
\bauthor{\bsnm{Yuan}, \binits{Y.}},
\bauthor{\bsnm{Yu}, \binits{G.}}:
\bctitle{Siamfc++: Towards robust and accurate visual tracking with target estimation guidelines}.
In: \bbtitle{Proceedings of the AAAI Conference on Artificial Intelligence},
vol. \bseriesno{34},
pp. \bfpage{12549}--\blpage{12556}
(\byear{2020})
\end{bchapter}
\endbibitem

%%% 66
\bibitem{Ocean}
\begin{bchapter}
\bauthor{\bsnm{Zhang}, \binits{Z.}},
\bauthor{\bsnm{Peng}, \binits{H.}},
\bauthor{\bsnm{Fu}, \binits{J.}},
\bauthor{\bsnm{Li}, \binits{B.}},
\bauthor{\bsnm{Hu}, \binits{W.}}:
\bctitle{Ocean: Object-aware anchor-free tracking}.
In: \bbtitle{European Conference on Computer Vision},
pp. \bfpage{771}--\blpage{787}
(\byear{2020}).
\bcomment{Springer}
\end{bchapter}
\endbibitem

%%% 67
\bibitem{KYS}
\begin{bchapter}
\bauthor{\bsnm{Bhat}, \binits{G.}},
\bauthor{\bsnm{Danelljan}, \binits{M.}},
\bauthor{\bsnm{Gool}, \binits{L.V.}},
\bauthor{\bsnm{Timofte}, \binits{R.}}:
\bctitle{Know your surroundings: Exploiting scene information for object tracking}.
In: \bbtitle{European Conference on Computer Vision},
pp. \bfpage{205}--\blpage{221}
(\byear{2020}).
\bcomment{Springer}
\end{bchapter}
\endbibitem

%%% 68
\bibitem{SiamCAR}
\begin{bchapter}
\bauthor{\bsnm{Guo}, \binits{D.}},
\bauthor{\bsnm{Wang}, \binits{J.}},
\bauthor{\bsnm{Cui}, \binits{Y.}},
\bauthor{\bsnm{Wang}, \binits{Z.}},
\bauthor{\bsnm{Chen}, \binits{S.}}:
\bctitle{Siamcar: Siamese fully convolutional classification and regression for visual tracking}.
In: \bbtitle{Proceedings of the IEEE/CVF Conference on Computer Vision and Pattern Recognition},
pp. \bfpage{6269}--\blpage{6277}
(\byear{2020})
\end{bchapter}
\endbibitem

%%% 69
\bibitem{PrDiMP}
\begin{botherref}
\oauthor{\bsnm{Danelljan}, \binits{M.}},
\oauthor{\bsnm{Gool}, \binits{L.V.}},
\oauthor{\bsnm{Timofte}, \binits{R.}}:
Probabilistic regression for visual tracking.
2020 IEEE Conference on Computer Vision and Pattern Recognition (CVPR)
(2020)
\end{botherref}
\endbibitem

%%% 70
\bibitem{TCTrack}
\begin{bchapter}
\bauthor{\bsnm{Cao}, \binits{Z.}},
\bauthor{\bsnm{Huang}, \binits{Z.}},
\bauthor{\bsnm{Pan}, \binits{L.}},
\bauthor{\bsnm{Zhang}, \binits{S.}},
\bauthor{\bsnm{Liu}, \binits{Z.}},
\bauthor{\bsnm{Fu}, \binits{C.}}:
\bctitle{Tctrack: Temporal contexts for aerial tracking}.
In: \bbtitle{Proceedings of the IEEE/CVF Conference on Computer Vision and Pattern Recognition},
pp. \bfpage{14798}--\blpage{14808}
(\byear{2022})
\end{bchapter}
\endbibitem

%%% 71
\bibitem{chatfield2014return}
\begin{botherref}
\oauthor{\bsnm{Chatfield}, \binits{K.}},
\oauthor{\bsnm{Simonyan}, \binits{K.}},
\oauthor{\bsnm{Vedaldi}, \binits{A.}},
\oauthor{\bsnm{Zisserman}, \binits{A.}}:
Return of the devil in the details: Delving deep into convolutional nets.
arXiv preprint arXiv:1405.3531
(2014)
\end{botherref}
\endbibitem

%%% 72
\bibitem{HOG}
\begin{bchapter}
\bauthor{\bsnm{Dalal}, \binits{N.}},
\bauthor{\bsnm{Triggs}, \binits{B.}}:
\bctitle{Histograms of oriented gradients for human detection}.
In: \bbtitle{2005 IEEE Computer Society Conference on Computer Vision and Pattern Recognition (CVPR'05)},
vol. \bseriesno{1},
pp. \bfpage{886}--\blpage{893}
(\byear{2005}).
\bcomment{Ieee}
\end{bchapter}
\endbibitem

%%% 73
\bibitem{van2009learning}
\begin{barticle}
\bauthor{\bsnm{Van De~Weijer}, \binits{J.}},
\bauthor{\bsnm{Schmid}, \binits{C.}},
\bauthor{\bsnm{Verbeek}, \binits{J.}},
\bauthor{\bsnm{Larlus}, \binits{D.}}:
\batitle{Learning color names for real-world applications}.
\bjtitle{IEEE Transactions on Image Processing}
\bvolume{18}(\bissue{7}),
\bfpage{1512}--\blpage{1523}
(\byear{2009})
\end{barticle}
\endbibitem

%%% 74
\bibitem{AlexNet}
\begin{botherref}
\oauthor{\bsnm{Krizhevsky}, \binits{A.}},
\oauthor{\bsnm{Sutskever}, \binits{I.}},
\oauthor{\bsnm{Hinton}, \binits{G.E.}}:
Imagenet classification with deep convolutional neural networks.
Advances in neural information processing systems
\textbf{25}
(2012)
\end{botherref}
\endbibitem

%%% 75
\bibitem{FastRCNN}
\begin{bchapter}
\bauthor{\bsnm{Girshick}, \binits{R.}}:
\bctitle{Fast r-cnn}.
In: \bbtitle{Proceedings of the IEEE International Conference on Computer Vision},
pp. \bfpage{1440}--\blpage{1448}
(\byear{2015})
\end{bchapter}
\endbibitem

%%% 76
\bibitem{ResNet}
\begin{bchapter}
\bauthor{\bsnm{He}, \binits{K.}},
\bauthor{\bsnm{Zhang}, \binits{X.}},
\bauthor{\bsnm{Ren}, \binits{S.}},
\bauthor{\bsnm{Sun}, \binits{J.}}:
\bctitle{Deep residual learning for image recognition}.
In: \bbtitle{Proceedings of the IEEE Conference on Computer Vision and Pattern Recognition},
pp. \bfpage{770}--\blpage{778}
(\byear{2016})
\end{bchapter}
\endbibitem

%%% 77
\bibitem{FCOS}
\begin{bchapter}
\bauthor{\bsnm{Tian}, \binits{Z.}},
\bauthor{\bsnm{Shen}, \binits{C.}},
\bauthor{\bsnm{Chen}, \binits{H.}},
\bauthor{\bsnm{He}, \binits{T.}}:
\bctitle{Fcos: Fully convolutional one-stage object detection}.
In: \bbtitle{Proceedings of the IEEE/CVF International Conference on Computer Vision},
pp. \bfpage{9627}--\blpage{9636}
(\byear{2019})
\end{bchapter}
\endbibitem

%%% 78
\bibitem{FasterRCNN}
\begin{botherref}
\oauthor{\bsnm{Ren}, \binits{S.}},
\oauthor{\bsnm{He}, \binits{K.}},
\oauthor{\bsnm{Girshick}, \binits{R.}},
\oauthor{\bsnm{Sun}, \binits{J.}}:
Faster r-cnn: Towards real-time object detection with region proposal networks.
Advances in neural information processing systems
\textbf{28}
(2015)
\end{botherref}
\endbibitem

%%% 79
\bibitem{IoU-Net}
\begin{bchapter}
\bauthor{\bsnm{Jiang}, \binits{B.}},
\bauthor{\bsnm{Luo}, \binits{R.}},
\bauthor{\bsnm{Mao}, \binits{J.}},
\bauthor{\bsnm{Xiao}, \binits{T.}},
\bauthor{\bsnm{Jiang}, \binits{Y.}}:
\bctitle{Acquisition of localization confidence for accurate object detection}.
In: \bbtitle{Proceedings of the European Conference on Computer Vision (ECCV)},
pp. \bfpage{784}--\blpage{799}
(\byear{2018})
\end{bchapter}
\endbibitem

%%% 80
\bibitem{detone2018superpoint}
\begin{bchapter}
\bauthor{\bsnm{DeTone}, \binits{D.}},
\bauthor{\bsnm{Malisiewicz}, \binits{T.}},
\bauthor{\bsnm{Rabinovich}, \binits{A.}}:
\bctitle{Superpoint: Self-supervised interest point detection and description}.
In: \bbtitle{Proceedings of the IEEE Conference on Computer Vision and Pattern Recognition Workshops},
pp. \bfpage{224}--\blpage{236}
(\byear{2018})
\end{bchapter}
\endbibitem

%%% 81
\bibitem{sarlin2020superglue}
\begin{bchapter}
\bauthor{\bsnm{Sarlin}, \binits{P.-E.}},
\bauthor{\bsnm{DeTone}, \binits{D.}},
\bauthor{\bsnm{Malisiewicz}, \binits{T.}},
\bauthor{\bsnm{Rabinovich}, \binits{A.}}:
\bctitle{Superglue: Learning feature matching with graph neural networks}.
In: \bbtitle{Proceedings of the IEEE/CVF Conference on Computer Vision and Pattern Recognition},
pp. \bfpage{4938}--\blpage{4947}
(\byear{2020})
\end{bchapter}
\endbibitem

%%% 82
\bibitem{tan2020efficientdet}
\begin{bchapter}
\bauthor{\bsnm{Tan}, \binits{M.}},
\bauthor{\bsnm{Pang}, \binits{R.}},
\bauthor{\bsnm{Le}, \binits{Q.V.}}:
\bctitle{Efficientdet: Scalable and efficient object detection}.
In: \bbtitle{Proceedings of the IEEE/CVF Conference on Computer Vision and Pattern Recognition},
pp. \bfpage{10781}--\blpage{10790}
(\byear{2020})
\end{bchapter}
\endbibitem

%%% 83
\bibitem{GIoU}
\begin{bchapter}
\bauthor{\bsnm{Rezatofighi}, \binits{H.}},
\bauthor{\bsnm{Tsoi}, \binits{N.}},
\bauthor{\bsnm{Gwak}, \binits{J.}},
\bauthor{\bsnm{Sadeghian}, \binits{A.}},
\bauthor{\bsnm{Reid}, \binits{I.}},
\bauthor{\bsnm{Savarese}, \binits{S.}}:
\bctitle{Generalized intersection over union: A metric and a loss for bounding box regression}.
In: \bbtitle{Proceedings of the IEEE/CVF Conference on Computer Vision and Pattern Recognition},
pp. \bfpage{658}--\blpage{666}
(\byear{2019})
\end{bchapter}
\endbibitem

%%% 84
\bibitem{DIoU}
\begin{bchapter}
\bauthor{\bsnm{Zheng}, \binits{Z.}},
\bauthor{\bsnm{Wang}, \binits{P.}},
\bauthor{\bsnm{Liu}, \binits{W.}},
\bauthor{\bsnm{Li}, \binits{J.}},
\bauthor{\bsnm{Ye}, \binits{R.}},
\bauthor{\bsnm{Ren}, \binits{D.}}:
\bctitle{Distance-iou loss: Faster and better learning for bounding box regression}.
In: \bbtitle{Proceedings of the AAAI Conference on Artificial Intelligence},
vol. \bseriesno{34},
pp. \bfpage{12993}--\blpage{13000}
(\byear{2020})
\end{bchapter}
\endbibitem

\end{thebibliography}

\end{document}